\documentclass{elsarticle}
\usepackage{diagbox}
\usepackage{graphicx} 
\usepackage{epstopdf}
\usepackage{caption}
\usepackage{tabularx}
\usepackage{xcolor}%
\usepackage{colortbl,booktabs}%
\usepackage{multicol}
\usepackage{multirow}
\usepackage{booktabs}
\usepackage{threeparttable}
\usepackage{tabu}
\usepackage{colortbl,booktabs}%
\usepackage{subfigure}
\usepackage{epsfig}
\usepackage{graphicx}
\usepackage[labelfont=bf]{caption}

\usepackage{algorithm}
\usepackage{algorithmic}
\usepackage{graphics}
\usepackage[fleqn]{amsmath}  %
\usepackage{verbatim}
\usepackage{makecell}
\usepackage{setspace}
\usepackage{geometry}
\usepackage{enumerate}%
\usepackage{etoolbox}
\makeatletter
\g@addto@macro\normalsize{%
    \setlength\abovedisplayskip{2pt} %
    \setlength\belowdisplayskip{2pt}
    \setlength\abovedisplayshortskip{2pt}
    \setlength\belowdisplayshortskip{2pt}%
}
\makeatother

\makeatletter   %
\renewcommand{\maketag@@@}[1]{\hbox{\m@th\normalsize\normalfont#1}}%
\makeatother

\usepackage{hyperref}
\usepackage{cleveref} %
\newtheorem{mydef}{Definition}
\newtheorem{theorem}{Theorem}
\newtheorem{example}{Example}
\newtheorem{property}{Property}
\newtheorem{remark}{Remark}
\usepackage{amssymb}
\biboptions{numbers,sort&compress}
\usepackage{array, caption, threeparttable}
\usepackage[font=small,labelfont=bf,labelsep=none]{caption}

\biboptions{numbers,sort&compress}
\hypersetup{hidelinks}
\hypersetup{
colorlinks=true,
linkcolor=blue
}
\journal{arXiv}

\usepackage{doi}

\begin{document}
\begin{spacing}{1.1}

\begin{frontmatter}
\title{Sequential three-way group decision-making for double hierarchy hesitant fuzzy linguistic term set}
\author[a,b]{Nanfang Luo}
\author[a,b,c]{Qinghua Zhang\corref{mycorrespondingauthor}}
\cortext[mycorrespondingauthor]{Corresponding author}
\ead{zhangqh@cqupt.edu.cn}
\author[a,b]{Qin Xie}
\author[b,c]{Yutai Wang}
\author[b,c]{Longjun Yin}
\author[b,c,d]{Guoyin Wang}
\address[a]{Chongqing Key Laboratory of Tourism Multisource Data Perception and Decision Ministry of Culture and Tourism, Chongqing University of Posts and Telecommunications, Chongqing 400065, China}
\address[b]{Chongqing Key Laboratory of Computational Intelligence, Chongqing University of Posts and Telecommunications, Chongqing 400065, China}
\address[c]{Key Laboratory of Big Data Intelligent Computing, Chongqing University of Posts and Telecommunications, Chongqing 400065, China}
\address[d]{College of Computer and Information Science, Chongqing Normal University, Chongqing, 401331, China}

\begin{abstract}\label{abstract}
Group decision-making (GDM) characterized by complexity and uncertainty is an essential part of various life scenarios.
Most existing researches lack tools to fuse information quickly and interpret decision results for partially formed decisions.
This limitation is particularly noticeable when there is a need to improve the efficiency of GDM.
To address this issue, a novel multi-level sequential three-way decision for group decision-making (S3W-GDM) method is constructed from the perspective of granular computing.
This method simultaneously considers the vagueness, hesitation, and variation of GDM problems under double hierarchy hesitant fuzzy linguistic term sets (DHHFLTS) environment.
First, for fusing information efficiently, a novel multi-level expert information fusion method is proposed, and the concepts of expert decision table and the extraction/aggregation of decision-leveled information based on the multi-level granularity are defined.
Second, the neighborhood theory, outranking relation and regret theory (RT) are utilized to redesign the calculations of conditional probability and relative loss function.
Then, the granular structure of DHHFLTS based on the sequential three-way decision (S3WD) is defined to improve the decision-making efficiency, and the decision-making strategy and interpretation of each decision-level are proposed. Furthermore, the algorithm of S3W-GDM is given.
Finally, an illustrative example of diagnosis is presented, and the comparative and sensitivity analysis with other methods are performed to verify the efficiency and rationality of the proposed method.
\end{abstract}


\end{frontmatter}


\section{Introduction}\label{Introduction}
Modern medical decision-making privileges rapid and accurate diagnosis to improve patient care.
While a comprehensive examination can accomplish this goal, it is often time-consuming and can be emotionally stressful for the patient, especially during the diagnostic phase of the disease.
An effective diagnostic strategy that quickly determines the need for further examination after the initial assessment would be of great advantage.
This would not only shorten the diagnostic time but also reduce uncertainty and psychological stress for the patient.
This challenge is particularly evident in complex diseases such as systemic lupus erythematosus (SLE).
The diagnosis of SLE is characterized by a wide range of symptoms and requires multidisciplinary collaboration between experts \cite{zirkzee2012prospective}.
This collaborative process, known as group decision-making (GDM), inherently involves uncertainty in the form of vagueness, hesitation, and variation.
A patient-centered approach through GDM can optimize the diagnostic process while prioritizing patient well-being, reflecting the dual needs of efficiency and humanistic care in modern medicine.

\subsection{A brief review of GDM}
GDM is a common process in many fields, especially when dealing with complex issues that require multiple experts.
GDM can be challenging due to the inherent complexity of the decision-making process and the involvement of multiple experts with varying expertise and perspectives \cite{herrera2020revisiting}.
One of the main challenges lies in effectively combining and utilizing the information collected from different experts.
This involves ensuring that all relevant information is considered, that experts' opinions are not distorted and the information is integrated in a way that leads to sound decision-making.
Existing research is not limited to the construction of models that quantitatively evaluate information but qualitative representations that match human usage habits are becoming a hot topic for more model construction research.
Rodriguez et al. \cite{rodriguez2013group} significantly advanced GDM by introducing a model that effectively utilizes hesitant fuzzy linguistic term sets (HFLTS) to handle comparative linguistic expressions, enhancing the precision and flexibility of decision information representation.
Pang et al. \cite{PangPLT} introduced probabilistic linguistic term sets (PLTS) for more accurately collecting and expressing decision information in multi-attribute GDM.
The most straightforward way to evaluate the information is to use an aggregation operators \cite{xu2015hesitant,liu2018some} to fuse the information in order to obtain a comprehensive evaluation information.
The comprehensive evaluation information is then processed by classical multi-attribute decision-making methods, including TOPSIS, VIKOR, and MULTIMOORA.
Another challenge is dealing with the uncertainty that arises from the diverse backgrounds and experiences of the experts.
Individual preferences, knowledge gaps, and varying interpretations of information can all contribute to uncertainty in the decision-making process.
In studying the decision-making behaviours of decision-makers, the close connection between behavioral economics \cite{kahneman2013prospect,loomes1982regret} and decision-making science has led to a number of noteworthy results \cite{liu2019extended,zhang2016regret}.
The emergency decision-making usually involves multiple experts as well, which is also a challenge in GDM problems.
The dynamic character of emergencies further adds to the complexity of  GDM problems, including temporary hospital site selection \cite{chen2023integrated}, emergency management \cite{sun2021new}.
Despite these efforts, GDM remains a complex and challenging area, and requires continuous research to develop more effective ways of dealing with uncertainty and making wiser decisions in a better way \cite{herrera2020revisiting}.

\subsection{A brief review of DHHFLTS}
Double hierarchy hesitant fuzzy linguistic term set (DHHFLTS) is a composite information collection tool that plays a crucial role in collecting vague and hesitant decision-making information.
The uniqueness of DHHFLTS is that it embeds a layer of linguistic scale based on the HFLTS.
This layer of embedded linguistic term set serves as a subscale of the main linguistic scale, providing a richer foundation for qualitative expression.
This dual structure allows DHHFLTS to effectively capture and represent the nuanced opinions of experts.
In recent years, researchers have explored the application of DHHFLTS in various fields and extended its capability to handle more complex decision-making scenarios.
For instance, Gou et al. \cite{GOU2017multimoora} applied DHHFLTS to the MULTIMOORA method in their study, demonstrating its flexibility and effectiveness in complex decision environments.
To quantify the relationships between different DHHFLTS, Gou et al. \cite{gou2018multiple} proposed various distance measurement models, enhancing the accuracy and effectiveness of decision analysis.
As a type of linguistic data, DHHFLTS has been gradually integrated into traditional multi-attributes decision-making methods.
Liu et al. \cite{liu2019novel} combined DHHFLTS with traditional aggregation operators to propose a new multi-attributes decision-making model.
Gou et al. \cite{gou2021probabilistic} studied probabilistic DHHFLTS and integrated it with the VIKOR method.
In addition, DHHFLTS has achieved some results through theoretically extended researches \cite{liu2022magdm,montserrat2019free}.
Some researches have increasingly validated the potential of DHHFLTS in dynamic decision-making scenarios.
Liu et al. \cite{liu2023improved} improved the ELECTRE II method, extending the application scope of DHHFLTS in emergency logistics provider selection.
Also, Gou et al. \cite{gou2024medical} proposed an improved ORESTE method utilizing linguistic preference orderings to evaluate medical resource allocation in public health emergencies, which demonstrates that DHHFLTS has a better handling capability for dynamic problems in healthcare management.
Currently, DHHFLTS has also produced noteworthy works.
Cheng et al. \cite{cheng2024large} introduced a large-scale GDM model that considers risk attitudes and dynamic role changes, showcasing the high adaptability of DHHFLTS in a practical emergency problem.
Furthermore, a large-scale GDM opinions updating model driven by autonomous learning was developed \cite{cheng2024opinions}, further expanding the application scope of DHHFLTS in GDM.
DHHFLTS increases the flexibility of expert linguistic expression and provides support for information representation in more complex GDM.

\subsection{A brief review of S3WD}
\begin{figure}[htbp]
  \centering
  \includegraphics[width=6cm]{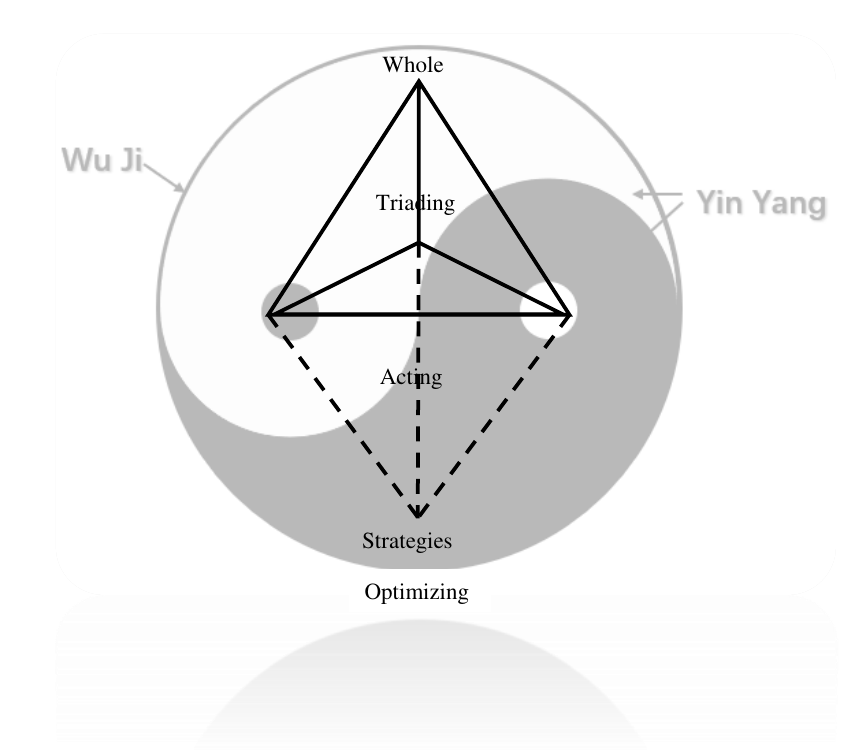}
  \caption{\textbf{.} TAO and DAO.}
  \label{fig1}
\end{figure}

Sequential three-way decision (S3WD)  is a typical model of three-way decision (3WD) for handling dynamic problems.
The recent comprehensive work \cite{yao2023dao} offers a new interpretation of the 3WD framework, TAO (Triading-Acting-Optimizing).
TAO provides a theory for dynamic decisions that aligns with traditional Chinese culture.
It has delved into four types of triadic structures of three worlds and explained the Dao, the way of three-world thinking by using examples.
The concept of TAO, similar to the "Wuji" state in traditional Chinese culture, breaks the binary or bipolar thinking by introducing an intermediate state for handling uncertainty.
Fig. \ref{fig1} exemplifies this using the "Yin Yang" symbol.
The outer circle, representing "Wuji", gives rise to the opposing poles of "Yin" and "Yang" within "Tai Chi".
In "Tai Chi", the opposites of "Yin" and "Yang" are unified and constantly transform, producing "Change"\footnote{Xinbo Gao. "Understanding Chongqing Nanshan and Promoting the Spirit in CQUPT." January 1, 2024. \url{https://mp.weixin.qq.com/s/dD1U1xhkuB6uZuNSLRY_Dw}
}.
This dynamic transform interplay drives the world, just as the interaction between heaven, earth, and man influences decision-making.
Yao \cite{yao2013granular} has studied S3WD from the perspective of granular computing in 2013 and analysed that the decision cost of this dynamic 3WD is smaller than that of two-way decision (2WD).
2WD is mostly found in established GDM methods.
Regarding decision objects, the decision results are either accepted or rejected, with a risk of decision errors.
Integrating the concept of 3WD into GDM can effectively reduce the risk of decision errors.
Recently, very noteworthy results have been achieved in GMD methods for 3WD under linguistic terms environments.
Liu et al. \cite{liu2024three} proposed a 3WGDM method, which combines comparative linguistic expressions and personalised numerical scales to improve the accuracy of decision making and effectively solve the problem of uncertainty in GDM.
Yang et al. \cite{yang2024dynamic} integrated basic uncertain linguistic information and decision-theoretic rough sets into a dynamic 3WD framework based on the temporal dimension, enhancing the flexibility and accuracy of the multi-attribute decision-making process and making a valuable contribution.
The study of S3WD in multi-levels of granularity decision information processing is more suitable for improving efficiency and reducing decision costs.
This multi-level processing follows the dynamic expansion of information from coarse-grained to fine-grained.
When the available information is insufficient to support a decision (accept or reject) at the current decision-level, additional information is added to support the next decision-level, reducing the decision cost for some objects that can be accepted or rejected.
The approximate computing capability of S3WD makes it an important research topic in the field of granular computing.
Yang et al. \cite{yang2020multilevel} considered the rules of hierarchical granulation in both horizontal and vertical directions, and proposed a general multi-level neighborhood sequential decision approach that improves the applicability on cognitive science applications.
It has also been correspondingly extended to different information environment \cite{hu2022novel,zhang2019sequential}.
Similarly, some work has used S3WD to solve GDM problems characterised by uncertainty.
Wang et al. \cite{wang2020sequential} integrated social influence dynamics into a S3WD framework, and proposed a STWMAGDM approach, addressing high decision risk and uncertainty.
Wang et al. \cite{wang2020bwm} builded a multigranulation S3WD model with cost-sensitive and obtained the expert GDM results using MULTIMOORA ranking the classification.
Further, the implementation of S3WD with multi-level granularity decision information processing will provide a better problem solving model for GDM.

\subsection{Motivations and contributions of this paper}
It is difficult for a single decision-maker to make a reasonable decision.
Similar to SLE multidisciplinary collaboration, GDM can provide effective solutions to decision problems in complex scenarios.
These scenarios often involve multiple experts evaluating multiple alternatives based on various attributes.
However, with the development of information science, the amount of decision information has greatly increased and the information dynamics changes rapidly.
GDM problems become more and more uncertain, and higher requirements are imposed on decision models.

Efficiency in GDM is crucial.
The granularity of information for decision is becoming increasingly refined, but using all available information indiscriminately would increase decision costs.
For example, doctors diagnosing abdominal pain do not require patients to undergo all related examinations.
If there are no concerning symptoms like bloody stool, vomiting, weight loss, or fever, then observation might be the most appropriate initial course of action.
If symptoms improve, there is no need for further intervention.
However, if the pain persists, an endoscopy might be necessary.
This strategy avoids unnecessary procedures, as endoscopies can have side effects.

Reliability in GDM is also crucial.
While expert preferences are intuitive, they often exhibit vagueness and hesitation.
Experts find it challenging to use precise numerical evaluations, which makes it difficult to collect decision information intuitively.
Qualitative expressions, being closer to human habits, have become a hot research.
However, existing single linguistic terms are often inflexible and do not easily meet human expression habits in applications.
Additionally, individuals tend to anchor their psychological preferences, which should be considered in qualitative expressions.
3WD provides a buffer zone by allowing an indeterminate state for alternatives.
However, does the indeterminate state remain unchanged?
With the addition of new decision information and finer granularity, some alternatives in the indeterminate state can be further decided upon, while those already decided upon do not increase decision costs.
This process is S3WD.
Usually, the static character is relative, whereas constant variation is the norm.
A dynamic approach to decision-making should be the strategic thinking provided by 3WD.

Therefore, improving efficiency and reducing decision costs are important issues.
This research is motivated by three key observations.

(1) Existing researches \cite{chen2023integrated,sun2021new,liu2023improved,krishankumar2020multi,krishankumar2019framework} primarily focus on 2WD, where decision results are typically either accepted or rejected, lacking a buffer zone of them.
In GDM scenarios, the number of experts, attributes, and costs associated with each decision increase the overall decision burden.
Few works consider adjusting the process of GDM information fusion to reduce decision costs.

(2) Few works focus on both S3WD and DHHFLTS.
DHHFLTS, as a composite linguistic term set, has been explored in various fields.
It is essential to promote research on GDM in qualitative dynamic problem-solving, enhancing the flexibility and rationality of DHHFLTS applications in S3WD.

(3) There is still a need to simultaneously address the characteristics of vagueness, hesitation, and variation in GDM problems.
Existing S3W-GDM methods based on granular computing \cite{wang2020bwm,hu2022novel} may not be applicable to deal the composite linguistic term model, and also lack the integration with PT \cite{kahneman2013prospect} or RT \cite{loomes1982regret}, etc. to further explore uncertainty.

With these motivations, the main contributions of this research are concluded as follows:

(1) A conditional probability calculation method for scenarios involving DHHFLTS based on the neighborhood theory and outranking relation is proposed to achieve greater generality.
Additionally, the concept of relative perceived utility is proposed to replace relative loss to better reflect real-world psychological preferences.

(2) The decision-level extracted and aggregated methods based on multi-level of granularity are introduced.
This provides a more flexible and relevant approach for handling complex GDM problems, allowing for a simplified analysis while preserving essential information.

(3) A novel method of S3W-GDM is established from granular computing perspective, which overcomes the shortcomings of existing GDM researches.
More importantly, compared with the existing 2WD methods, the proposed method is more effective and close to reality.

\begin{figure}[htbp]
  \centering
  \includegraphics[width=12cm]{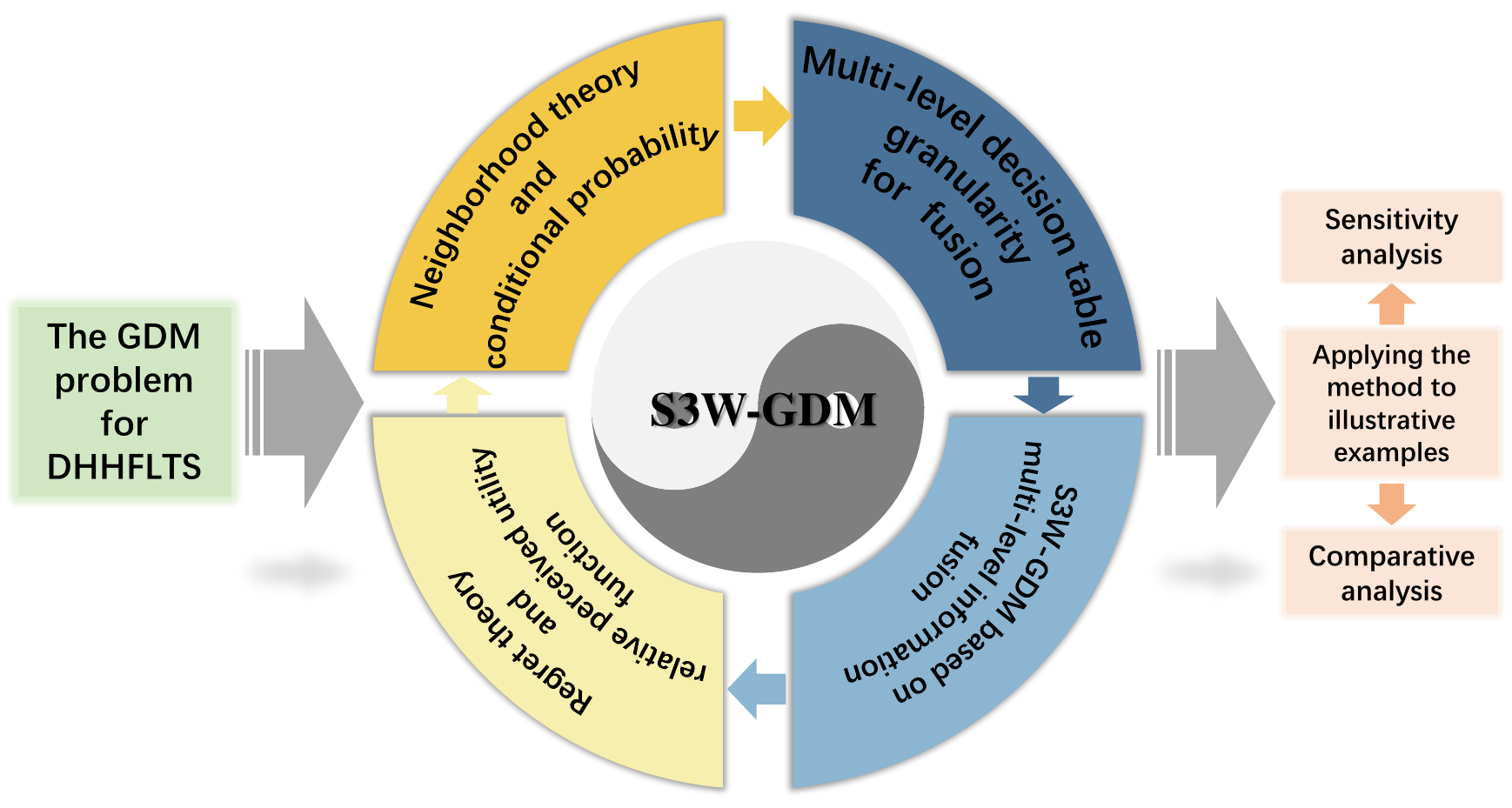}
  \caption{\textbf{.} The framework of this research.}
  \label{fig2}
\end{figure}

The structure of this paper is as follows.
Section \ref{Preliminaries} provides the backgroud of DHHFLTS, the S3WD model and RT.
In Section \ref{3WD model}, a S3WD model is developed based on neighborhood theory and RT.
In Section \ref{processing}, the S3W-GDM method for DHHFLTS based on multi-level of granularity is constructed and the related algorithms model is present.
Section \ref{illustrative and comparative} provides an illustrative example to show the applicability of the multi-level S3W-GDM method.
In Seection \ref{section6}, the comparative and sensitivity analysis verify the validity and rationality of this model.
Section \ref{conclusion} summarizes the conclusions and future work.
In addition, an overall diagram of the paper is given in Fig. \ref{fig2}.

\section{Preliminaries}\label{Preliminaries}
This section provides a brief review of concepts related to DHHFLTS, S3WD, and RT.

\subsection{DHHFLTS}\label{DHHFLTS}
Gou et al. \cite{GOU2017multimoora} proposed the DHHFLTS,  which can be represented to describe complex linguistic information more accurately with the form of ``adverb + adjective".

\begin{mydef}\cite{GOU2017multimoora}\label{mydef1}
Let $U$ be a finite universal set. A double hierarchy hesitant fuzzy linguistic term set (DHHFLTS) denotes as $H$ on $U$  is in mathematical form as follows:
$H = \left\{ {\left\langle {x,{h_{{S_O}}}\left( x \right)} \right\rangle \left| {x \in U} \right.} \right\}$,
where ${h_{{S_O}}}\left( x \right){\rm{ = }}\left\{ {{s_{{\phi _l}\left\langle {{o_{{\varphi _l}}}} \right\rangle }}\left( x \right)\left| {{s_{{\phi _l}\left\langle {{o_{{\varphi _l}}}} \right\rangle }} \in {S_O};l = 1,2,...,L;} \right.{\phi _l} = \left[ { - \tau ,\tau } \right];} \right.$\\
$\left. {{\varphi _l} = \left[ { - \varsigma ,\varsigma } \right]} \right\}$ called double hierarchy hesitant fuzzy linguistic element (DHHFLE) is a set of some value denotes the possible degree of $x$ to double hierarchy linguistic term set (DHLTS) ${S_O}$, ${s_{{\phi _l}\left\langle {{o_{{\varphi _l}}}} \right\rangle }}\left( x \right)$ is the continuous terms in ${S_O}$, $L$ is the number of DHLTSs in ${h_{{S_O}}}\left( x \right)$.
\end{mydef}
To better understand DHHFLTS,  two examples for illustration are given.

\begin{example}\label{Example 1}
Suppose $S = \left\{ {{s_{ - 3}} = none,{s_{ - 2}} = very\;low,{s_{ - 1}} = low,{s_0} = medium,{s_1} = high,} \right.$\\$\left. {{s_2} = very\;high,{s_3} = perfect} \right\}$ is the first hierarchy linguistic term set.
And $O = \left\{ {{o_{ - 2}} = far} \right.$\\$\left. {from,{o_{ - 1}} = a\;little,{o_0} = just\;right,{o_1} = much,{o_2} = very\;much} \right\}$ is the second hierarchy linguistic term sets, which is the subscale of linguistic embedded in the main linguistic $s_1$ scale.
When some complicated linguistic terms are described as - ``a little high" and ``between much medium and just right very high".
They could use DHHFLEs $\left\{ {{s_{1\left\langle {{o_-1}} \right\rangle }}} \right\}$ and $\left\{ {{s_{0\left\langle {{o_1}} \right\rangle }},{s_1},{s_{2\left\langle {{o_0}} \right\rangle }}} \right\}$.
\end{example}

\begin{figure}[htbp]
  \centering
  \includegraphics[width=8cm]{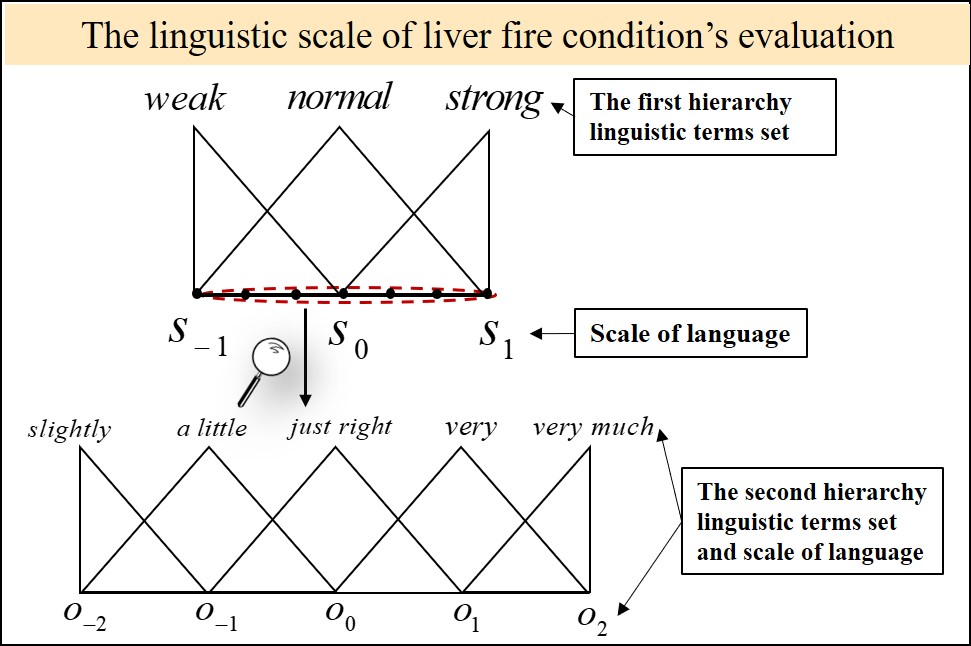}
  \caption{\textbf{.} The linguistic scale of liver fire conditions evaluation.}
\label{liverfire}
\end{figure}

\begin{example}\label{Example 2}
In traditional Chinese medicine (TCM), diagnosis often relies on sensory methods like inspection, smelling, inquiry, and palpation. Verbal descriptions are a cornerstone of this evaluation process. According to Fig. \ref{liverfire}, a TCM practitioner assessed the liver fire condition of three patients and subsequently developed a linguistic scale to categorize their conditions.
These scales are the first hierarchy linguistic terms set $S = \left\{ {{s_{ - 1}} = weak,{s_0} = normal,{s_1} = strong} \right\}$ and the second hierarchy linguistic terms set $O = \left\{ {{o_{ - 2}} = slightly,{o_{ - 1}}} \right.$
$\left. { = a\;little,{o_0} = just\;right,{o_1} = very,{o_2} = very\;much} \right\}$.
The diagnosis resulted in three patients evaluations are denoted as ${h_{{s_o}1}} = \left\{ {{s_{1\left\langle {{o_{ - 1}}} \right\rangle }},{s_{1\left\langle {{o_{ - 2}}} \right\rangle }}} \right\}$, ${h_{{s_o}2}} = \left\{ {{s_{0\left\langle {{o_0}} \right\rangle }}} \right\}$ and ${h_{{s_o}3}} = \left\{ {{s_{ 1\left\langle {{o_{ - 2}}} \right\rangle }},{s_{0\left\langle {{o_0}} \right\rangle }}} \right\}$.
\end{example}

\begin{mydef} \cite{GOU2017multimoora}\label{mydef2}
Let ${S_O} = \left\{ {{s_{{\phi _l}\left\langle {{o_{{\varphi _l}}}} \right\rangle }}\left| {{\phi _l} \in \left[ { - \tau ,\tau } \right]} \right.,{\varphi _l} \in \left[ { - \varsigma ,\varsigma } \right]} \right\}$ be a DHLTS, ${h_{{S_O}}}{\rm{ = }}$\\
$\left\{ {{s_{{\phi _l}\left\langle {{o_{{\varphi _l}}}} \right\rangle }}\left| {{s_{{\phi _l}\left\langle {{o_{{\varphi _l}}}} \right\rangle }} \in {S_O};l = 1,2,...,L} \right.} \right\}$ be a DHHFLE, and ${h_\gamma }{\rm{ = }}\left\{ {{\gamma _l}\left| {{\gamma _l} \in \left[ {0,1} \right];l = 1,2,...,L} \right.} \right\}$ be a set of hesitant fuzzy element (HFE).
There are two transformation functions $f$ and ${f^{ - 1}}$ as follows:
\begin{equation}\label{eq1}
f:\left[ { - \tau ,\tau } \right] \times \left[ { - \varsigma ,\varsigma } \right] \to \left[ {0,1} \right],
f\left( {{\phi _l},{\varphi _l}} \right) = \frac{{{\varphi _l} + \left( {\tau  + {\phi _l}} \right)\varsigma }}{{2\varsigma \tau }},
\end{equation}

\begin{equation}\label{eq2}
\begin{split}
 & f^{-1}:[0,1] \rightarrow [-\tau, \tau] \times [-\varsigma, \varsigma], f^{-1}(\gamma_{l}) = [2\tau \gamma_{l} - \tau]\langle o_{\varsigma(2\tau \gamma_{l} - \tau - [2\tau \gamma_{l} - \tau])} \rangle \\
 & \qquad\qquad\qquad\qquad\qquad\qquad\qquad\qquad= [2\tau \gamma_{l} - \tau] + 1 \langle o_{\varsigma((2\tau \gamma_{l} - \tau - [2\tau \gamma_{l} - \tau]) - 1)} \rangle.
\end{split}
\end{equation}
The transformation functions $F$ and ${F^{ - 1}}$ are established as follows:
\begin{equation}\label{eq3}
F:\Phi  \times \Psi  \to \Theta ,
F\left( {{h_{{S_O}}}} \right) = \left\{ {{\gamma _l}\left| {{\gamma _l} = } \right.f\left( {{\phi _l},{\varphi _l}} \right)} \right\} = {h_\gamma },
\end{equation}

\begin{equation}\label{eq4}
{F^{ - 1}}:\Theta  \to \Phi  \times \Psi ,
{F^{ - 1}}\left( {{h_\gamma }} \right) = \left\{ {{s_{{\phi _l}\left\langle {{o_{{\varphi _l}}}} \right\rangle }}\left| {{\phi _l}\left\langle {{\varphi _l}} \right\rangle  = } \right.{f^{ - 1}}\left( {{\gamma _l}} \right)} \right\} = {h_{{S_O}}}.
\end{equation}
\end{mydef}

\begin{mydef} \cite{GOU2017multimoora}
Let ${S_O}$ be a continuous DHLTS, ${h_{{S_O}}}$, ${h_{{S_O}}}_1$, and ${h_{{S_O}}}_2$ be any three DHHFLEs, and $\mu $ be a constant. Then some operational laws between DHHFLEs are defined as follows:
\vspace{-0.2cm}
\begin{equation}\label{}
Addition:
\begin{aligned}
\begin{array}{l}
{h_{{S_O}}}_1 \oplus {h_{{S_O}}}_2 = {F^{ - 1}}\left( {\mathop  \cup \limits_{{\gamma _1} \in F\left( {{h_{{S_O}}}_1} \right),{\gamma _2} \in F\left( {{h_{{S_O}}}_2} \right)} \left\{ {{\gamma _1} + {\gamma _2} - {\gamma _1}{\gamma _2}} \right\}} \right),
\end{array}
\end{aligned}
\end{equation}

\begin{equation}\label{}
Multiplication:
\begin{aligned}
\begin{array}{l}
\mu {h_{{S_O}}} = {F^{ - 1}}\left( {\mathop  \cup \limits_{\gamma  \in F\left( {{h_{{S_O}}}} \right)} \left\{ {1 - {{\left( {1 - \gamma } \right)}^\mu }} \right\}} \right),
\end{array}
\end{aligned}
\end{equation}

\begin{equation}\label{}
Power:
\begin{aligned}
\begin{array}{l}
{\left( {{h_{{S_O}}}} \right)^\mu } = {F^{ - 1}}\left( {\mathop  \cup \limits_{\gamma  \in F\left( {{h_{{S_O}}}} \right)} \left\{ {{\gamma ^\mu }} \right\}} \right),
\end{array}
\end{aligned}
\end{equation}

\begin{equation}\label{}
Complementary:
\begin{aligned}
\begin{array}{l}
{\left( {{h_{{S_O}}}} \right)^C} = {F^{ - 1}}\left( {\mathop  \cup \limits_{\gamma  \in F\left( {{h_{{S_O}}}} \right)} \left\{ {1 - \gamma } \right\}} \right).
\end{array}
\end{aligned}
\end{equation}
\label{}
\end{mydef}

According to the definition of DHHFLE, there are situations where different DHHFLEs have different numbers of DHLTs.
In order to have a reasonable normalization process and to keep all the integrity of the linguistic information during the computation, Gou et al.  \cite{gou2019group} developed a linguistic expected-value for DHHFLE.
\begin{mydef} \cite{gou2019group}
\label{mydef3}
Let ${S_O} = \left\{ {{s_{{\phi _l}\left\langle {{o_{{\varphi _l}}}} \right\rangle }}\left| {{\phi _l} \in \left[ { - \tau ,\tau } \right]} \right.,{\varphi _l} \in \left[ { - \varsigma ,\varsigma } \right]} \right\}$ be a continuous DHLTS, ${h_{{S_O}}}{\rm{ = }}$$\left\{ {{s_{{\phi _l}\left\langle {{o_{{\varphi _l}}}} \right\rangle }}\left| {{s_{{\phi _l}\left\langle {{o_{{\varphi _l}}}} \right\rangle }} \in {S_O};l = 1,2,...,L} \right.} \right\}$ be a DHHFLE,
$\Phi  \times \Psi $ be the set of all DHHFLEs over ${S_O}$.
Then a linguistic expected-value of ${h_{{S_O}}}$ is obtained as follows:
\begin{equation}\label{functionle}
le:\Phi  \times \Psi  \to {S_O},le\left( {{h_{{S_O}}}} \right) = \frac{1}{L} \oplus {s_{{\phi _l}\left\langle {{o_{{\varphi _l}}}} \right\rangle }} = {s_{le\left( {{\phi _l}} \right)\left\langle {{o_{le\left( {{\varphi _l}} \right)}}} \right\rangle }}.
\end{equation}
where $le\left( {{\phi }_{l}} \right)=\frac{1}{L}\sum\limits_{l=1}^{L}{{{\phi }_{l}}}$ and $le\left( {{\varphi }_{l}} \right)=\frac{1}{L}\sum\limits_{l=1}^{L}{{{\varphi }_{l}}}$.
\end{mydef}
\begin{example}\label{Example le}
From Example \ref{Example 2}, patient 1's liver fire condition is ${h_{{s_o}1}} = \left\{ {{s_{1\left\langle {{o_{ - 1}}} \right\rangle }},{s_{1\left\langle {{o_{ - 2}}} \right\rangle }}} \right\}$.
The numbers of DHLTs are 2 and can be normalized by Eq.(\ref{functionle}) as
$le\left( {{h_{{s_o}1}}} \right) = \left\{ {{s_{1\left\langle {{o_{ - 1.5}}} \right\rangle }}} \right\}$.
TCM practitioner, through questioning, gets the condition of liver fire of patient 1 as ``a little strong and slightly strong", and through the $le$ function, the hesitation in ``a little strong" and ``slightly strong" is neutralized, thus achieving the purpose of normalization.
\end{example}

\subsection{Multi-level structure of S3WD}\label{S3WD}
The three-way thinking of granular computing proposed by Yao \cite{yao2013granular}, states that a granular structure comprises three key elements: granules, layers, and hierarchical structures.
Granules are the fundamental units of information, while layers are composed of granules with the same granularity level.
These layers, when arranged according to their granularity levels, form a multi-level structure.
This concept is particularly relevant to dynamic decision-making problems, where a multi-level structure with increasing granularity from coarse to fine can be achieved by progressively increasing conditional attributes.

Given a quadruple of decision table ${\left( {DT} \right)_i} = \left( {{U_i},{Z_i},{V_i},{g_i}} \right)$$\left( {i = 1,2, \ldots ,k} \right)$, at the $i$th decision-level, $U_i$ is a finite and nonempty universal set of alternatives.
$Z_i$ is a finite and nonempty subset of conditional attributes, $A$ is the set of all conditional attributes, satisfies ${Z_1} \subseteq {Z_2} \subseteq  \ldots  \subseteq {Z_i} \subseteq  \ldots  \subseteq {Z_k} \subseteq A$, and $V_i$ denotes the domain of the conditional attributes.
${g_i}:{U_i} \times {A_i} \to {V_i}$ denotes an information function mapping. $\forall {Z_i} \subseteq A$, the equivalence relation ${R_{{Z_i}}}$ induced by ${Z_i}$ on the universe ${U_i}$ is defined by:
\begin{equation}
{R_{{Z_i}}} = \left\{ {\left( {x,y} \right) \in {U_i} \times {U_i}\left| {{g_i}\left( {x,a} \right) = {g_i}\left( {y,a} \right)} \right.,\forall a \in {Z_i}} \right\}.
\end{equation}

With the equivalence relations induced by different subsets of conditional attributes $Z_i$, the universal set $U_i$ can be partitioned.
Then the multi-level granular structure can be obtained, denoted as $Gl = \left\{ {G{l_1},G{l_2}, \ldots ,G{l_i}, \ldots ,G{l_k}} \right\}$.

\begin{mydef} \cite{wang2020sequential}\label{mydef4}
Given a multi-level granular structure $Gl = \left\{ G{l_1}, G{l_2}, \ldots , G{l_i}, \ldots , G{l_k} \right\}$, define each granular structure $G{l_i} = \left( {{U_i},{Z_i},pr\left( {{\pi _i}\left| x \right.} \right),{\Lambda _i}} \right)$.
In the $i$th decision-level's granular structure $G{l_i}$, $x \in {U_i}$ is an alternatives, ${Z_i}$ denotes the subset of conditional attributes, $pr\left( {{\pi _i}\left| x \right.} \right)$ denotes the conditional probability of the alternatives $x$ and ${ \Lambda _i}$ denotes loss functions.
\end{mydef}

Taking advantage of  the S3WD concept, Bayesian decision process is modeled using a state set ${\Gamma _i} = \left\{ {{\pi _i},\neg {\pi _i}} \right\}$ and a set $\mathfrak{A} = \left\{ {{\partial _P},{\partial _B},{\partial _N}} \right\}$ of three actions.
These states ${\pi _i}$ and $\neg {\pi _i}$ are complementary, while actions ${\partial _P}$, ${\partial _B}$, and ${\partial _N}$ represent the decision choices of classifying an alternative into acceptance region, uncertainty region and rejection region at the $i$th decision-level, as shown in Table \ref{lossfunction}.

\begin{table}[htbp]
\centering
\renewcommand{\arraystretch}{1.33}%
\setlength{\abovecaptionskip}{1.5pt}%
\captionsetup{justification=centering, singlelinecheck=false, labelsep=quad}
\caption{The unit loss function of $x$.}
\label{lossfunction}
\setlength{\tabcolsep}{4mm}{
\begin{tabular}{cc|cc}
\hline
\multicolumn{2}{c|}{}                              & ${\pi _i}$   & $\neg {\pi _i}$ \\ \hline
\multicolumn{1}{c|}{\multirow{3}{*}{$x$}} & ${{\partial _P}}$ & ${\lambda _i^{PP}}$    & ${\lambda _i^{PN}}$ \\ \cline{2-2}
\multicolumn{1}{c|}{}                        & ${{\partial _B}}$ & ${\lambda _i^{BP}}$ & ${\lambda _i^{BN}}$  \\ \cline{2-2}
\multicolumn{1}{c|}{}                        & ${{\partial _N}}$ & ${\lambda _i^{NP}}$ & ${\lambda _i^{NN}}$ \\ \hline
\end{tabular}
}
\end{table}

The unit loss function is the result of $x$ in the granular structure $G{l_i}$ under a conditional attribute $a$($a \in {Z_i}$), which is composed of the set ${ \Lambda _i} = \left\{ {\lambda _i^{PP},\lambda _i^{BP},\lambda _i^{NP},\lambda _i^{PN},\lambda _i^{BN},\lambda _i^{NN}} \right\}$.
$\lambda _i^{PP}$, $\lambda _i^{BP}$, and $\lambda _i^{NP}$ are the losses incurred for taking actions of ${\partial _P}$, ${\partial _B}$, and ${\partial _N}$, respectively, when an alternative belongs to ${\pi _i}$. Similarly, $\lambda _i^{PN}$, $\lambda _i^{BN}$, and $\lambda _i^{NN}$ express the losses incurred for taking the same actions when the alternative belongs to $\neg {\pi _i}$.
The loss function satisfy the following relationship: $\lambda _i^{PP} \le \lambda _i^{BP} \le \lambda _i^{NP}$ and $\lambda _i^{PN} \le \lambda _i^{BN} \le \lambda _i^{NN}$.
The expected losses of taking actions ${\partial _P}$, ${\partial _B}$, and ${\partial _N}$ at the $i$th decision-level are calculated as follows:
\begin{equation}\label{}
\begin{array}{l}
E{L_i}\left( {{\partial _P}\left| x \right.} \right) = \lambda _i^{PP}pr\left( {{\pi _i}\left| x \right.} \right) + \lambda _i^{PN}pr\left( {\neg {\pi _i}\left| x \right.} \right),\\
E{L_i}\left( {{\partial _B}\left| x \right.} \right) = \lambda _i^{BP}pr\left( {{\pi _i}\left| x \right.} \right) + \lambda _i^{BN}pr\left( {\neg {\pi _i}\left| x \right.} \right),\\
E{L_i}\left( {{\partial _N}\left| x \right.} \right) = \lambda _i^{NP}pr\left( {{\pi _i}\left| x \right.} \right) + \lambda _i^{NN}pr\left( {\neg {\pi _i}\left| x \right.} \right),
\end{array}
\end{equation}
In the $i$th decision-level's granular structure $G{l_i}$, the universal ${U_i}$ can be divided into three regions with Bayesian minimum expected loss rule.
These rules are derived in the following:

(P) If $EL{_i}\left( {{\partial _P}\left| x \right.} \right) \ge EL{_i}\left( {{\partial _B}\left| x \right.} \right)$ and $EL{_i}\left( {{\partial _P}\left| x \right.} \right) \ge EL{_i}\left( {{\partial _N}\left| x \right.} \right)$, then $x \in PO{S_i}\left( {{\pi _i}} \right)$,

(B) If $EL{_i}\left( {{\partial _B}\left| x \right.} \right) \ge EL{_i}\left( {{\partial _P}\left| x \right.} \right)$ and $EL{_i}\left( {{\partial _B}\left| x \right.} \right) \ge EL{_i}\left( {{\partial _N}\left| x \right.} \right)$, then $x \in BN{D_i}\left( {{\pi _i}} \right)$,

(N) If $EL{_i}\left( {{\partial _N}\left| x \right.} \right) \ge EL{_i}\left( {{\partial _P}\left| x \right.} \right)$ and $EL{_i}\left( {{\partial _N}\left| x \right.} \right) \ge EL{_i}\left( {{\partial _B}\left| x \right.} \right)$, then $x \in NE{G_i}\left( {{\pi _i}} \right)$.
where satisfy $PO{S_i}\left( {{\pi _i}} \right)\bigcup {BN{D_i}\left( {{\pi _i}} \right)\bigcup {NE{G_i}\left( {{\pi _i}} \right)} }  = {U_i}$.

\subsection{Regret theory}\label{RT}
Loomes et al. \cite{loomes1982regret} and Bell  \cite{bell1982regret} both introduced regret theory in 1982.
This theory suggests that individuals seek to maximize their utility while minimizing potential regret from missed opportunities.
People generally seek to avoid choices that might lead to higher levels of regret.
For example, research has shown that when making decisions, people may anticipate the possibility of feeling regret once uncertainty is resolved, and therefore factor in their decision-making the desire to eliminate or reduce this potential regret.
RT models capture decision-makers' choice behavior under uncertainty by considering the impact of anticipated regret.
In general, RT models include a regret term in the utility function.
This regret is negatively correlated with the realized result and positively correlated with the best alternative result after the uncertainty is resolved.

The original model of regret theory is based on comparing the results of two alternative options.
Before making a decision, the decision-maker evaluates the results of the chosen alternative against those of the rejected ones.
If the chosen result is better, they feel happy; if not, they feel regret. The perceived utility value for the decision-maker consists of two parts: the utility value of the current result and the regret or rejoice derived from comparing it to another alternative.
Recognizing that real decision-making often involves multiple alternatives, Quiggin \cite{quiggin1994regret} extended regret theory to encompass multiple alternatives.
Let ${x_1},{x_2}, \ldots ,{x_n}$ be $n$ alternatives, where ${x_p}$ represents the $p$th alternative.
Let ${\chi _1},{\chi _2}, \ldots ,{\chi _n}$ be the results of the $n$ alternatives, where ${\chi _p}$ represents the $p$th result.
Then the perceived utility of ${x_p}$ is:
\begin{equation}\label{perceivedutility}
V\left( {{\chi _p}} \right) =  \mathfrak{u}\left( {{\chi _p}} \right) +\mathfrak{v} \left( {\mathfrak{u}\left( {{\chi _p}} \right) - \mathfrak{u}\left( {{\chi ^ + }} \right)} \right),
\end{equation}
where ${\chi ^ + } = \max \left\{ {{\chi _p}\left| {p = 1,2, \ldots ,n} \right.} \right\}$, which means the regret value $\mathfrak{v} \left( {\mathfrak{u}\left( {{\chi _p}} \right) - \mathfrak{u}\left( {{\chi ^ + }} \right)} \right)\le 0$.

In Eq.(\ref{perceivedutility}),  $\mathfrak{u}\left( \cdot \right)$ is a utility function, $\mathfrak{v}\left(\cdot  \right)$ is the regret-rejoice function.
To simulate the utility of the evaluation information, the power function \cite{zhang2016regret} $ \mathfrak{u}\left( {{\chi_p}} \right) = {\left( {{\chi_p}} \right)^\theta }$ are usually used, where $\theta$ is decision meker's risk aversion, satisfies $0 < \theta  < 1$.
It has special meaning for the regret-rejoice function.
$\mathfrak{v}\left(  \cdot  \right) = 0$ implies that neither regret nor rejoice is felt when two alternatives' results equally.
$\mathfrak{v}\left(  \cdot  \right) > 0$ indicates that $\mathfrak{v}\left(\cdot  \right)$ is strictly increasing.
Regret aversion produces a unique prediction of RT, implying that $\mathfrak{v}\left(\cdot  \right)$ is concave and $\mathfrak{v}''\left(  \cdot  \right) < 0$.
In view of these characteristics, the regret-rejoice function is usually represented as:
$\mathfrak{v} \left( {\mathfrak{u}\left( {{\chi _p}} \right) - \mathfrak{u}\left( {{\chi ^ + }} \right)} \right) = 1 - {e^{ - \delta \left( {\mathfrak{u}\left( {{\chi _p}} \right) - \mathfrak{u}\left( {{\chi ^ + }} \right)} \right)}}$, where $\delta  \ge 0$ denotes the decision-maker's regret aversion.

\section{The S3WD of DHHFLTS model}\label{3WD model}
In real life, linguistic evaluative information such as ``slightly poor", ``average", ``good", ``excellent", etc. tends to be more common, and DHHFLTS excels at collecting this type of linguistic expressions.
However, there are inefficiencies in dealing with GDM problems under double hierarchy hesitant fuzzy linguistic environment according to the traditional method. The primary challenge in integrating the S3WD model with DHHFLTS is to efficiently capture uncertain decision-making information and trap the psychological factors of decision-making given by bounded rational people.
To address this issue, this section proposes a S3WD of DHHFLTS model.
This model seamlessly integrates DHHFLTS into the S3WD framework, enabling it to effectively process and utilize linguistic information

\subsection{Conditional probability based on neighborhood theory}\label{3.1}
Existing S3WD models cannot be directly applied under DHHFLTS environment due to the complex nature of this linguistic term, which significantly impacts the selection of models and methods used in GDM.
On one hand, expert evaluations, which are based on experience and knowledge, are expressed using DHHFLTSs, but there is a lack of relevant methods to integrate DHHFLTS for calculating the attribute values of alternatives.
On the other hand, decision tables of alternatives constructed from expert evaluations in GDM typically lack pre-defined labels.
In contrast, traditional methods \cite{wang2020bwm,hu2022novel} depend on pre-defined decision attributes to estimate conditional probabilities, which becomes impractical and inefficient when dealing with numerous alternatives.
Additionally, conditional probabilities based on the decision-maker's prior knowledge can lead to significant errors.
To integrate S3WD models into GDM, it is crucial to establish a set of universal and feasible methods for defining equivalence relationships between alternatives.
Yang et al. \cite{yang2020multilevel} proposed a method for dividing general equivalence classes using neighborhood covering approach, which extends the concept from binary equivalence relations to more general binary similarity relationships.
The neighborhood relation serves as a more generalized form of equivalence classes. Their research also demonstrated that a universe-based neighborhood covering approach can enhance the model's classification performance.

To begin with, a decision table of DHHFLTS should be defined.
Then, the neighborhood binary relation among different alternatives is constructed.
Using this relationship, the neighborhood relation is introduced, and equivalence classes are formed.
Finally, a method for calculating conditional probability is presented.
It is assumed that the S3WD models in this section are all constructed when the conditional attributes are subsets $Z$ at the $i$th decision-level.

\begin{mydef}\label{mydef7}
Gievn the set $A$ of all conditional attributes and a quadruple of double hierarchy hesitant fuzzy linguistic decision table $DHHFLDT = \left( {U,Z,V,g} \right)$,
$A = \left\{ {{a_1},{a_2}, \ldots ,{a_m}} \right\}$  is a finite and nonempty set,
$Z \subseteq A$,
$U = \left\{ {{x_1},{x_2}, \ldots ,{x_n}} \right\}$ is a finite and nonempty universal set of alternatives.
$g:U \times Z \to V$ is a complete information function, $V$ denotes the domain of these attributes.
For ${x_p} \in U\left( {p = 1,2, \ldots ,n} \right)$ and  ${a_q} \in Z\left( {q = 1,2, \ldots ,m} \right)$, $g\left( {{x_p},{a_q}} \right) \in V$, $g\left( {{x_p},{a_q}} \right)$ is defined as a DHHFLE denoted by $h_{{S_O}pq}^{}$.
\end{mydef}

Equivalence classes play a crucial role in deriving conditional probabilities.
Traditionally, equivalence classes are formed by partitioning the universe based on an equivalence relation.
Neighborhood relations extend the concept of binary equivalence relations to a more general computational framework.
Research on similarity relations in DHHFLTS is essential for constructing neighborhood relations.
Thus, the binary similarity degree and neighborhood granules based on the $\kappa$-cut are introduced for DHHFLTS.

\begin{mydef}\label{mydef9}
For any alternatives ${x_p},{x_y} \in U\left( {p,y = 1,2, \ldots ,n} \right)$ in $DHHFLDT$, a universe $U$ can be partitioned by $\kappa $-cut neighborhood binary relation
${\Re _{{\aleph _Z}}} = \left\{ {\left. {\left( {{x_p},{x_y}} \right) \in U \times U} \right|} \right.$\\
$\left. {{{\rm{T}}_Z}\left( {{x_p},{x_y}} \right) \ge \kappa } \right\}$, and a family of neighborhood granule ${\kappa _{{\aleph _Z}}}\left( {{x_p}} \right)$ of ${x_p}$ in $Z$  is defined as:
\begin{equation}\label{}
{\kappa _{{\aleph _Z}}}\left( {{x_p}} \right) = \left\{ {{x_y}\left| {{x_y} \in U} \right.,{T_Z}\left( {{x_p},{x_y}} \right) \ge \kappa } \right\},
\end{equation}
\end{mydef}
where $0 \le \kappa  \le 1$, $Z$ is the subset of conditional attributes, $Z \subseteq A$, and ${T_Z}$ is the similarity degrees under conditional attributes subset $Z$.
A universe $U$ be granulated by neighborhood binary relation ${\Re _{{\aleph _Z}}}$.

\begin{property}\label{property}
For any alternatives ${x_p},{x_y} \in U$, the similarity degree ${T_Z}$ with the conditional attribute $Z$ has the following properties:
\begin{enumerate}[(1)]
\item $0 \le {T_Z}\left( {{x_p},{x_y}} \right) \le 1$,
\item ${T_Z}\left( {{x_p},{x_y}} \right) = 1$ if and only if $g\left( {{x_p},{a_q}} \right) = g\left( {{x_y},{a_q}} \right)$,
\item ${T_Z}\left( {{x_p},{x_y}} \right) = {T_Z}\left( {{x_y},{x_p}} \right)$.
\end{enumerate}
\end{property}

When evaluating decision alternatives using DHHFLTS in GDM scenarios, further investigation of similarity measure is warranted.
Similarity and distance measures are often interrelated.
A smaller distance between two alternatives typically indicates higher similarity, and vice versa.
However, the $1-d$ similarity measure may not accurately capture specific similarity concepts and may not be suitable for all scenarios.
To address this limitation, Gou et al. \cite{gou2018multiple} has developed a set of real functions that can be adapted to different scenarios.
Most distance-based methods are built on the assumption that the distance metric itself is sufficient to represent the differences between alternatives.
If the distance between any two alternatives is equal, then their similarity is also equal.
Although these real functions $\Im \left(  \bullet  \right)$ are strictly monotonically decreasing functions, there are inevitably cases where the practical meanings and arithmetic values of two DHHFLEs are not the same \cite{luo2024three}.
It is important to note that relying solely on distance values to determine the similarity may not provide an accurate representation.
The following example will illustrate this in more details.

\begin{example}\label{Example3}
For the alternatives ${x_1},{x_2} \in U$ denoted by ${{h_{{S_O}}}\left( {{x_1}} \right)}$ and ${{h_{{S_O}}}\left( {{x_2}} \right)}$ , the similarity measure \cite{gou2018multiple} between them could be calculated by the following equation:
\begin{equation}\label{}
T\left( {{x_1},{x_2}} \right) = \frac{{\Im \left( {d\left( {{x_1},{x_2}} \right)} \right) - \Im \left( 1 \right)}}{{\Im \left( 0 \right) - \Im \left( 1 \right)}}.
\end{equation}
where the Euclidean distance measure is defined as:
\begin{equation}\label{}
d\left( {{x_1},{x_2}} \right) = {\left( {\frac{1}{L}\sum\limits_l^L {{{\left( {\left| {F'\left( {{h_{{S_O}}}\left( {{x_1}} \right)} \right) - F'\left( {{h_{{S_O}}}\left( {{x_2}} \right)} \right)} \right|} \right)}^2}} } \right)^{{1 \mathord{\left/
 {\vphantom {1 2}} \right.
 \kern-\nulldelimiterspace} 2}}}.
\end{equation}
Suppose the real function is $\Im \left( \upsilon  \right) = 1 - d\left( \upsilon  \right)$ and a DHHFLE ${h_{{S_O}}}$ only has a DHLT ${s_{{\phi _l}\left\langle {{o_{{\varphi _l}}}} \right\rangle }}$, namely, ${h_{{S_O}}}\left( x \right) = \left\{ {{s_{{\phi _l}\left\langle {{o_{{\varphi _l}}}} \right\rangle }}} \right\}$, then the transformation function $F$ reduces to $F'$, $F'\left( {{h_{{S_O}}}\left( x \right)} \right) = F'\left( {{s_{{\phi _l}\left\langle {{o_{{\varphi _l}}}} \right\rangle }}} \right) = f\left( {{\phi _l},{\varphi _l}} \right)$.
Assuming ${{h_{{S_O}}}\left( {{x_1}} \right)} = \left\{ {{s_{1\left\langle {{o_{ - 3}}} \right\rangle }},{s_{2\left\langle {{o_{ - 3}}} \right\rangle }}} \right\}$ and ${{h_{{S_O}}}\left( {{x_2}} \right)}  = \left\{ {{s_{ - 1\left\langle {{o_3}} \right\rangle }},{s_{0\left\langle {{o_3}} \right\rangle }}} \right\}$, it could get $d\left( {{x_1},{x_2}} \right) = {\left( {\frac{{{{\left( {0.5 - 0.5} \right)}^2} + {{\left( {0.67 - 0.67} \right)}^2}}}{2}} \right)^{{1 \mathord{\left/
 {\vphantom {1 2}} \right.
 \kern-\nulldelimiterspace} 2}}} = 0$ and $T\left( {{x_1},{x_2}} \right) = 1$.
Similarly, it could get $T\left( {{x_3},{x_4}} \right) = 1$ with ${h_{{S_O}3}} = \left\{ {{s_{0\left\langle {{o_3}} \right\rangle }},{s_{1\left\langle {{o_0}} \right\rangle }},{s_{2\left\langle {{o_{ - 3}}} \right\rangle }}} \right\}$
and ${h_{{S_O}4}} = \left\{ {{s_{1\left\langle {{o_0}} \right\rangle }},{s_{2\left\langle {{o_{ - 3}}} \right\rangle }}} \right\}$ using Eq.(\ref{eq1}) and (\ref{eq3}).
It can be seen from these two calculation results that the distance result have a great influence on the similarity result.
But ${{h_{{S_O}}}\left( {{x_1}} \right)}$ and ${{h_{{S_O}}}\left( {{x_2}} \right)}$ have different evaluation forms, indicating different alternative preferences.
The same goes for ${{h_{{S_O}}}\left( {{x_3}} \right)}$ and ${{h_{{S_O}}}\left( {{x_4}} \right)}$ .
\end{example}

As highlighted earlier, a similarity measurement will be explored to address this issue.
On one hand, considering the different situations that arise from varying practical meanings and arithmetic values, as well as the anchoring effect observed in human thinking, the superior gradus \cite{luo2024three} is designed to address such scenarios.
On the other hand, the introduction of the Gaussian Kernel function enables the nonlinear transformation of raw distance measurements into a more interpretable similarity space.
Unlike linear transformations, the Gaussian Kernel doesn't simply convert distance to similarity in a straight line.
Instead, it utilizes an exponential function to map the distance between alternatives onto a similarity value between 0 and 1.
This approach proves particularly valuable in situations where two alternatives are highly similar but not identical.
In conclusion, the use of superior gradus-based Gaussian Kernel function will have a better advantage in dealing with complex similarity relationships and nonlinear data distributions in GDM scenarios evaluated using DHHFLTS than distance-based similarity methods.

For any alternatives ${x_p},{x_y} \in U\left( {p,y = 1,2, \ldots ,n} \right)$ in $DHHFLDT$, the similarity degree-based gaussian kernel function \cite{hu2010gaussian} of the alternative ${x_p}$ and ${x_y}$ on the conditional attributes subset $Z$, $Z \subseteq A$ can be expressed as:
\begin{equation}\label{originalgaussian}
{{T_Z}\left( {{x_p},{x_y}} \right) = {{\rm{K}}_Z}\left( {{x_p},{x_y}} \right) = \exp \left( { - \frac{{\left\| {{x_p} - {x_y}} \right\|_Z^2}}{{2{\sigma ^2}}}} \right)},
\end{equation}
where $\sigma $ is the width parameter of Gaussian Kernel function.

To better capture the similarity between alternatives, a combination of the superior gradus \cite{luo2024three} and the Euclidean norm ${\left\| {{x_p} - {x_y}} \right\|_Z^2}$ is used.
As a result, Eq.(\ref{originalgaussian}) is modified.

\begin{theorem}\label{theorem2}
Given any two alternatives ${x_p},{x_y} \in U$ in $DHHFLDT$ denoted by DHHFLEs ${{h_{{S_O}}}\left( {{x_p}} \right)}$ and ${{h_{{S_O}}}\left( {{x_y}} \right)}$, respectively, the superior gradus-based Gaussian Kernel function of the alternative ${x_p}$ and ${x_y}$ on the conditional attributes subset $Z$ ($Z \subseteq A$) can be expressed as:
\begin{equation}\label{simlilaritydegeree}
{{\rm{K}}_Z}\left( {{x_p},{x_y}} \right) = {\rm{exp}}\left( { - \frac{{{{\sum\nolimits_{q = 1}^{\left| Z \right|} {\left( {SG\left( {{h_{{S_O}}}\left( {{x_p}} \right)} \right) - SG\left( {{h_{{S_O}}}\left( {{x_y}} \right)} \right)} \right)} }^2}}}{{2{\sigma ^2}}}} \right),
\end{equation}
where $SG\left( {{h_{{S_O}}}\left( x \right)} \right) = \frac{1}{L}\sum\limits_{l = 1}^L {sg\left( {{s_{{\phi _l}\left\langle {{o_{{\varphi _l}}}} \right\rangle }}} \right)}$, $sg\left( {{s_{{\phi _l}\left\langle {{o_{{\varphi _l}}}} \right\rangle }}} \right) = \frac{{\left( {{e^\alpha } + \beta } \right) - 1}}{{e - 1}}$, $\alpha  = \left( {\frac{{{\phi _l}}}{{2\tau }} + \frac{1}{2}} \right)$, and $\beta  = \frac{{{\varphi _l}}}{{2\varsigma \tau }}$.
Obviously, Eq.(\ref{simlilaritydegeree}) also satisfies Property \ref{property}.
\end{theorem}

\begin{theorem}\label{theorem}
Given a $DHHFLDT$, for  ${\sigma _1} \le {\sigma _2}$ and any ${x_p},{x_y} \in U$, it has ${{\rm{K}}_Z}{\left( {{x_p},{x_y}} \right)_{^{{\sigma _1}}}} \le {{\rm{K}}_Z}{\left( {{x_p},{x_y}} \right)_{^{{\sigma _2}}}}$.
\end{theorem}

\begin{remark}
Based on the above, Theorem \ref{theorem} is not difficult to prove.
Previous research \cite{yang2020multilevel} has shown that as decision information increases from coarse to fine, adjusting the parameter $\sigma$ according to a certain rule can improve the accuracy of the decision results.
When there is limited decision-making information, only a small amount of information is used for comparative analysis, leading to high uncertainty in the results.
As more decision-making information becomes available, the basis for comparison becomes more accurate, reducing the uncertainty of the decision results.
The flexibility of the similarity measure can be controlled by adjusting $\sigma$: a larger $\sigma$ results in lower differentiation between alternatives, while a smaller $\sigma$ leads to higher differentiation, thus providing more certain decision information.
\end{remark}

The S3WD framework relies on classifying alternatives based on their similarity.
A crucial step in this process involves constructing a specific type of binary relation.
A common representation is to use the $\kappa$-cut neighborhood binary relation ${\Re _{{\aleph _Z}}}$ to capture the similarity relationships between alternatives within the context of the decision problem \cite{hu2008neighborhood}. Given a $\kappa$-cut neighborhood binary relation ${\Re _{{\aleph _Z}}}$$\left( {\kappa  \in \left[ {0,1} \right]} \right)$, the neighborhood relation matrix ${R_{{\aleph _Z}}} = {\left( {{{\rm{r}}_{py}}} \right)_{n \times n}}$ for any ${x_p},{x_y} \in U$ can be defined by:
\begin{equation}\label{cutbinaryrelation}
{r_{py}} = \left\{ {\begin{array}{*{20}{c}}
{1,\left( {{x_p},{x_y}} \right) \in {\Re _{{\aleph _Z}}}}\\
{0,otherwise}
\end{array}} \right.,
\end{equation}
where $\kappa$-cut neighborhood binary relation ${\Re _{{\aleph _Z}}}$ satisfies reflexivity and symmetry.
Then, a family of neighborhood granule ${\kappa _{{\aleph _Z}}}\left( {{x_p}} \right)$ for different alternatives with varying $\kappa $ thresholds can be obtained.
By dynamically conditioning the attributes using specific rules, sequences with varying neighborhood granularity are created.
These sequences are characterized by subsets of conditional attributes.

In traditional S3WD models, the calculation of conditional probabilities relies on known prior probabilities of alternatives, typically derived from decision attributes.
However, integrating DHHFLTS into the S3WD model presents a significant challenge: the lack of prior knowledge about alternatives makes it impossible to compute these prior probabilities directly.
Although studies such as \cite{wang2020bwm,hu2022novel} propose a modified S3WD approach to address GDM problems across multi-level granularity, their reliance on decision attributes in the information system limits their applicability.
Consequently, a novel approach for calculating conditional probabilities based on the outranking relation is introduced \cite{ye2020novel}.
This method assesses the relative advantage of each alternative compared to others, aiming to evaluate the likelihood of different alternatives achieving specific results through comparative advantage relationships.
It gives a membership function ${\pi _Z}\left( {{x_p}} \right) = \sum\limits_{q = 1}^n {{w_q}}  \oplus g\left( {{x_p},{a_q}} \right)$.
And two complementary state concepts are defined based on this function, where the concept of ``good state" ${\pi _Z}$ is the total membership of the alternative $x_p$, denoted as ${\pi _Z} = \frac{{{\pi _Z}\left( {{x_1}} \right)}}{{{x_1}}} + \frac{{{\pi _Z}\left( {{x_2}} \right)}}{{{x_2}}} +  \ldots  + \frac{{{\pi _Z}\left( {{x_n}} \right)}}{{{x_n}}}$, and the other concept ``bad state" $\neg {\pi _Z}$ is the complement of concept ${\pi _Z}$.

\begin{mydef}\label{conditionalprab}
For any alternative ${x_p}$ belong to the ``good state" ${\pi _Z}$ in neighborhood granules ${\kappa _{{\aleph _Z}}}\left( {{x_p}} \right)$, the conditional probability for the alternative ${x_p}$ is derived as follows:
\begin{equation}\label{outrankingrelationconditionalprobability}
pr\left( {{\pi _Z}\left| {{\kappa _{{\aleph _Z}}}\left( {{x_p}} \right)} \right.} \right) = \frac{{\sum\limits_{{x_p} \in {\kappa _{{\aleph _Z}}}\left( {{x_p}} \right)}^{} {{\pi _Z}\left( {{x_p}} \right)} }}{{\left| {{\kappa _{{\aleph _Z}}}\left( {{x_p}} \right)} \right|}},
\end{equation}
where $\left| {{\kappa _{{\aleph _Z}}}\left( {{x_p}} \right)} \right|$ represents the cardinality of the neighborhood granule ${\kappa _{{\aleph _Z}}}\left( {{x_p}} \right)$.
\end{mydef}

An example is used to illustrate the conditional probability based on the neighborhood granule ${\kappa _{{\aleph _Z}}}\left( {{x_p}} \right)$.
\begin{example}
If ${\kappa _{{\aleph _Z}}}\left( {{x_1}} \right) = \left\{ {{x_1},{x_3},{x_5}} \right\}$,
the conditional attribute $Z = \left\{ {{a_1},{a_2},{a_3} ,{a_4}} \right\}$ weights are denoted by the vector $w = \left\{ {{w_1},{w_2},{w_3} ,{w_4}} \right\}$, $g\left( {{x_p},{a_q}} \right)$ is a DHHFLE $h_{{S_O}pq}^{}\left( {p = 1,3,5;} \right.$\\$\left. {q = 1,2,3,4} \right)$. Then,

$\begin{array}{l}
{\pi _Z}\left( {{x_1}} \right) = {w_1}  {h_{{S_O}}}_{11} \oplus {w_2} {h_{{S_O}}}_{12} \oplus {w_3} {h_{{S_O}}}_{13} \oplus {w_4} {h_{{S_O}}}_{14}\\
{\pi _Z}\left( {{x_3}} \right) = {w_1}  {h_{{S_O}}}_{31} \oplus {w_2} {h_{{S_O}}}_{32} \oplus {w_3}  {h_{{S_O}}}_{33} \oplus {w_4} {h_{{S_O}}}_{34}\\
{\pi _Z}\left( {{x_5}} \right) = {w_1}  {h_{{S_O}}}_{51} \oplus {w_2} {h_{{S_O}}}_{52} \oplus {w_3} {h_{{S_O}}}_{53} \oplus {w_4} {h_{{S_O}}}_{54}
\end{array}$

According to Definition \ref{conditionalprab}, it could get:
$pr\left( {{\pi _Z}\left| {{\kappa _{{\aleph _Z}}}\left( {{x_1}} \right)} \right.} \right) = \frac{{{\pi _Z}\left( {{x_1}} \right) + {\pi _Z}\left( {{x_3}} \right) + {\pi _Z}\left( {{x_5}} \right)}}{3}$.
\end{example}

\subsection{The relative perceived utility function via RT}\label{3.2}
The development of the loss function in the S3WD model has been a critical area of research, focusing on how to more accurately reflect losses in decision-making processes.
Initially, Yao \cite{yao2010three} proposed loss function values based on individual experience, which however, were fixed and did not accurately mirror the realistic loss scenarios.
The concept of ``relative"  has been progressively introduced to more accurately characterize the semantics of loss.
Jia and Liu \cite{jia2019novel} explored a relative loss function for different alternatives, a concept further developed by Liang et al. \cite{liang2019heterogeneous}, who constructed both relative loss and benefit functions.
Lei et al. \cite{lei2020multigranulation} extended these concepts with PT under hesitant fuzzy linguistic environment, acknowledging that while relative loss functions approach reality, decision-makers often prefer knowledge of their gains (``what I get").

Traditional relative loss functions either rely on the subjective assignment of losses incurred from actions by experts or are based on the principle that the greater the value generated by an action, the smaller the loss, according to the assumption of complete rationality.
This is particularly true in fuzzy environments of multi-attribute decision-making.
However, evaluation information, especially in the context of DHHFLTS, is often provided subjectively by individuals, thereby increasing the uncertainty of decision-making. Depicting human psychological preferences solely based on a linear relationship is somewhat simplistic
This subjectivity challenges the assumption of complete rationality which is required to convert qualitative evaluations into quantitative expressions.
Although relative loss functions bring us closer to reality, they do not completely solve the psychological issues in decision-making, particularly the emotions of regret and delight that significantly impact human decisions.
Decision-makers' evaluations of alternatives are often influenced by anchoring, showing a preference for current alternatives, which is undoubtedly closely related to the concept of relativity.
RT can analyze the consequences of irrational behavior in the decision-making process.
Through mathematical representation, it is possible to better understand people's decision-making behavior under uncertain conditions, especially when potential regret or rejoice becomes a significant influencing factor.

\begin{figure}[h]
    \centering
    \begin{minipage}[b]{0.35\textwidth}
        \centering
        \includegraphics[width=\textwidth]{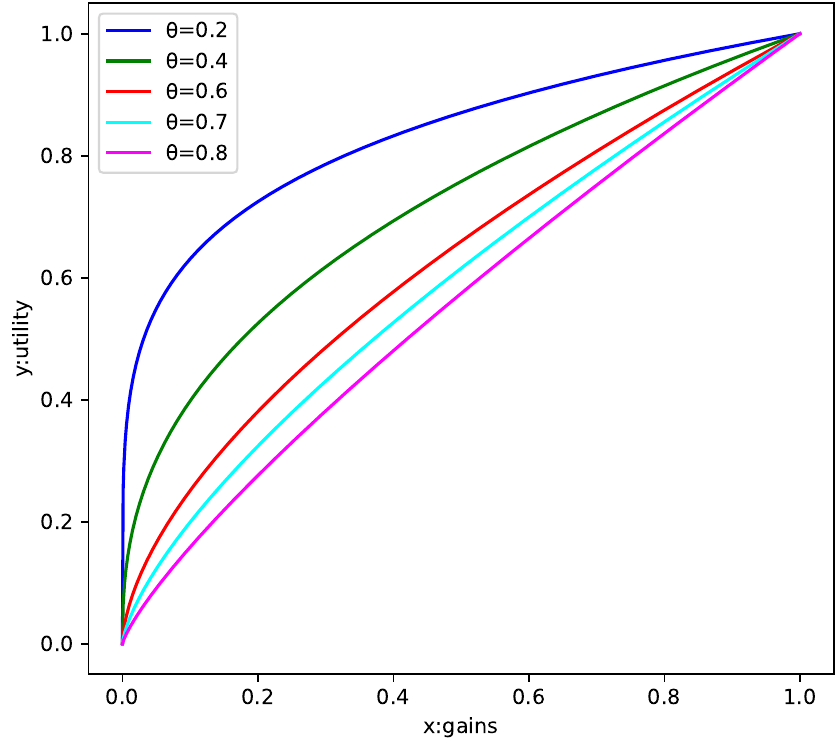}
    \end{minipage}
    \hspace{0.02\textwidth} %
    \begin{minipage}[b]{0.355\textwidth}
        \centering
        \includegraphics[width=\textwidth]{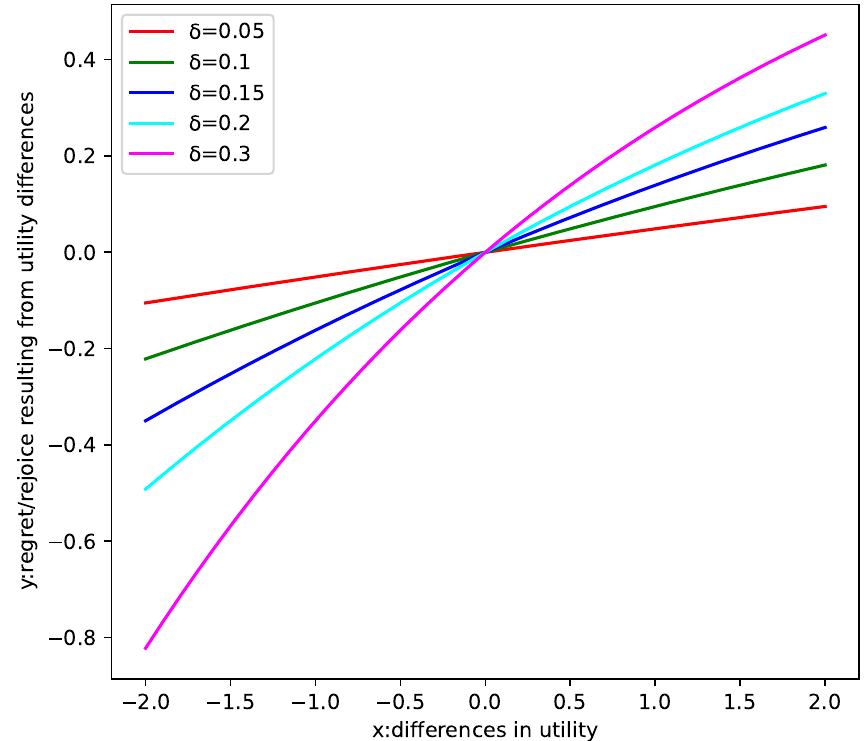}
    \end{minipage}
    \caption{\textbf{.} Utility function $ \mathfrak{u}$ and regret/rejoice function $\mathfrak{v}$.}
\label{regret}
\end{figure}

Qualitative evaluation is essentially an intuitive method of characterizing gains.
These evaluations incorporate the decision-maker's perceived sensitivity to risk, acknowledging their limited rationality.
In reality, this sensitivity manifests as a decline in the rate of utility growth as returns increase, shown in Fig. \ref{regret}  below.
The relative gain function serves as the foundation for constructing the relative perceived utility function in RT, which further enhances the decision-making process by explicitly accounting for the decision-makers' bounded rationality and sensitivity to risk and regret.

Given a $DHHFLDT$, the evaluation value of the alternative ${x_p}$ with respect to the conditional attribute ${a_q}$ is $g\left( {{x_p},{a_q}} \right)$, denoted as $h_{{S_O}pq}^{}$.
${\pi _Z}$ and $\neg {\pi _Z}$ represent whether the alternative ${x_p}$ is in good state or not under $Z \subseteq A$.
There are three actions: ${\partial _P}$, ${\partial _B}$, and ${\partial _N}$, which denote the actions of dividing ${x_p}$ into acceptance region, uncertainty region and rejection region, respectively.
The unit relative gain function is the result of ${x_p}$ under a conditional attribute $a_q$$\left( {{a_q} \in Z} \right)$, ${\Lambda _{{a_q}}} = \left\{ {\chi _{pq}^{PP},\chi _{pq}^{BP},\chi _{pq}^{NP},\chi _{pq}^{PN},\chi _{pq}^{BN},\chi _{pq}^{NN}} \right\}$ includes six kinds of relative gains.
These gains are produced by taking different actions in different states.
When ${x_p}$ with the conditional attribute ${a_q}$ belongs to ${\pi _Z}$, $\chi _{pq}^{PP},\chi _{pq}^{BP},\chi _{pq}^{NP}$ represent the relative gain from taking actions ${\partial _P}$, ${\partial _B}$, and ${\partial _N}$, respectively.
Conversely, $\chi _{pq}^{PN},\chi _{pq}^{BN},\chi _{pq}^{NN}$ represent the relative gains from taking the same actions when ${x_p}$ with the attribute ${a_q}$ belongs to $\neg {\pi _Z}$.
In Table \ref{relativegainfunction}, this relative gain function serves as a basic unit to clearly present the relative gain function of an alternative with a certain conditional attribute.
The number of units is related to the product of the number of alternatives and the number of conditional attributes.
\begin{table}[htbp]
\centering
\renewcommand{\arraystretch}{1.33}%
\setlength{\abovecaptionskip}{1.5pt}%
\captionsetup{justification=centering, singlelinecheck=false, labelsep=quad}
\caption{The unit of relative gain function.}
\label{relativegainfunction}
\setlength{\tabcolsep}{5.2mm}{
\begin{tabular}{{cc|cc}}
\hline
\multicolumn{2}{c|}{}                              & ${\pi _Z}$   & $\neg {\pi _Z}$ \\ \hline
\multicolumn{1}{c|}{\multirow{3}{*}{$x$}} & ${{\partial _P}}$ & $\chi _{pq}^{PP}$    & $\chi _{pq}^{PN}$ \\ \cline{2-2}
\multicolumn{1}{c|}{}                        & ${{\partial _B}}$ & ${\chi _{pq}^{BP}}$ & ${\chi _{pq}^{BN}}$  \\ \cline{2-2}
\multicolumn{1}{c|}{}                        & ${{\partial _N}}$ & ${\chi _{pq}^{NP}}$ & ${\chi _{pq}^{NN}}$ \\ \hline
\end{tabular}
}
\end{table}

$\chi _{pq}^{ \circ  \bullet }$ $\left( { \circ  = P,B,N; \bullet  = P,N} \right)$ represents the relative gain functions with different conditional attributes.
As the evaluation information for decision-makers, it applies the form of DHHFLE.
Unlike the concept of a relative loss function, a relative gain function is based on the expression of the gain that can be visualized for that alternative taken by a certain action.
The content of two states and three actions in this basic unit can produce six behavioral results.
For the action result $\chi _{pq}^{PP}$, it is the full gains obtained by alternative ${x_p}$ in one situation where the conditional attribute is ${a_q}$ and the state is ${\pi _Z}$.
For the delayed generation behavior $\chi _{pq}^{BP}$, it is part of the gains from alternative ${x_p}$ in the above situation.
Refusal to make a behavior will result in no gains, which means that $\chi _{pq}^{NP}$ has no value.

In another situation where the conditional attribute is ${a_q}$ and the state is $\neg {\pi _Z}$, the gains $\chi _{pq}^{PN}$ implies that taking action ``without the right place, the right time, and the right people" would be meaningless, i.e., no gains.
On the contrary, if such behavior is not made, gains can be preserved instead.
The value of preserved gains can also be understood in terms of the concept of opportunity cost in economics.
This means that taking action in the state ${\pi _Z}$ yields a gain of $\chi _{pq}^{PP}$, which corresponds to the potential loss of not taking action in state ${\pi _Z}$.
In the state $\neg {\pi _Z}$, choosing not to take action results in potential gains $\chi _{pq}^{NN}$, which means giving up the gains $\chi _{pq}^{PP}$.
The resulting potential gains are complementary to the gains $\chi _{pq}^{PP}$.
Additionally, delaying the action $\chi _{pq}^{BN}$ represents part of the gains $\chi _{pq}^{NN}$.
The connection between these concepts can be expressed as follows:
$\chi _{pq}^{BP} = \eta  \chi _{pq}^{PP}$, $\chi _{pq}^{NN} = {\left( {\chi _{pq}^{PP}} \right)^C}$, and $\chi _{pq}^{BN} = \eta \chi _{pq}^{NN}$.

Since some DHHFLEs often have a different number of DHLTs.
There are two notable points in the computation: the first is to characterize the subtle differences between different DHHFLEs, and the second is to compare DHHFLEs when obtaining regret and rejoice values.
For this purpose, a superior gradus \cite{luo2024three} is introduced as a preprocessing step for the relative gain function.
Especially, superior gradus can solve the problem of overlapping transform values in DHHFLE.
To simplify the calculation, the unit relative gain function ${\Lambda _{{a_q}}} = \left\{ {\chi _{pq}^{ \circ  \bullet }} \right\}$ can be further expressed as a simplified unit relative gain function ${\overline \Lambda  _{{a_q}}} = \left\{ {b_{pq}^{ \circ  \bullet }} \right\}$, where $b_{pq}^{ \circ  \bullet } = SG\left( {\chi _{pq}^{\circ  \bullet }} \right)$.

\begin{mydef}\label{}
Given a simplified unit relative gain function ${\overline \Lambda  _{{a_q}}} = \left\{ {b_{pq}^{ \circ  \bullet }} \right\}$, the utility of the evaluation information and the relative perceived utility are proposed as respectively:
\begin{equation}\label{SGdirectutility}
\mathfrak{u} \left( {{b_{pq}^{ \circ  \bullet }}} \right)={\left( { {b_{pq}^{ \circ  \bullet }}} \right)^\theta },
\end{equation}
\begin{equation}\label{relative perceived utility}
V_{pq}^{ \circ  \bullet } = V\left( {{b _{pq}^{ \circ  \bullet }} } \right) = \mathfrak{u}\left( { {b _{pq}^{ \circ  \bullet }}} \right) + \mathfrak{v}\left( {\mathfrak{u}\left( {{b _{pq}^{ \circ  \bullet }}} \right) - \mathfrak{u}{^ + }\left( { {b _{pq}^{ \circ  \bullet }} } \right)} \right).
\end{equation}
\end{mydef}

Currently, a computational model of the perceived behavior of alternatives in qualitative decision making under a given conditional attribute is obtained by detailing the above.
At the core of the model are six action results, each represented by an RT model.
Details regarding the specific RT model and the unit relative perceived utility function are presented in Table \ref{relativeperceivedfunction}.
The total number of units in the model is calculated by multiplying the number of alternatives under consideration by the number of conditional attributes associated with each alternative.

\begin{table}[htbp]
\centering
\renewcommand{\arraystretch}{1.33}%
\setlength{\abovecaptionskip}{1.5pt}%
\captionsetup{justification=centering, singlelinecheck=false, labelsep=quad}
\caption{The unit of relative perceived utility function.}
\label{relativeperceivedfunction}
\setlength{\tabcolsep}{7mm}{
\begin{tabular}{{cc|cc}}
\hline
\multicolumn{2}{c|}{}                              & ${\pi _Z}$   & $\neg {\pi _Z}$ \\ \hline
\multicolumn{1}{c|}{\multirow{3}{*}{$x$}} & ${{\partial _P}}$ & $V _{pq}^{PP}$    & $V_{pq}^{PN}$ \\ \cline{2-2}
\multicolumn{1}{c|}{}                        & ${{\partial _B}}$ & ${V _{pq}^{BP}}$ & ${V_{pq}^{BN}}$  \\ \cline{2-2}
\multicolumn{1}{c|}{}                        & ${{\partial _N}}$ & ${V _{pq}^{NP}}$ & ${V _{pq}^{NN}}$ \\ \hline
\end{tabular}
}
\end{table}

For the subset of the conditional attributes $Z \subseteq A$, the relative comprehensive perceived utility $\mathbb{V}_p^{ \circ  \bullet }$ of the alternative ${x_p}$ is computed as below:
\begin{equation}\label{comprehensiveperceivedutility}
\mathbb{V}_p^{ \circ  \bullet } = \sum\limits_{m = 1}^q {{w_Z}} V_{pq}^{ \circ  \bullet },
\end{equation}
where ${w_Z}$ is the weight corresponding to the condition attribute in $Z$.
The number of $q$ are related to the elements contained in $Z$, satisfy $p = 1,2, \ldots ,n;q = 1,2, \ldots ,m$ .
The subset of conditional attribute weight ${w_Z}$ satisfies $0 \le {w_{Z}} \le 1$ and $\sum\limits_{q = 1}^m {{w_Z}}  = 1$.

These two subsections address two crucial aspects related to S3WD of DHHFLTS models.
They serve as the foundation of this work's endeavor to investigate the sequential process.
The extraction and aggregation model of these information units and multi-level information fusion sequential process will be demonstrated in the next section.

\section{The S3W-GDM for DHHFLTS}\label{processing}
From the perspective of granular computing, S3WD is an effective method for dealing with complex and dynamic uncertainty problems.
To achieve an efficient decision-making process, this section constructs a multi-level S3WD for GDM (S3W-GDM) based on the ``trisect-and-conquer" strategy.
It implements two issues at each level, ``Triading" before ``Optimizing", which involves dividing the alternatives into three parts and then taking a matching action for each stage.

\subsection{Statement of the problem}\label{Statement of the problem}
The GDM problem of DHHFLTS can be explained as follows: there are $n$ alternatives in the finite set $U = \left\{ {{x_1},{x_2}, \ldots ,{x_n}} \right\}$, $m$ conditional attributes in the finite set $A = \left\{ {{a_1},{a_2}, \ldots ,{a_m}} \right\}$ and $e$ experts in the finite set $E = \left\{ {{E_1},{E_2}, \ldots ,{E_e}} \right\}$.
The conditional attribute weights are denoted by the vector $w = \left\{ {{w_1},{w_2}, \ldots ,{w_m}} \right\}$, where $0 \le {w_q} \le 1$, $\sum\limits_{q = 1}^m {{w_q}}  = 1$, and $w_{q}\in w$.
It is typical to discuss the type of attributes, namely cost or benefit types.
Since the semantic analysis of attributes is crucial for effectively resolving complex decision issues \cite{ye2022variable}.
Failing to distinguish between attributes in decision analysis can easily lead to incorrect decisions.
In the context of diagnosing a disease, there are two crucial evaluation attributes: the severity of symptoms and the side effects of treatment.
The severity of symptoms is a health attribute that indicates the impact of the disease on the patient, where a higher severity level requires urgent treatment.
On the other hand, the side effects of treatment represent a treatment risk attribute, where fewer side effects mean the treatment is more beneficial for the patient.
Therefore, the types of evaluation attributes in the evaluation process need to be defined before the experts give their evaluations.
When multiple experts are involved, their opinions are usually given different weights depending on their level of expertise and credibility.
The weight of different experts are denoted by ${w_E} = \left\{ {w_E^1,w_E^2, \ldots ,w_E^{\rm{e}}} \right\}$, where $0 \le w_E^j \le 1$, $\sum\limits_{j = 1}^e {w_E^j}  = 1\left( {j = 1,2, \ldots ,e} \right)$, and $w_E^j\in w_E$.
Each expert provides their own evaluation for all the alternatives based on different conditional attributes.
These evaluations constitute with personal preferences that includes the number of $n \times m$ DHHFLEs $h_{{S_O}pq}^j$, as shown in Table \ref{DHHFLDTofj-expert}.
The $\left( {DHHFLDT} \right)_{}^j$ can also be denoted as matrices ${H^j} = \left\{ {h_{{S_O}pq}^j} \right\}$, which denotes the evaluation of the alternative ${x_p}$ with the conditional attribute ${a_q}$ given by the $jth$ expert.
\begin{table}[htbp]
\centering
\renewcommand{\arraystretch}{1.4}%
\setlength{\abovecaptionskip}{1.5pt}%
\captionsetup{justification=centering, singlelinecheck=false, labelsep=quad}
\caption{The $DHHFLDT$ of the $jth$ expert.}
\label{DHHFLDTofj-expert}
\setlength{\tabcolsep}{3mm}{
\begin{tabular}{ccccc}
\hline
           & ${a_1}$           & ${a_2}$           & $ \cdots $ & ${a_m}$           \\ \hline
${x_1}$    & $h_{{S_{O11}}}^j$ & $h_{{S_{O12}}}^j$ & $ \cdots $ & $h_{{S_{O1m}}}^j$ \\
${x_2}$    & $h_{{S_{O21}}}^j$ & $h_{{S_{O22}}}^j$ & $ \cdots $ & $h_{{S_{O2m}}}^j$ \\
$ \vdots $ & $ \vdots $        & $ \vdots $        & $ \vdots $ & $ \vdots $        \\
${x_n}$    & $h_{{S_{On1}}}^j$ & $h_{{S_{On2}}}^j$ & $ \cdots $ & $h_{{S_{Onm}}}^j$ \\ \hline
\end{tabular}
}
\end{table}

This section proposes a novel approach to extract and aggregate expert evaluation information in GDM.
The granular computing is used to improve the traditional GDM fusion method.
The novel method involves sequentially considering the importance of conditional attributes and establishing multi-level granularity of evaluation information.
This novel method builds on the core concepts presented in Section \ref{3WD model} and is the basis of this study.

\subsection{Multi-level decision table granularity for fusion}\label{multi-level}
Many decision-making scenarios benefit from starting with coarse-grained information.
This approach proves valuable in initial evaluations, as seen in tasks like selecting contestants, screening resumes, or filtering emails.
Coarse-grained information allows for a quick and efficient assessment of a large number of alternatives.
However, after the initial selection, further refinement of the evaluation process is often required.
This is where fine-grained information becomes critical.
By continuously refining the decision attributes through successive levels of analysis, decision-makers can make more detailed and precise judgments.
This iterative process ultimately improves the overall effectiveness of the decision-making process.
While more detailed information is essential for final decisions, coarse-grained information plays a vital role in enhancing decision-making efficiency.
After all, decision-making incurs costs related to information collection and the actual situation.
Therefore, a multi-level S3WD process offers a compelling approach by progressively analyzing decision-making information, starting from coarse-grained to fine-grained levels. This iterative strategy enables a balance between efficiency and accuracy, leading to better decision results.

\begin{figure}[htbp]
  \centering
  \includegraphics[width=13cm]{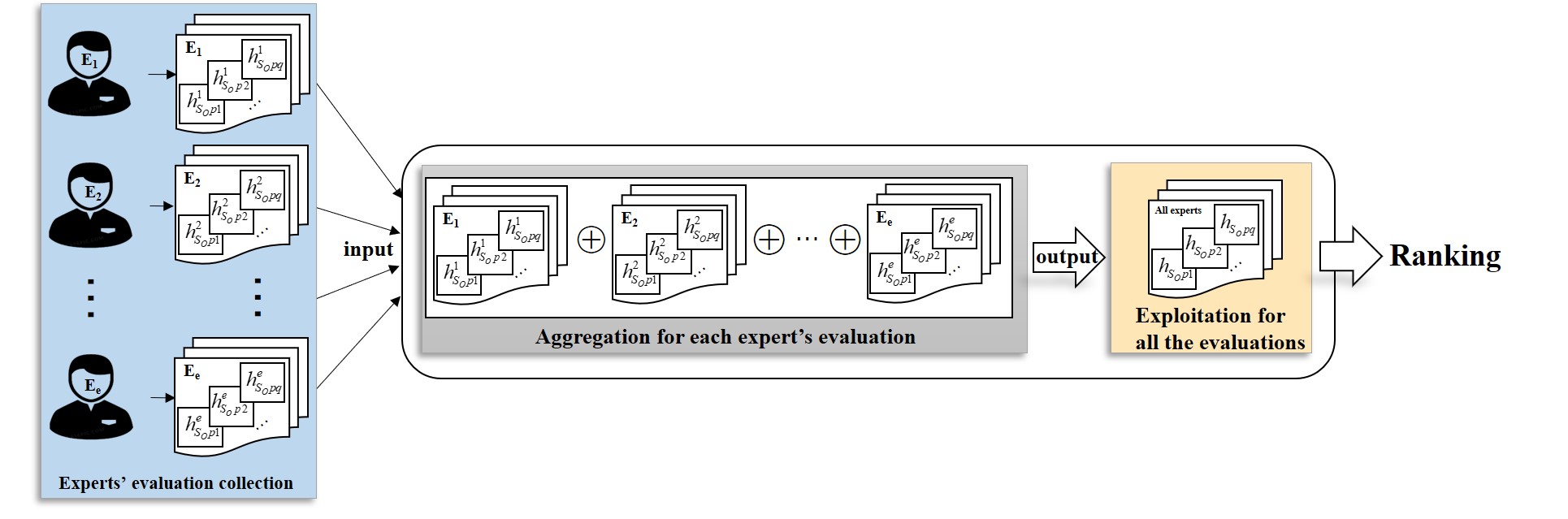}
  \caption{\textbf{.} The traditional framework of fusing the evaluation information.}
  \label{oldfusionmodel}
\end{figure}

\begin{figure}[htbp]
  \centering
  \includegraphics[width=13cm]{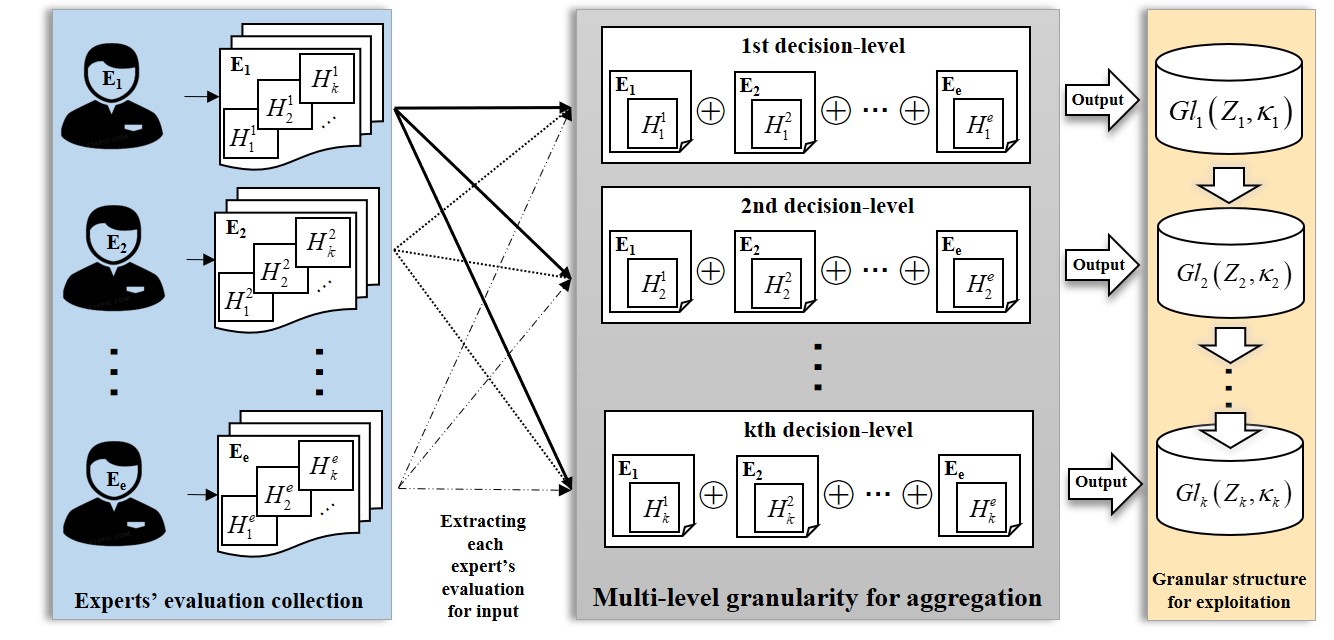}
  \caption{\textbf{.} The novel framework for fusing the evaluation information.}
  \label{newfusionmodel}
\end{figure}

Information fusion models, particularly those dealing with vague and uncertain evaluations, often rely heavily on aggregation operators.
Extensive research has been conducted in this area, as evidenced by these works \cite{xu2015hesitant,liu2018some,yager1988ordered}.
Regardless of the specific type of operator chosen, the core method remains consistent.
Experts begin by evaluating each alternative based on various attributes.
An aggregation operator is then applied to integrate these individual evaluations for each alternative across all attributes, resulting in a set of integrated evaluations.
Finally, an exploitation method is used to rank the alternatives based on the integrated evaluations, ultimately selecting the optimal one (as illustrated in Fig. \ref{oldfusionmodel}).
Constructing a multi-level expert information aggregation granularity method can effectively improve the efficiency of alternative categorization and ranking for GDM problem.
The novel fusion framework are shown in Fig. \ref{newfusionmodel}.
To facilitate a coarse-to-fine presentation of input information, the expert evaluation information table needs to be defined.

\begin{mydef}\label{extraction function}
Let $\left( {DHHFLDT} \right)_{}^j = \left( {U,A,{V^j},{g^j}} \right)\left( {j = 1,2, \ldots ,e} \right)$ be the decision table denoted as ${H^j}$, which means the evaluation of the alternative ${x_p}$ with the conditional attribute ${a_q}$ given by the $jth$ expert.
There exists an evaluation information extraction function $\mathfrak{d}$ satisfies $\mathfrak{d}\left( {{H^j}} \right) = \left\{ {H_1^j,H_2^j, \ldots ,H_i^j, \ldots ,H_k^j} \right\}$,
where ${H_i^j}$ is the matrix representation of the decision extraction table $\left( {DHHFLDT} \right)_i^j = \left( {{U_i},{Z_i},V_i^j,g_i^j} \right)$ for the $jth$ expert under ${Z_i}$, where ${V_i^j}$ denotes the domain of the conditional attributes, and ${{g_i^j}}:U_i^{} \times Z_i^{} \to {V_i^j}$ is the complete information function.
${Z_i}$ be the set of $z$ attributes whose conditional attributes are ranked in the top $i$ in terms of importance.
${Z_i}$ satisfy ${Z_1} \subseteq {Z_2} \subseteq  \ldots  \subseteq {Z_i} \subseteq  \ldots  \subseteq {Z_k} \subseteq A\left( {i = 1,2, \ldots ,k} \right)$.
\end{mydef}

At each decision-level, the evaluation information of the corresponding attribute of each expert is extracted in turn according to the order of importance of the conditional attribute to get the ${H_i^j}$  of different experts under the subset conditional attribute of current decision-level.

\begin{mydef}\label{aggregation function}
Let $\left( {DHHFLDT} \right)_i^j $ be the extraction decision table for the $jth$ expert under ${Z_i}$, denoted as ${H_i^j}$.
There exists an evaluation information aggregation function $\mathfrak{a}$ satisfies $\mathfrak{a} \left( {H_i^1,H_i^2, \ldots ,H_i^j, \ldots ,H_i^e} \right) = {H_i}$,
where ${H_i}$ is the matrix representation of the fusion decision table ${\left( {DHHFLDT} \right)_i} = \left( {{U_i},{Z_i},{V_i},{g_i}} \right)$ of $e$ experts under ${Z_i}$.
\end{mydef}

A novel fusion framework is developed to propose a method that incorporates multi-level of granularity.
Functions for evaluating information extraction and aggregation have also been designed. Given the design of the aggregation function discussed above, common aggregation function can be considered as aggregation operators used for information fusion.
Various types of linguistic weighted operators have been developed, including Muirhead mean aggregation operators \cite{liu2018some}, and linguistic ordered weighted distance operators \cite{xu2015hesitant}.
The weighted average operatora is widely used and well-characterized aggregation operator proposed by Yager  \cite{yager1988ordered}.
For ease of use, this paper extends the application of the DHHFLWA \cite{gou2019group} operator as the evaluation information aggregation function.

\begin{mydef}
Let a group matrix of $\left\{ {H_i^1,H_i^2, \cdots ,H_i^j, \cdots ,H_i^e} \right\}$ be the extraction decision tables for e experts under ${Z_i}$.
Then, the double hierarchy hesitant fuzzy linguistic matrix weighted averaging operator (DHHFLMWA) is defined as below:
\begin{equation}\label{}
DHHFLMWA\left( {H_i^1,H_i^2, \cdots ,H_i^j, \cdots ,H_i^e} \right) = \mathop  \oplus \limits_{j = 1}^e \left( {w_E^j \cdot H_i^j} \right)
\end{equation}
$\left( {j = 1,2, \cdots ,e;i = 1,2, \ldots ,k} \right)$, where $0 \le w_E^j \le 1$ and $\sum\limits_{j = 1}^e {w_E^j}  = 1$.
\end{mydef}

\begin{theorem}
Let a group matrix of $\left\{ {H_i^1,H_i^2, \cdots ,H_i^j, \cdots ,H_i^e} \right\}$ be the extraction decision table for e experts under ${Z_i}$ and $H_i^j = \left\{ {{{\left( {h_{{S_O}}^j} \right)}_i}} \right\}$. A collection of DHHFLEs are ${\left( {h_{{S_O}}^j} \right)_i} = \left\{ {{s_{\phi _l^j{{\left\langle {{o_{\varphi _l^j}}} \right\rangle }_i}}}\left| {{s_{\phi _l^j{{\left\langle {{o_{\varphi _l^j}}} \right\rangle }_i}}} \in {S_O};l = 1,2,...,{L^j}} \right.} \right\}$, and the weight of experts are denoted by ${w_E} = \left\{ {w_E^1,w_E^2, \ldots ,w_E^{\rm{e}}} \right\}$. Then the DHHFLMWA with linguistic expected-value is calculated as below:
\begin{equation}\label{}
\begin{array}{*{20}{l}}
{DHHFLMWA\left( {H_i^1,H_i^2, \cdots ,H_i^j, \cdots ,H_i^e} \right)}\\
{ = le\left[ {\left( {w_E^1H_i^1} \right) \oplus \left( {w_E^2H_i^2} \right) \oplus  \cdots  \oplus \left( {w_E^jH_i^j} \right) \oplus  \cdots  \oplus \left( {w_E^eH_i^e} \right)} \right]}\\
{ = w_E^1\left\{ {le{{\left( {h_{{S_O}}^1} \right)}_i}} \right\} \oplus w_E^2\left\{ {le{{\left( {h_{{S_O}}^2} \right)}_i}} \right\} \oplus  \cdots  \oplus w_E^j\left\{ {le{{\left( {h_{{S_O}}^j} \right)}_i}} \right\} \oplus  \cdots  \oplus w_E^e\left\{ {le{{\left( {h_{{S_O}}^e} \right)}_i}} \right\}}
\end{array}.
\end{equation}
Substituting into Eq.(\ref{functionle}) gives:
\begin{equation}\label{}
\begin{array}{*{20}{l}}
{w_E^1\left\{ {le{{\left( {h_{{S_O}}^1} \right)}_i}} \right\} \oplus w_E^2\left\{ {le{{\left( {h_{{S_O}}^2} \right)}_i}} \right\} \oplus  \cdots  \oplus w_E^j\left\{ {le{{\left( {h_{{S_O}}^j} \right)}_i}} \right\} \oplus  \cdots  \oplus w_E^e\left\{ {le{{\left( {h_{{S_O}}^e} \right)}_i}} \right\}}\\
{ = w_E^1\left\{ {{s_{{\phi ^1}{{\left\langle {{o_{{\varphi ^1}}}} \right\rangle }_i}}}} \right\} \oplus w_E^2\left\{ {{s_{{\phi ^2}{{\left\langle {{o_{{\varphi ^2}}}} \right\rangle }_i}}}} \right\} \oplus  \cdots  \oplus w_E^j\left\{ {{s_{{\phi ^j}{{\left\langle {{o_{{\varphi ^j}}}} \right\rangle }_i}}}} \right\} \oplus  \cdots  \oplus w_E^e\left\{ {{s_{{\phi ^e}{{\left\langle {{o_{{\varphi ^e}}}} \right\rangle }_i}}}} \right\}}
\end{array},
\end{equation}
where $w_E^j\left\{ {{s_{{\phi ^j}{{\left\langle {{o_{{\varphi ^j}}}} \right\rangle }_i}}}} \right\} = w_E^j\left\{ {{s_{\frac{1}{{{L^j}}}\sum\limits_{l = 1}^{{L^j}} {\phi _l^j} {{\left\langle {{o_{\frac{1}{{{L^j}}}\sum\limits_{l = 1}^{{L^j}} {\varphi _l^j} }}} \right\rangle }_i}}}} \right\}$, using Eq.(\ref{functionle}) could get the final aggregation results of $e$ experts under ${Z_i}$ is ${H_i} = \left\{ {{s_{\sum\limits_{j = 1}^e {w_E^j{\phi ^j}} {{\left\langle {{o_{\sum\limits_{j = 1}^e {w_E^j{\varphi ^j}} }}} \right\rangle }_i}}}} \right\}$.
\end{theorem}

\subsection{S3W-GDM method based on multi-level granularity fusion}\label{S3WD}
When it comes to solving complex GDM processes, based on S3WD the multi-level of granularity is a useful approach.
Alternatives at each decision-level are classified into three possible results: acceptance, rejection, or non-commitment, as a way to increase the efficiency of the decision alternatives selection process.
The following will demonstrate how it's combined with a S3WD model into the GDM process.

\begin{figure}[htbp]
  \centering
  \includegraphics[width=13cm]{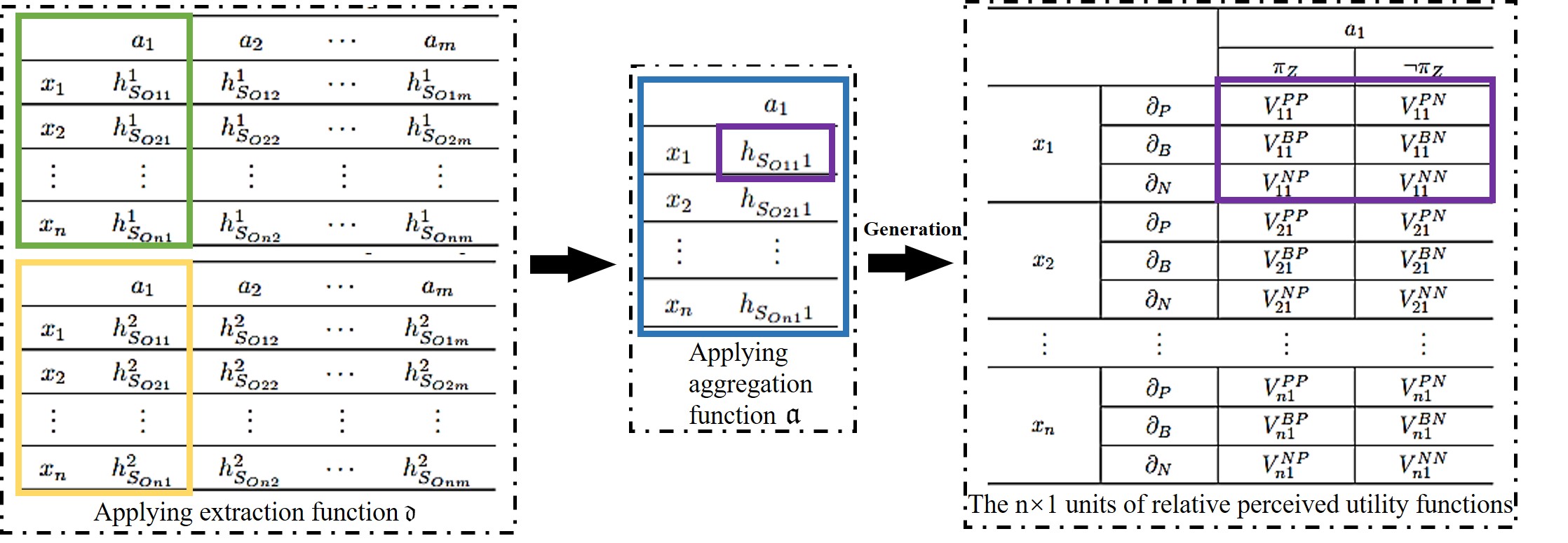}
  \caption{\textbf{.} Multi-level granularity fusion by two experts' decision tables.}
  \label{modelpresentation}
\end{figure}

Given a ${\left( {DHHFLDT} \right)_i} = \left( {{U_i},{Z_i},{V_i},{g_i}} \right)$, ${U_i}$ denotes the processing alternatives, $G{l}\left( {{Z},} \right.$\\$\left. {{\kappa }} \right)$ is the 4-tuple, $Z = \left\{ {{Z_1},{Z_2}, \ldots ,{Z_i}, \ldots ,{Z_k}} \right\}$ be a nested sequence of conditional attributes, and $\kappa  = \left\{ {{\kappa _1},{\kappa _2}, \ldots ,{\kappa _i}, \ldots ,{\kappa _k}} \right\}$ be a sequence of similarity thresholds.
The multi-level granular structure $Gl\left( {Z,\kappa } \right) = \left\{ {G{l_1}\left( {{Z_1},{\kappa _1}} \right),G{l_2}\left( {{Z_2},{\kappa _2}} \right), \ldots ,G{l_k}\left( {{Z_k},{\kappa _k}} \right)} \right\}$ is defined as:
\begin{equation}\label{granularstructure}
\begin{array}{*{20}{l}}
{1st\;decision - level:G{l_1} = \left\{ {{{\left( {DHHFLDT} \right)}_1},{\kappa _1},pr\left( {{\pi _{{Z_1}}}\left| {{\kappa _{{\aleph _{{Z_1}}}}}\left( x \right)} \right.} \right),V_{{Z_1}}^{ \circ  \bullet }} \right\}}\\
{2nd\;decision - level:G{l_2} = \left\{ {{{\left( {DHHFLDT} \right)}_2},{\kappa _2},pr\left( {{\pi _{{Z_2}}}\left| {{\kappa _{{\aleph _{{Z_2}}}}}\left( x \right)} \right.} \right),V_{{Z_2}}^{\circ  \bullet }} \right\}}\\
{\;\;\;\;\; \vdots \;\;\;\;\;\;\;\;\;\;\;\;\;\;\;\;\;\;\;\;\; \vdots \;\;\;\;\;\;\;\;\;\;\;\;\;\;\;\;\;\;\;\;\;\;\;\;\;\;\;\;\;\;\;\;\;\;\;\;\;\;\;\;\;\;\; \vdots \;\;\;}\\
{ith\;decision - level:G{l_i} = \left\{ {{{\left( {DHHFLDT} \right)}_i},{\kappa _i},pr\left( {{\pi _{{Z_i}}}\left| {{\kappa _{{\aleph _{{Z_i}}}}}\left( x \right)} \right.} \right),V_{{Z_i}}^{\circ  \bullet }} \right\}}\\
{\;\;\;\;\; \vdots \;\;\;\;\;\;\;\;\;\;\;\;\;\;\;\;\;\;\;\;\; \vdots \;\;\;\;\;\;\;\;\;\;\;\;\;\;\;\;\;\;\;\;\;\;\;\;\;\;\;\;\;\;\;\;\;\;\;\;\;\;\;\;\;\;\; \vdots \;}\\
{kth\;decision - level:G{l_k} = \left\{ {{{\left( {DHHFLDT} \right)}_k},{\kappa _k},pr\left( {{\pi _{{Z_k}}}\left| {{\kappa _{{\aleph _{{Z_k}}}}}\left( x \right)} \right.} \right),V_{{Z_k}}^{\circ  \bullet }} \right\}}
\end{array},
\end{equation}
where at the $ith$ decision-level of $G{l_i}\left( {{Z_i},{\kappa _i}} \right)$, ${U_i}$ satisfies $\left| {{U_1}} \right| \ge \left| {{U_2}} \right| \ge  \ldots  \ge \left| {{U_i}} \right| \ge  \ldots  \ge \left| {{U_k}} \right|$, ${Z_i}$ satisfies ${Z_1} \subseteq {Z_2} \subseteq  \ldots  \subseteq {Z_i} \subseteq  \ldots  \subseteq {Z_k} \subseteq A$, ${\kappa _i}$ satisfies $0 \le {\kappa _1} \le {\kappa _2} \le  \ldots  \le {\kappa _k} \le 1$, ${\sigma _1} \ge {\sigma _2} \ge  \ldots  \ge {\sigma _i} \ge  \ldots  \ge {\sigma _k}$, $pr\left( {{\pi _{{Z_i}}}\left| {{\kappa _{{\aleph _{{Z_i}}}}}\left( x \right)} \right.} \right)$ denotes conditional probabilitise, and $V_{{Z_i}}^{ \circ  \bullet }\left( { \circ  = P,B,N; \bullet  = P,N} \right)$ denotes relative perceived utility functions.
To demonstrate a detailed method of operation, an example is given below.
\begin{example}
Two experts have already given evaluation decision tables ${\left( {DHHFLDT} \right)^1}$ and ${\left( {DHHFLDT} \right)^2}$, with conditional attributes ordered by importance.
Assuming that $a_1$ is the most important attributes, it means ${Z_1} = \left\{ {{a_1}} \right\}$.
According to Definition \ref{extraction function}, the decision tables $\left( {DHHFLDT} \right)_1^1$ and $\left( {DHHFLDT} \right)_1^2$ of the two experts under the $1st$ decision-level can be extracted respectively.
Definition \ref{aggregation function} can then be used to obtain the important granular structure $Gl_1\left( {Z_1,\kappa_1 } \right) $ of the two experts at the $1st$ decison-level, resulting in the fused decision table ${\left( {DHHFLDT} \right)_1}$.
Clearly, $U_1 = \left\{ {{x_1},{x_2}, \ldots ,{x_n}} \right\}$.
Having obtained this level of granularity, the unit of relative gain function introduced in Section \ref{3WD model} can be constructed.
The $n \times 1$ unit relative perceived functions ${V_{{Z_1}}^{ \circ  \bullet }}$ can be created.
Deriving $pr\left( {{\pi _{{Z_1}}}\left| {{\kappa _{{\aleph _{{Z_1}}}}}\left( x \right)} \right.} \right)$ is straightforward using the method described in Section \ref{3.1}.
The specific operation of the above process is demonstrated in Fig. \ref{modelpresentation}. Finally, the rules of the S3WD are utilized to derive the results of the $1st$ decision-level of granular structure, which is used to drive the S3W-GDM process.
\end{example}

Given a multi-level structure $Gl\left( {Z,\kappa } \right) = \left\{ {G{l_1}\left( {{Z_1},{\kappa _1}} \right),} \right.$$\left. {G{l_2}\left( {{Z_2},{\kappa _2}} \right), \ldots ,G{l_k}\left( {{Z_k},{\kappa _k}} \right)} \right\}$ for any the $ith$ decision-level of $G{l_i}\left( {{Z_i},{\kappa _i}} \right)$, the expected perceived utility of ${{\kappa _{{\aleph _{{Z_i}}}}}\left( x \right)}$ is:
\begin{equation}\label{expectedperceivedutility}
\begin{array}{*{20}{l}}
{E\mathbb{V}{_i}\left( {{\partial _P}\left| x \right.} \right) = \mathbb{V}_{{Z_i}}^{PP}pr\left( {{\pi _{{Z_i}}}\left| {{\kappa _{{\aleph _{{Z_i}}}}}\left( x \right)} \right.} \right) + \mathbb{V}_{{Z_i}}^{PN}pr\left( {\neg {\pi _{{Z_i}}}\left| {{\kappa _{{\aleph _{{Z_i}}}}}\left( x \right)} \right.} \right),}\\
{E\mathbb{V}{_i}\left( {{\partial _B}\left| x \right.} \right) = \mathbb{V}_{{Z_i}}^{BP}pr\left( {{\pi _{{Z_i}}}\left| {{\kappa _{{\aleph _{{Z_i}}}}}\left( x \right)} \right.} \right) + \mathbb{V}_{{Z_i}}^{BN}pr\left( {\neg {\pi _{{Z_i}}}\left| {{\kappa _{{\aleph _{{Z_i}}}}}\left( x \right)} \right.} \right),}\\
{E\mathbb{V}{_i}\left( {{\partial _N}\left| x \right.} \right) = \mathbb{V}_{{Z_i}}^{NP}pr\left( {{\pi _{{Z_i}}}\left| {{\kappa _{{\aleph _{{Z_i}}}}}\left( x \right)} \right.} \right) + \mathbb{V}_{{Z_i}}^{NN}pr\left( {\neg {\pi _{{Z_i}}}\left| {{\kappa _{{\aleph _{{Z_i}}}}}\left( x \right)} \right.} \right).}
\end{array}
\end{equation}

The perceived utility from regret theory is chosen as the basis for formulating decision rules, as relative perceived utility aligns with the subjective of utility.
Therefore, the maximized expected perceived utility is selected as the most appropriate action according to Bayesian.
These rules are derived as follows:

(P1) If $E\mathbb{V}{_i}\left( {{\partial _P}\left| x \right.} \right) \ge E\mathbb{V}{_i}\left( {{\partial _B}\left| x \right.} \right)$ and $E\mathbb{V}{_i}\left( {{\partial _P}\left| x \right.} \right) \ge E\mathbb{V}{_i}\left( {{\partial _N}\left| x \right.} \right)$, then $x \in PO{S_i}\left( {{\pi _{{Z_i}}}} \right)$,

(B1) If $E\mathbb{V}{_i}\left( {{\partial _B}\left| x \right.} \right) \ge E\mathbb{V}{_i}\left( {{\partial _P}\left| x \right.} \right)$ and $E\mathbb{V}{_i}\left( {{\partial _B}\left| x \right.} \right) \ge E\mathbb{V}{_i}\left( {{\partial _N}\left| x \right.} \right)$, then $x \in BN{D_i}\left( {{\pi _{{Z_i}}}} \right)$,

(N1) If $E\mathbb{V}{_i}\left( {{\partial _N}\left| x \right.} \right) \ge E\mathbb{V}{_i}\left( {{\partial _P}\left| x \right.} \right)$ and $E\mathbb{V}{_i}\left( {{\partial _N}\left| x \right.} \right) \ge E\mathbb{V}{_i}\left( {{\partial _B}\left| x \right.} \right)$, then $x \in NE{G_i}\left( {{\pi _{{Z_i}}}} \right)$.
where $PO{S_i}\left( {{\pi _{{Z_i}}}} \right)\bigcup {BN{D_i}\left( {{\pi _{{Z_i}}}} \right)\bigcup {NE{G_i}\left( {{\pi _{{Z_i}}}} \right)} }  = {U_i}$.

The granular structure $Gl\left( {Z,\kappa } \right)$ of each decision-level is used for decision-making, and is able to do the classification of each alternative to generate the decision results for the current decision-level.
Seven S3WD models were summarized for different dynamic scenarios \cite{yang2020multilevel}.
The feasibility of these seven dynamic decision-making solutions is supported by the fact that the decision data set has prior information, which can be tested for accuracy in each region.
However, since there is no prior information in GDM, the decision-level is based on the results derived from the granular structure of the previous level.
Therefore, this paper considers not only improving the efficiency of decision-making but also enhancing the rationality of decision results.
The most reasonable dynamic decision-making situation is set up through the GDM problems, and the relevant parameter settings are strictly followed.
In the absence of prior information, the rationality of parameter settings can ensure the accuracy of decision results to a greater extent.

\begin{figure}[htbp]
  \centering
  \includegraphics[width=6cm]{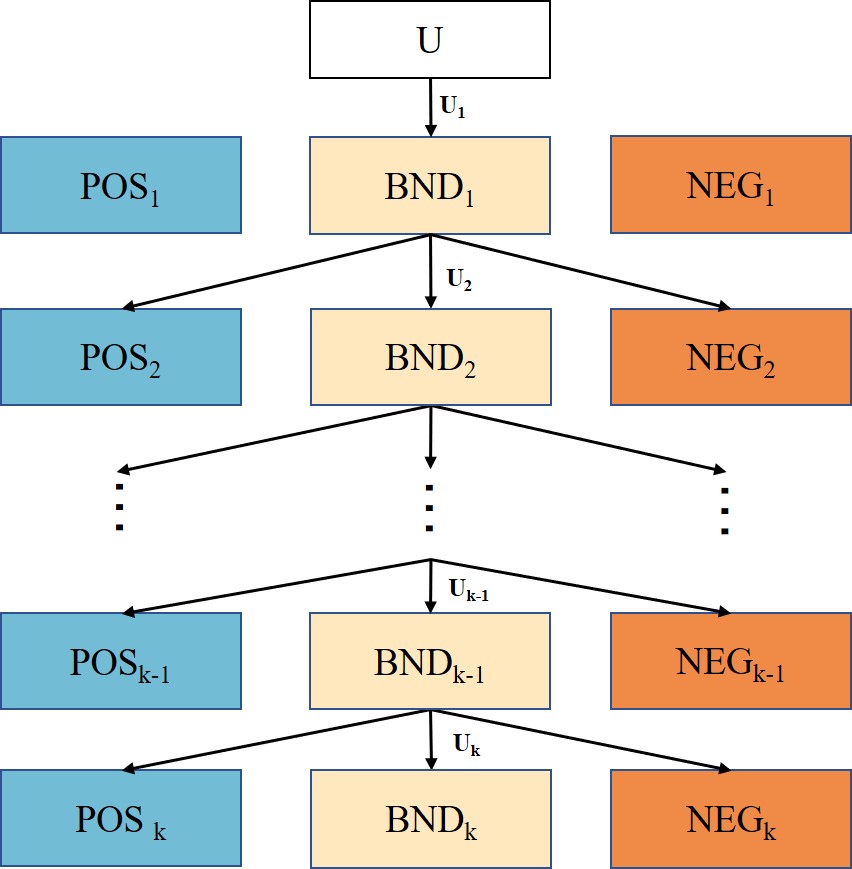}
  \caption{\textbf{.} The process of multi-level S3WD.}
  \label{universeprocess}
\end{figure}

Assume ${U_i} = BN{D_{i - 1}}\left( {{\pi _{i - 1}}} \right)\left( {1 < i \le k} \right)$ from $\left( {i - 1} \right)th$ decision-level to $ith$ level if a top-down manner is adopted.
Fig. \ref{universeprocess} clearly illustrates one of the most common S3WD models.
Another crucial problem to be solved for the GDM problem is the ranking of the alternatives classified at each level, especially those classified in the positive (negative) domain, according to the target concept.

The ranking of alternatives follows a specific order based on their expected perceived utility, denoted as $PO{S_i} \succ BN{D_i} \succ NE{G_i}$.
When the target concept is accepted and alternatives fall within the positive domain, a higher expected perceived utility signifies stronger alignment with the desired result.
Conversely, if the target concept is rejected and alternatives are classified into the negative domain, a higher expected perceived utility value indicates a greater degree of misalignment with the undesirable result.
In essence, prioritization within the positive domain favors alternatives with higher expected utility (positive direction), while prioritization within the negative domain favors alternatives with lower expected utility (reverse direction).
If an alternative is ultimately classified in the boundary domain, the ranking of its expected perceived utility is determined based on the results of the final analysis at this level.

\subsection{The algorithms of S3W-GDM}\label{algorithms}
Based on the multi-level granularity fusion, S3WD is a useful model to address GDM problems.
Additionally, neighborhood theory and regret theory support the combination of DHHFLTS and the S3WD model.
With these works, the novel S3W-GDM method is constructed.
Two algorithms will be presented in the following sections.
Algorithm \ref{algorithm1} dynamically extracts and fuses decision tables by iterating through each decision-level based on the importance of conditional attributes.
Algorithm \ref{algorithm2} describes the detailed dynamic decision-making process of the multi-level S3WD, including the information aggregation of decision-level, the construction method of $\kappa$-cut neighborhood binary relation at each decision-level, the coarse-grained representation of the relative utility functions, and the expectation of the integrated perceived S3WD rule for comparison.
\begin{algorithm}[htpb]
\caption{The extraction and aggregation of expert decision table.}
\label{algorithm1}
\begin{algorithmic}[1]
\REQUIRE
The decision table of experts $\left( {DHHFLDT} \right)_{}^j$;
${w_E} = \left\{ {w_E^1,w_E^2, \ldots ,w_E^{\rm{e}}} \right\}$;
$w = \left\{ {{w_1},{w_2}, \ldots ,{w_m}} \right\}$; and $k_{max}$.
\ENSURE
The fused decision table ${\left( {DHHFLDT} \right)_i} $.
\STATE \textbf{begin}
\FOR{$w = \left\{ {{w_1},{w_2}, \ldots ,{w_m}} \right\}$}
\STATE Rank by value in descending order to obtain a nested sequence of conditional attributes $Z_1 \subseteq Z_2 \subseteq ... \subseteq Z_k \subseteq A $.
\ENDFOR
\FOR{each level $k$ from 1 to $k_{max}$}
    \FOR{each $\left( {DHHFLDT} \right)_{}^j$ }
      \STATE Extract the top $i$ attributes in $Z_i$ and add to $\left( {DHHFLDT} \right)_i^j $.
    \ENDFOR
\ENDFOR
\FOR{each $\left( {DHHFLDT} \right)_i^j $}
\STATE Apply ${w_E}$ to compute a weighted average for fusing $\left( {DHHFLDT} \right)_i^j $.
\ENDFOR
\STATE \textbf{return} The fused decision table ${\left( {DHHFLDT} \right)_i} $.
\STATE \textbf{end}
\end{algorithmic}
\end{algorithm}

\begin{algorithm}[ht]
\caption{The multi-level S3W-GDM method.}
\label{algorithm2}
\begin{algorithmic}[1]
        \REQUIRE
        The fused decision table ${\left( {DHHFLDT} \right)_i}$;
        A nested sequence of conditional attributes and parameters $Z_1 \subseteq Z_2 \subseteq ... \subseteq Z_k \subseteq A $, $0 \le {\kappa _1} \le {\kappa _2} \le  \ldots  \le {\kappa _k} \le 1$ and ${\sigma _1} \ge {\sigma _2} \ge  \ldots  \ge {\sigma _i} \ge  \ldots  \ge {\sigma _k}$;
        The first hierarchy linguistic scale $\tau $, the second hierarchy linguistic scale $\varsigma $;
        $\theta ,\eta $;
        $w = \left\{ {{w_1},{w_2}, \ldots ,{w_m}} \right\}$.
        \ENSURE
        The results of each alternative ${x_p}\left( {p= 1, \ldots ,n} \right)$ in POS, BNG, and NEG.
        \STATE \textbf{begin}
        \STATE $POS = \emptyset, BND = \emptyset, NEG = \emptyset;$
        \STATE $i = 1; U_1 = U; \pi_1 = \pi;$
        \FOR{$x_p \in U_i$}
            \STATE Calculate the conditional probabilities based on Eq.(\ref{simlilaritydegeree}-\ref{outrankingrelationconditionalprobability})
        \ENDFOR
        \FOR{$x_p \in U_i$}
        \STATE Compute the relative perceived utility functions based on Eq.(\ref{SGdirectutility}-\ref{comprehensiveperceivedutility})
        \ENDFOR
        \WHILE{$U_i \neq \emptyset$ and $1 \leq i \leq k$}
            \STATE  Construct the $ith$ level of $G{l_i}\left( {{Z_i},{\kappa _i}} \right)$ with Eq.(\ref{granularstructure})
            \FOR{$x_p \in U_i$}
            \STATE Decision-making through Eq.(\ref{expectedperceivedutility}) and (P1)-(N1).
               \IF {$PO{S_i}\left( {{\pi _{{Z_i}}}} \right)$ }
               \STATE ${x_p} \Rightarrow {\partial _P}$.
               \ENDIF
               \IF {$BN{D_i}\left( {{\pi _{{Z_i}}}} \right)$ }
               \STATE ${x_p} \Rightarrow {\partial _B}$.
               \ENDIF
               \IF {$NE{G_i}\left( {{\pi _{{Z_i}}}} \right)$ }
               \STATE ${x_p} \Rightarrow {\partial _N}$.
               \ENDIF
            \ENDFOR
            \STATE $POS = POS \cup PO{S_i}\left( {{\pi _{{Z_i}}}} \right);$
            \STATE $BND = BND \cup BN{D_i}\left( {{\pi _{{Z_i}}}} \right);$
            \STATE $NEG = NEG \cup NE{G_i}\left( {{\pi _{{Z_i}}}} \right);$
            \STATE $\pi_{i} = \pi_i \cap U_{i+1};$
            \STATE $i = i + 1;$
        \ENDWHILE
        \STATE \textbf{return} The results of each alternative ${x_p}\left( {p= 1, \ldots ,n} \right)$ in POS, BNG, and NEG.
        \STATE \textbf{end}
\end{algorithmic}
\end{algorithm}

\section{An illustrative example}\label{illustrative and comparative}
This section performs case analysis based on an illustrative example to verify the applicability of the established multi-level S3W-GDM method.
In Section \ref{5.1}, multiple experts' SLE diagnostic problem characterized by vagueness, hesitation, and variation is presented.
The S3W-GDM method is applied to obtain the classification and ranking results of SLE patients in Section \ref{5.2}.

\subsection{Description of the problem}\label{5.1}
In recent years, granular computing and S3WD have been widely applied in the medical field, primarily focusing on the prediction and classification of patient populations.
Furthermore, existing literature indicates that granular computing and S3WD provide an effective reasoning paradigm for dynamic medical diagnostics.
However, the reality of dynamic decision-making in medical diagnostics is exceedingly complex, involving the diversity of disease characteristics, the uncertainty of treatment decisions, and the phased involvement of multidisciplinary experts.
Taking the diagnosis of SLE as an example, such decision-making issues typically exhibit characteristics of vagueness, hesitation, and variation.
Regrettably, there is a lack of further research on existing decision-making methods or models that consider these characteristics simultaneously.
And the current decision-making methods regarding DHHFLTS are not suitable for GDM problems with the above characteristics.
In light of this, Section \ref{processing} establishes the S3W-GDM method.
To demonstrate the applicability of the established models, an illustrative example of SLE diagnosis is presented.

SLE is an autoimmune disease that primarily affects multiple systems and organs, leading to complex and varied clinical manifestations.
It predominantly occurs in women of childbearing age.
Regular follow-up and monitoring of the condition are necessary to detect and manage relapses early and to achieve accurate diagnosis and delineation in SLE patients.
Some characteristic clinical manifestations can provide clues for the early diagnosis of SLE, such as arthralgia, rash, nephritis, serological changes, immunosuppression, and psychiatric symptoms.
Due to its impact on multiple systems and organs, SLE results in a diverse range of clinical manifestations.
In the face of this complexity, there is a need to combine the instructions of experts from different disciplines as quickly as possible to give the results of the patient's diagnosis, assessment, prediction, and other decision-making.

Let $U = \left\{ {{x_1},{x_2},{x_3},{x_4},{x_5},{x_6},{x_7},{x_8}} \right\}$ be the set of women of childbearing age who may have SLE.
The hospital has implemented a multidisciplinary dynamic diagnosis approach for a particular group of patients to enhance diagnostic efficiency and accuracy.
This approach involves conducting multi-level diagnoses through a joint consultation of experts from various fields such as rheumatology, nephrology, and dermatology, who collaborate to diagnose the patients.
Since SLE cannot be diagnosed based solely on the unique indicators of any single discipline, all experts use a consistent set of attributes to arrive at a diagnosis.
These attributes include Antinuclear antibody (ANA), Anti-double-stranded DNA antibody (Anti-dsDNA), Complement protein (C3 and C4) levels, Skin and mucous membrane damage, arthritis, and kidney involvement.
These attributes not only help in the diagnosis of SLE but also aid in ruling out other diseases.
Research \cite{ye2022variable} has been conducted to determine whether these attributes are associated with cost or benefit.
To gather accurate evaluations from experts, the collection of evaluation content is adjusted and the types of attributes used are shown in Table \ref{attributestable}.
$A = \left\{ {{a_1},{a_2},{a_3},{a_4},{a_5},{a_6}} \right\}$ is represented by those six attributes.
According to the order in which experts prioritize these attributes in SLE diagnosis, their importance is set as $w = \left\{ {0.2,0.3,0.15,0.1,0.1,0.15} \right\}$.
For each patient, it is assumed that they exist in two states $\pi $ and $\neg {\pi}$, corresponding to having SLE and not having SLE,
At the same time, there are three actions ${\partial _P}$, ${\partial _B}$, and ${\partial _N}$, which correspond to a confirmed diagnosis requiring immediate treatment, a pending diagnosis requiring further testing, and no treatment but requiring regular inspection and monitoring.

\begin{figure}[t]
  \centering
  \includegraphics[width=7.3cm]{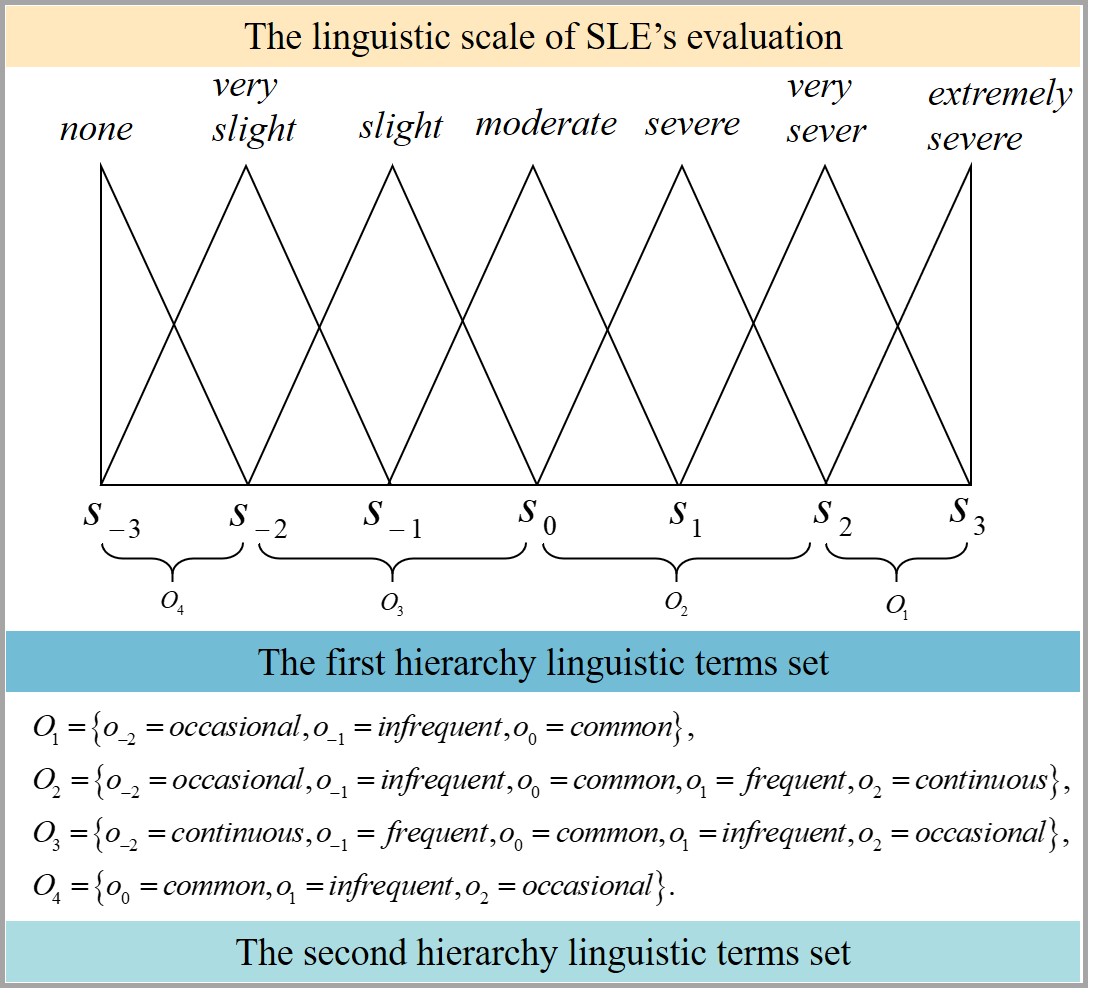}
  \caption{\textbf{.} Distributions of the first and second hierarchy linguistic scales.}
  \label{fig5}
\end{figure}

\begin{table}[htpb]\small
\centering
\renewcommand{\arraystretch}{1.5}%
\setlength{\abovecaptionskip}{1.5pt}%
\captionsetup{justification=centering, singlelinecheck=false, labelsep=quad}
\caption{The explanations of conditional attributes.}
\label{}
\setlength{\tabcolsep}{2mm}{  %
\begin{tabular}{ll}
\hline
\multicolumn{1}{l}{Conditional sattributes} & \multicolumn{1}{l}{The connotation of key indicators for diagnosing SLE}  \\	
\hline
${a_1}$ ANA positivity& \makecell[l]{The ANA positivity rate can reach more than 95$\%$ \\and is the most sensitive screening indicator for SLE.}		\\
${a_2}$ Degree of Anti-dsDNA		& \makecell[l]{It is closely related to SLE disease activity, \\and an increase in titer indicates disease activity.}	\\
${a_3}$ Reduction degree of C3 and C4	& \makecell[l]{Represents complement activation, and the decrease \\in C3 and C4 indicates disease activity.}	\\
${a_4}$ Degree of skin and mucosal damage	& \makecell[l]{It means that SLE has active manifestations, \\such as discoid erythema, oral ulcers, etc.}\\
${a_5}$ Degree of renal involvement		& \makecell[l]{Proteinuria or renal failure indicates that SLE \\involves the kidneys and requires active treatment.}		\\
${a_6}$ Arthritis pain level		& \makecell[l]{Pain and joint swelling indicate active arthritis in SLE.}		\\
\hline
  \end{tabular}}
\makecell[l]{\textbf{Note}: Conditional attributes are set with reference to https://pubmed.ncbi.nlm.nih.gov/, https://\\medlineplus.gov/, and https://rheumatology.org/. In this case, the conditional attributes are all\\ cost-type, i.e., the greater the rating, the worse the indicator.}
\label{attributestable}
\end{table}
Based on these attributes, the prepared linguistic scale of the DHHFLTS is designed to collect expert evaluation information, shown in Fig. \ref{fig5}.
The first and second linguisstic scale of the DHHFLTS are $\tau  = 3,\zeta  = 3$, respectively.
After consulting with medical experts in the field, they were invited to simulate a consultation based on their personal experience.
The three experts $e = 3$ assessed 8 patients ($n = 8$) by using the DHHFLTS,
and the results are collected into Table \ref{expert 1}-\ref{expert 3}.
The weight of the three experts is set to ${w_E} = \left\{ {0.5,0.3,0.2} \right\}$ based on their knowledge of the overall situation of SLE.
According to Section \ref{processing}, the algorithm repeatedly extracts and fuses the expert evaluation tables.
The parameters of each granular structure are adjusted to the same values for consistency.

\begin{table}[htpb]
\centering
\renewcommand{\arraystretch}{2}%
\caption{The evaluation decision table of $E_1$}
\label{expert 1}
\hspace*{-2cm}
{\small
\begin{tabular}{p{0.3cm}p{2cm}p{2.4cm}p{2.2cm}p{2.2cm}p{2.2cm}p{2.4cm}}
\hline
    & ${a_1}$ & ${a_2}$ & ${a_3}$ & ${a_4}$ & ${a_5}$ & ${a_6}$ \\
\hline
${x_1}$ & $\left\{ {{s_{ - 1\left\langle {{o_0}} \right\rangle }}} \right\}$ & $\left\{ {s_{1\left\langle {{o_{ - 2}}} \right\rangle }} \right\}$ & $\left\{ {{s_{ - 1\left\langle {{o_3}} \right\rangle }}} \right\}$ & $\left\{ {{s_{2\left\langle {{o_3}} \right\rangle }}} \right\}$ & $\left\{ {{s_{ - 2\left\langle {{o_0}} \right\rangle }}} \right\}$ & $\left\{ {{s_{ - 3\left\langle {{o_2}} \right\rangle }},{s_{ - 2\left\langle {{o_0}} \right\rangle }}} \right\}$ \\

${x_2}$ & $\left\{ {{s_{2\left\langle {{o_0}} \right\rangle }},{s_{2\left\langle {{o_1}} \right\rangle }}} \right\}$ & $\left\{ {{s_{3\left\langle {{o_0}} \right\rangle }}} \right\}$ & $\left\{ {{s_{ - 1\left\langle {{o_{ - 3}}} \right\rangle }}} \right\}$ & $\left\{ {{s_{2\left\langle {{o_{ - 2}}} \right\rangle }}} \right\}$ & $\left\{ {{s_{0\left\langle {{o_1}} \right\rangle }},{s_{ - 1\left\langle {{o_1}} \right\rangle }}} \right\}$ & $\left\{ {{s_{ - 1\left\langle {{o_{ - 3}}} \right\rangle }}} \right\}$ \\

${x_3}$ & $\left\{ {{s_{3\left\langle {{o_{ - 1}}} \right\rangle }}} \right\}$ & $\left\{ {{s_{0\left\langle {{o_{ - 3}}} \right\rangle }},{s_{ - 1\left\langle {{o_3}} \right\rangle }}} \right\}$ & $\left\{ {{s_{1\left\langle {{o_{ - 3}}} \right\rangle }}} \right\}$ & $\left\{ {{s_{0\left\langle {{o_1}} \right\rangle }},{s_{0\left\langle {{o_{ - 3}}} \right\rangle }}} \right\}$ & $\left\{ {{s_{2\left\langle {{o_3}} \right\rangle }}} \right\}$ & $\left\{ {{s_{ - 1\left\langle {{o_3}} \right\rangle }},{s_{0\left\langle {{o_{ - 1}}} \right\rangle }}} \right\}$ \\

${x_4}$ & $\left\{ {{s_{2\left\langle {{o_{ - 2}}} \right\rangle }}} \right\}$ & $\left\{ {{s_{3\left\langle {{o_{ - 3}}} \right\rangle }},{s_{3\left\langle {{o_{ - 1}}} \right\rangle }}} \right\}$ & $\left\{ {{s_{1\left\langle {{o_2}} \right\rangle }},{s_{0\left\langle {{o_{ - 1}}} \right\rangle }}} \right\}$ & $\left\{ {{s_{ - 1\left\langle {{o_{ - 3}}} \right\rangle }}} \right\}$ & $\left\{ {{s_{ - 1\left\langle {{o_{ - 1}}} \right\rangle }}} \right\}$ & $\left\{ {{s_{3\left\langle {{o_{ - 1}}} \right\rangle }}} \right\}$ \\

${x_5}$ & $\left\{ {{s_{ - 3\left\langle {{o_0}} \right\rangle }}} \right\}$ & $\left\{ {{s_{ - 2\left\langle {{o_3}} \right\rangle }}} \right\}$ & $\left\{ {{s_{1\left\langle {{o_2}} \right\rangle }}} \right\}$ & $\left\{ {{s_{2\left\langle {{o_2}} \right\rangle }},{s_{3\left\langle {{o_{-1}}} \right\rangle }}} \right\}$ & $\left\{ {{s_{ - 3\left\langle {{o_1}} \right\rangle }}} \right\}$ & $\left\{ {{s_{0\left\langle {{o_1}} \right\rangle }}} \right\}$ \\

${x_6}$ & $\left\{ {{s_{0\left\langle {{o_3}} \right\rangle }}} \right\}$ & $\left\{ {{s_{1\left\langle {{o_0}} \right\rangle }}} \right\}$ & $\left\{ {{s_{ - 2\left\langle {{o_1}} \right\rangle }},{s_{ - 3\left\langle {{o_3}} \right\rangle }}} \right\}$ & $\left\{ {{s_{ - 2\left\langle {{o_{ - 1}}} \right\rangle }}} \right\}$ & $\left\{ {{s_{ - 3\left\langle {{o_1}} \right\rangle }}} \right\}$ & $\left\{ {{s_{ - 2\left\langle {{o_2}} \right\rangle }}} \right\}$ \\

${x_7}$ & $\left\{ {{s_{1\left\langle {{o_{ - 2}}} \right\rangle }}} \right\}$ & $\left\{ {{s_{3\left\langle {{o_{ - 3}}} \right\rangle }}} \right\}$ & $\left\{ {{s_{1\left\langle {{o_0}} \right\rangle }}} \right\}$ & $\left\{ {{s_{1\left\langle {{o_1}} \right\rangle }}} \right\}$ & $\left\{ {{s_{1\left\langle {{o_3}} \right\rangle }}} \right\}$ & $\left\{ {{s_{2\left\langle {{o_1}} \right\rangle }}} \right\}$ \\

${x_8}$ & $\left\{ {{s_{ - 2\left\langle {{o_2}} \right\rangle }}} \right\}$ & $\left\{ {{s_{ - 1\left\langle {{o_{ - 1}}} \right\rangle }}} \right\}$ & $\left\{ {{s_{ - 1\left\langle {{o_{ - 3}}} \right\rangle }}} \right\}$ & $\left\{ {{s_{1\left\langle {{o_2}} \right\rangle }},{s_{2\left\langle {{o_0}} \right\rangle }}} \right\}$ & $\left\{ {{s_{2\left\langle {{o_{ - 3}}} \right\rangle }},{s_{1\left\langle {{o_3}} \right\rangle }}} \right\}$ & $\left\{ {{s_{ - 2\left\langle {{o_{ - 2}}} \right\rangle }}} \right\}$ \\
\hline
\end{tabular}
}
\end{table}

\begin{table}[htpb]
\centering
\renewcommand{\arraystretch}{2}%
\caption{The evaluation decision table of $E_2$ }
\label{expert 2}
\hspace*{-2cm}
{\small
\begin{tabular}{p{0.3cm}p{2.4cm}p{2.3cm}p{2.4cm}p{2.3cm}p{2.5cm}p{2.4cm}}
\hline
    & ${a_1}$ & ${a_2}$ & ${a_3}$ & ${a_4}$ & ${a_5}$ & ${a_6}$ \\
\hline
${x_1}$ & $\left\{ {{s_{ - 1\left\langle {{o_3}} \right\rangle }},{s_{1\left\langle {{o_{ 1}}} \right\rangle }}} \right\}$ & $\left\{ {{s_{ - 2\left\langle {{o_2}} \right\rangle }},{s_{ - 2\left\langle {{o_3}} \right\rangle }}} \right\}$ & $\left\{ {{s_{0\left\langle {{o_{ - 3}}} \right\rangle }}} \right\}$ & $\left\{ {{s_{2\left\langle {{o_3}} \right\rangle }},{s_{2\left\langle {{o_{  1}}} \right\rangle }}} \right\}$ & $\left\{ {{s_{ - 2\left\langle {{o_1}} \right\rangle }}} \right\}$ & $\left\{ {{s_{3\left\langle {{o_{-1}}} \right\rangle }}} \right\}$ \\

${x_2}$ & $\left\{ {{s_{ - 1\left\langle {{o_2}} \right\rangle }}} \right\}$ & $\left\{ {{s_{ - 1\left\langle {{o_{ - 1}}} \right\rangle }}} \right\}$ & $\left\{ {{s_{3\left\langle {{o_0}} \right\rangle }}} \right\}$ & $\left\{ {{s_{2\left\langle {{o_0}} \right\rangle }},{s_{1\left\langle {{o_3}} \right\rangle }}} \right\}$ & $\left\{ {{s_{1\left\langle {{o_0}} \right\rangle }},{s_{1\left\langle {{o_2}} \right\rangle }}} \right\}$ & $\left\{ {{s_{ - 3\left\langle {{o_1}} \right\rangle }},{s_{ - 3\left\langle {{o_2}} \right\rangle }}} \right\}$ \\

${x_3}$ & $\left\{ {{s_{0\left\langle {{o_1}} \right\rangle }}} \right\}$ & $\left\{ {{s_{2\left\langle {{o_2}} \right\rangle }}} \right\}$ & $\left\{ {{s_{ - 2\left\langle {{o_0}} \right\rangle }}} \right\}$ & $\left\{ {{s_{ - 1\left\langle {{o_2}} \right\rangle }},{s_{ - 1\left\langle {{o_3}} \right\rangle }}} \right\}$ & $\left\{ {{s_{0\left\langle {{o_0}} \right\rangle }}} \right\}$ & $\left\{ {{s_{2\left\langle {{o_0}} \right\rangle }}} \right\}$ \\

${x_4}$ & $\left\{ {{s_{3\left\langle {{o_0}} \right\rangle }}} \right\}$ & $\left\{ {{s_{ - 1\left\langle {{o_ 2}} \right\rangle }},{s_{0\left\langle {{o_{ - 1}}} \right\rangle }}} \right\}$ & $\left\{ {{s_{ - 3\left\langle {{o_1}} \right\rangle }}} \right\}$ & $\left\{ {{s_{1\left\langle {{o_3}} \right\rangle }}} \right\}$ & $\left\{ {{s_{0\left\langle {{o_1}} \right\rangle }}} \right\}$ & $\left\{ {{s_{3\left\langle {{o_0}} \right\rangle }}} \right\}$ \\

${x_5}$ & $\left\{ {{s_{3\left\langle {{o_{ - 1}}} \right\rangle }}} \right\}$ & $\left\{ {{s_{1\left\langle {{o_1}} \right\rangle }}} \right\}$ & $\left\{ {{s_{1\left\langle {{o_{ - 2}}} \right\rangle }},{s_{1\left\langle {{o_{ - 1}}} \right\rangle }}} \right\}$ & $\left\{ {{s_{0\left\langle {{o_{ - 2}}} \right\rangle }}} \right\}$ & $\left\{ {{s_{ - 1\left\langle {{o_3}} \right\rangle }}} \right\}$ & $\left\{ {{s_{1\left\langle {{o_1}} \right\rangle }}} \right\}$ \\

${x_6}$ & $\left\{ {{s_{2\left\langle {{o_{ - 2}}} \right\rangle }},{s_{2\left\langle {{o_1}} \right\rangle }}} \right\}$ & $\left\{ {{s_{ - 3\left\langle {{o_2}} \right\rangle }},{s_{2\left\langle {{o_{ - 2}}} \right\rangle }}} \right\}$ & $\left\{ {{s_{ - 1\left\langle {{o_2}} \right\rangle }}} \right\}$ & $\left\{ {{s_{3\left\langle {{o_{-1}}} \right\rangle }}} \right\}$ & $\left\{ {{s_{ - 2\left\langle {{o_{ - 3}}} \right\rangle }}} \right\}$ & $\left\{ {{s_{ - 2\left\langle {{o_ 2}} \right\rangle }},{s_{ - 1\left\langle {{o_{ - 1}}} \right\rangle }}} \right\}$ \\

${x_7}$ & $\left\{ {{s_{0\left\langle {{o_{ - 3}}} \right\rangle }},{s_{1\left\langle {{o_{ - 1}}} \right\rangle }}} \right\}$ & $\left\{ {{s_{0\left\langle {{o_0}} \right\rangle }}} \right\}$ & $\left\{ {{s_{1\left\langle {{o_{ - 2}}} \right\rangle }}} \right\}$ & $\left\{ {{s_{1\left\langle {{o_0}} \right\rangle }},{s_{0\left\langle {{o_0}} \right\rangle }}} \right\}$ & $\left\{ {{s_{1\left\langle {{o_0}} \right\rangle }}} \right\}$ & $\left\{ {{s_{ - 1\left\langle {{o_1}} \right\rangle }},{s_{0\left\langle {{o_{ - 2}}} \right\rangle }}} \right\}$ \\

${x_8}$ & $\left\{ {{s_{2\left\langle {{o_0}} \right\rangle }}} \right\}$ & $\left\{ {{s_{2\left\langle {{o_{ - 3}}} \right\rangle }}} \right\}$ & $\left\{ {{s_{ - 1\left\langle {{o_0}} \right\rangle }}} \right\}$ & $\left\{ {{s_{ - 3\left\langle {{o_3}} \right\rangle }},{s_{ - 3\left\langle {{o_1}} \right\rangle }}} \right\}$ & $\left\{ {{s_{ - 3\left\langle {{o_3}} \right\rangle }},{s_{ - 1\left\langle {{o_{ - 3}}} \right\rangle }}} \right\}$ & $\left\{ {{s_{2\left\langle {{o_{ - 1}}} \right\rangle }},{s_{2\left\langle {{o_{ - 2}}} \right\rangle }}} \right\}$ \\
\hline
\end{tabular}
}
\end{table}

\begin{table}[htpb]
\centering
\renewcommand{\arraystretch}{2}%
\caption{The evaluation decision table of $E_3$}
\label{expert 3}
\hspace*{-2cm}
{\small
\begin{tabular}{p{0.3cm}p{2.3cm}p{2.5cm}p{2.5cm}p{2.5cm}p{2.2cm}p{2.4cm}}
\hline
   & ${a_1}$ & ${a_2}$ & ${a_3}$ & ${a_4}$ & ${a_5}$ & ${a_6}$ \\
\hline
${x_1}$ & $\left\{ {{s_{ - 2\left\langle {{o_0}} \right\rangle }}} \right\}$ & $\left\{ {{s_{ - 1\left\langle {{o_3}} \right\rangle }},{s_{ - 1\left\langle {{o_{ - 3}}} \right\rangle }}} \right\}$ & $\left\{ {{s_{1\left\langle {{o_{ - 2}}} \right\rangle }}} \right\}$ & $\left\{ {{s_{ - 3\left\langle {{o_2}} \right\rangle }}} \right\}$ & $\left\{ {{s_{ - 1\left\langle {{o_3}} \right\rangle }}} \right\}$ & $\left\{ {{s_{ - 3\left\langle {{o_2}} \right\rangle }},{s_{ - 3\left\langle {{o_0}} \right\rangle }}} \right\}$ \\

${x_2}$ & $\left\{ {{s_{1\left\langle {{o_0}} \right\rangle }}} \right\}$ & $\left\{ {{s_{0\left\langle {{o_0}} \right\rangle }}} \right\}$ & $\left\{ {{s_{1\left\langle {{o_1}} \right\rangle }},{s_{1\left\langle {{o_{ - 1}}} \right\rangle }}} \right\}$ & $\left\{ {{s_{1\left\langle {{o_0}} \right\rangle }},{s_{1\left\langle {{o_1}} \right\rangle }}} \right\}$ & $\left\{ {{s_{ - 1\left\langle {{o_{ - 3}}} \right\rangle }}} \right\}$ & $\left\{ {{s_{1\left\langle {{o_0}} \right\rangle }},{s_{1\left\langle {{o_{ - 2}}} \right\rangle }}} \right\}$ \\

${x_3}$ & $\left\{ {{s_{ - 2\left\langle {{o_2}} \right\rangle }}} \right\}$ & $\left\{ {{s_{ - 1\left\langle {{o_1}} \right\rangle }}} \right\}$ & $\left\{ {{s_{3\left\langle {{o_{-1}}} \right\rangle }}} \right\}$ & $\left\{ {{s_{3\left\langle {{o_{-2}}} \right\rangle }}} \right\}$ & $\left\{ {{s_{2\left\langle {{o_3}} \right\rangle }}} \right\}$ & $\left\{ {{s_{1\left\langle {{o_3}} \right\rangle }},{s_{2\left\langle {{o_2}} \right\rangle }}} \right\}$ \\

${x_4}$ & $\left\{ {{s_{ - 1\left\langle {{o_0}} \right\rangle }},{s_{ - 2\left\langle {{o_0}} \right\rangle }}} \right\}$ & $\left\{ {{s_{ - 2\left\langle {{o_3}} \right\rangle }}} \right\}$ & $\left\{ {{s_{2\left\langle {{o_{ - 1}}} \right\rangle }},{s_{1\left\langle {{o_1}} \right\rangle }}} \right\}$ & $\left\{ {{s_{ - 2\left\langle {{o_0}} \right\rangle }},{s_{ - 1\left\langle {{o_{ - 3}}} \right\rangle }}} \right\}$ & $\left\{ {{s_{1\left\langle {{o_{ - 3}}} \right\rangle }},{s_{1\left\langle {{o_1}} \right\rangle }}} \right\}$ & $\left\{ {{s_{2\left\langle {{o_{ - 3}}} \right\rangle }}} \right\}$ \\

${x_5}$ & $\left\{ {{s_{0\left\langle {{o_{ - 1}}} \right\rangle }}} \right\}$ & $\left\{ {{s_{0\left\langle {{o_1}} \right\rangle }}} \right\}$ & $\left\{ {{s_{0\left\langle {{o_1}} \right\rangle }}} \right\}$ & $\left\{ {{s_{0\left\langle {{o_1}} \right\rangle }},{s_{2\left\langle {{o_{ - 3}}} \right\rangle }}} \right\}$ & $\left\{ {{s_{ - 1\left\langle {{o_{ - 1}}} \right\rangle }}} \right\}$ & $\left\{ {{s_{1\left\langle {{o_{ - 3}}} \right\rangle }}} \right\}$ \\

${x_6}$ & $\left\{ {{s_{ - 1\left\langle {{o_2}} \right\rangle }}} \right\}$ & $\left\{ {{s_{ - 2\left\langle {{o_3}} \right\rangle }}} \right\}$ & $\left\{ {{s_{0\left\langle {{o_1}} \right\rangle }}} \right\}$ & $\left\{ {{s_{ - 3\left\langle {{o_0}} \right\rangle }}} \right\}$ & $\left\{ {{s_{ - 3\left\langle {{o_2}} \right\rangle }}} \right\}$ & $\left\{ {{s_{2\left\langle {{o_{ - 1}}} \right\rangle }}} \right\}$ \\

${x_7}$ & $\left\{ {{s_{0\left\langle {{o_1}} \right\rangle }}} \right\}$ & $\left\{ {{s_{ - 1\left\langle {{o_1}} \right\rangle }},{s_{ - 1\left\langle {{o_2}} \right\rangle }}} \right\}$ & $\left\{ {{s_{ - 2\left\langle {{o_0}} \right\rangle }},{s_{ - 2\left\langle {{o_{ - 2}}} \right\rangle }}} \right\}$ & $\left\{ {{s_{0\left\langle {{o_1}} \right\rangle }},{s_{1\left\langle {{o_3}} \right\rangle }}} \right\}$ & $\left\{ {{s_{ - 1\left\langle {{o_0}} \right\rangle }}} \right\}$ & $\left\{ {{s_{ - 3\left\langle {{o_0}} \right\rangle }}} \right\}$ \\

${x_8}$ & $\left\{ {{s_{ - 1\left\langle {{o_1}} \right\rangle }},{s_{ - 1\left\langle {{o_2}} \right\rangle }}} \right\}$ & $\left\{ {{s_{2\left\langle {{o_1}} \right\rangle }}} \right\}$ & $\left\{ {{s_{3\left\langle {{o_0}} \right\rangle }}} \right\}$ & $\left\{ {{s_{2\left\langle {{o_0}} \right\rangle }}} \right\}$ & $\left\{ {{s_{1\left\langle {{o_{ - 1}}} \right\rangle }}} \right\}$ & $\left\{ {{s_{2\left\langle {{o_2}} \right\rangle }}} \right\}$ \\
\hline
\end{tabular}
}
\end{table}

\subsection{S3W-GDM process of SLE dignosis}\label{5.2}
\begin{figure}[htbp]
  \centering
  \includegraphics[width=8cm]{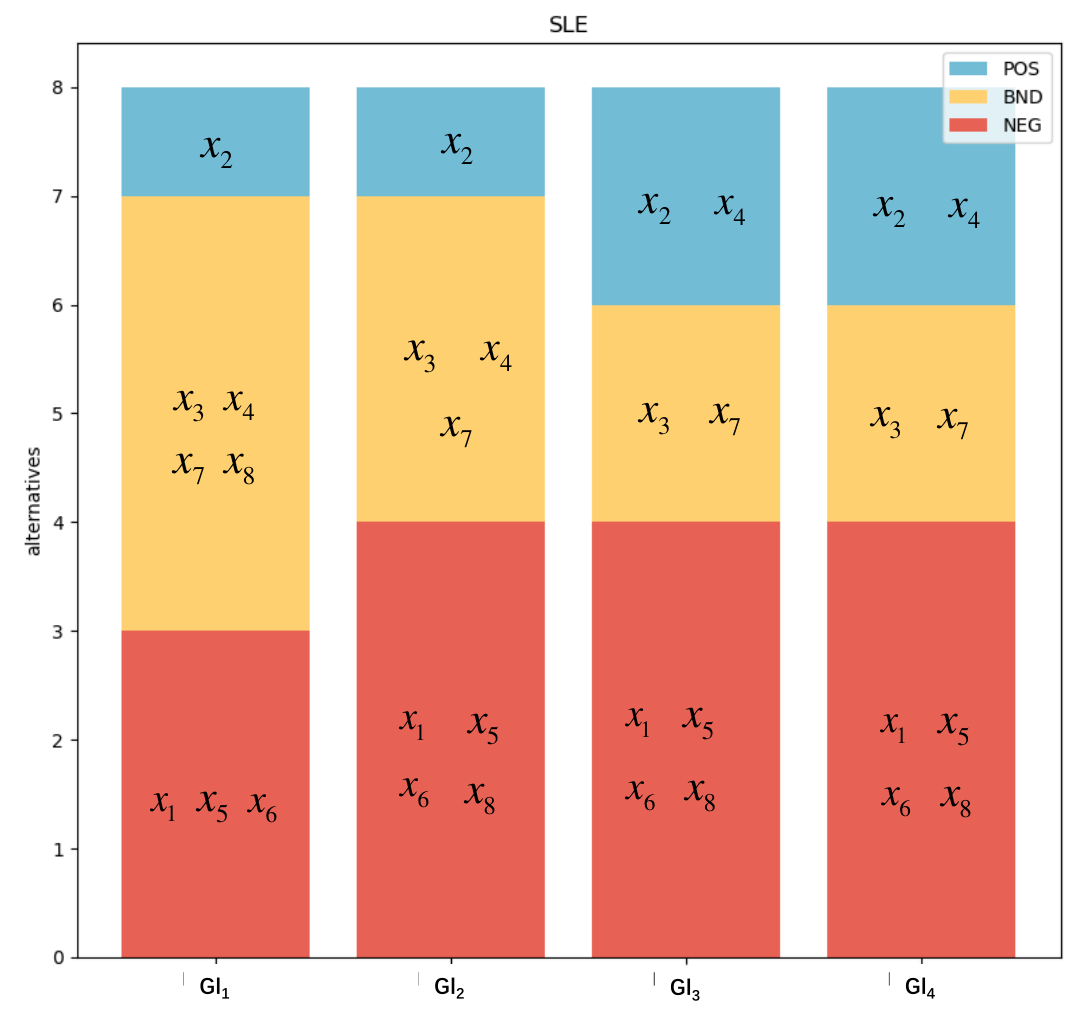}
  \caption{\textbf{.} The classification of SLE patients with different $G{l}$.}
  \label{SLEsexample}
\end{figure}

\begin{figure}[htbp]
\centering
       \subfigure{
                \begin{minipage}[t]{0.5\linewidth}
                \centering
                \includegraphics[height=4.2cm, width=6.5cm]{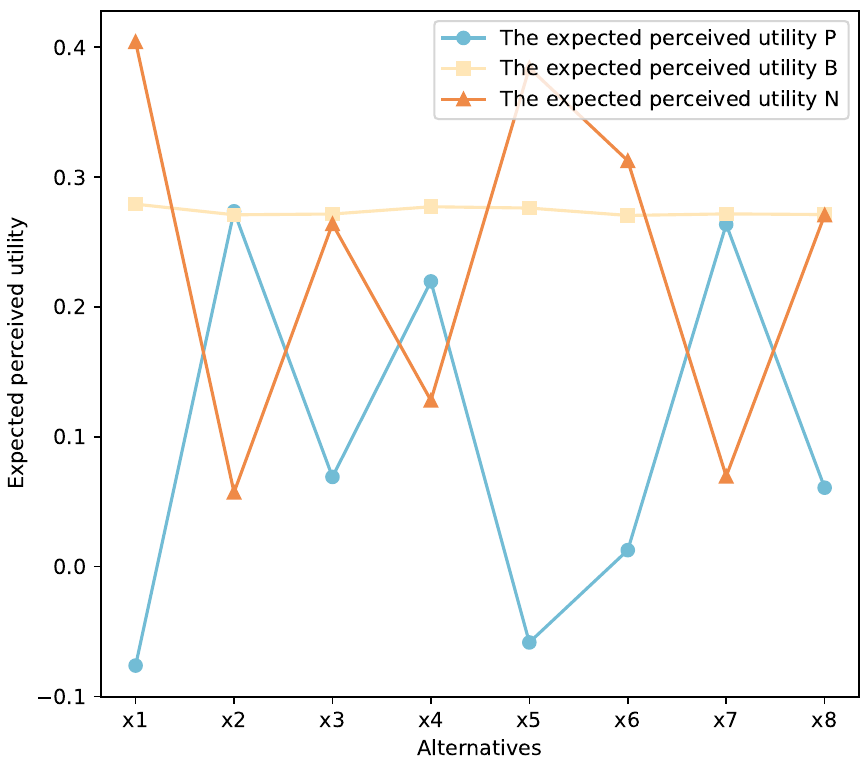}
                \small (a) The 1st decision-level.
                \end{minipage}%
                        }%
        \subfigure{
                \begin{minipage}[t]{0.5\linewidth}
                \centering
                \includegraphics[height=4.2cm, width=6.5cm]{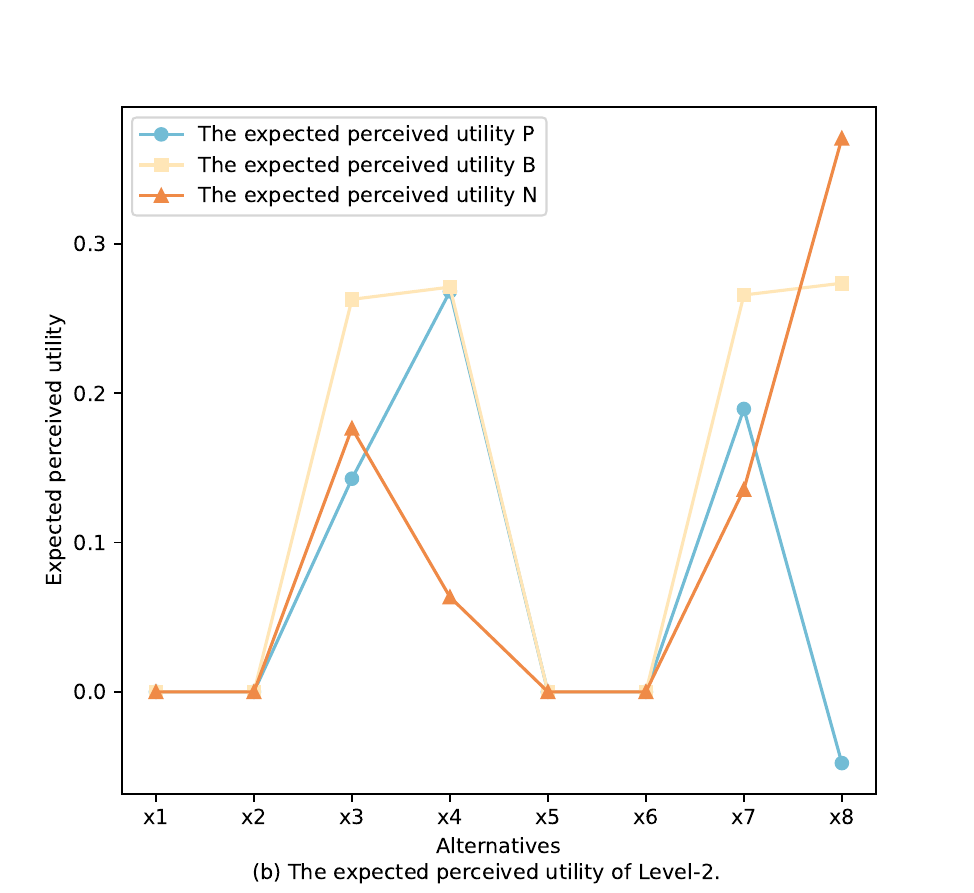}
                \small (b)  The 2nd decision-level.
                \end{minipage}
                        }
         \quad
\\
 \subfigure{
                \begin{minipage}[t]{0.5\linewidth}
                \centering
                \includegraphics[height=4.2cm, width=6.5cm]{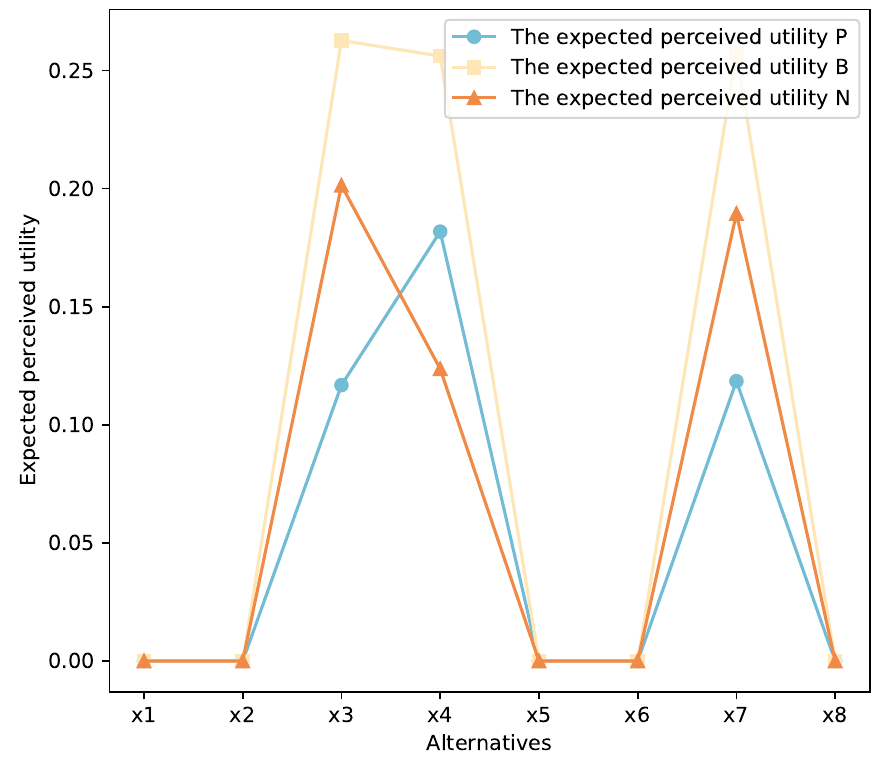}
                \small (c) The 3rd decision-level.
                \end{minipage}%
                        }%
        \subfigure{
                \begin{minipage}[t]{0.5\linewidth}
                \centering
                \includegraphics[height=4.2cm, width=6.5cm]{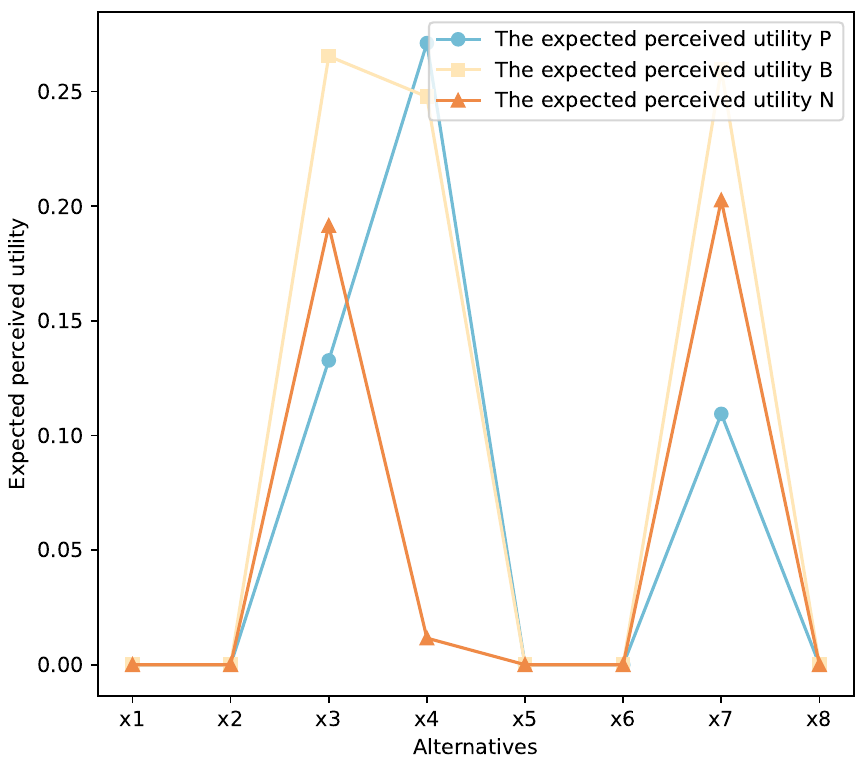}
                \small (d)  The 4th decision-level.
                \end{minipage}
                        }
\caption{\textbf{.} The expected perceived utility of alternatives at 4 decision-levels.}
\label{SLEsexample4}
\end{figure}

According to the contents of the previous two sections, the parameters involved in this illustration include the Gaussian kernel parameter $\sigma  = 0.7$, neighborhood cut parameter $\kappa  = 1$, relative gain parameter $\eta  = 0.6$, utility parameter $\theta  = 0.88$, and regret parameter $\delta  = 0.3$.
Reranking the conditional attributes $A = \left\{ {{a_1},{a_2},{a_3},{a_4},{a_5},{a_6}} \right\}$ based on weight vector $w = \left\{ {0.2,0.3,0.15,0.1,0.1,0.15} \right\}$, a sequence of conditional attribute subsets ${Z_1} = \left\{ {{a_2}} \right\}$, ${Z_2} = \left\{ {{a_1},{a_2}} \right\}$, ${Z_3} = \left\{ {{a_1},{a_2},{a_3},{a_6}} \right\}$, and ${Z_4} = \left\{ {{a_1},{a_2},{a_3},{a_4},{a_5},{a_6}} \right\}$ are obtained.
The S3W-GDM process will then be executed four times, once for each conditional attribute subset.
The calculation process is the same for each decision-level.
The calculation process for the $1st$ decision-level is illustrated.
\textbf{First}, the decision table under subset ${Z_1} = \left\{ {{a_2}} \right\}$ of conditional attributes from the three experts will be extracted at the $1st$ decision-level.
\textbf{Second}, the decision tables from these experts with the subset ${Z_1} = \left\{ {{a_2}} \right\}$ of conditional attributes will be aggregated to form the fused decision table.
\textbf{Then}, the granular structure $G{l_1} = \left\{ {{{\left( {DHHFLDT} \right)}_1},{\kappa _1},pr\left( {{\pi _1}\left| {{\kappa _{{\aleph _{{Z_1}}}}}\left( x \right)} \right.} \right),V_{{Z_1}}^{ \circ  \bullet }} \right\}$ will be obtained using the fused decision table.
\textbf{Finally}, the division results will be obtained by executing the S3W-GDM process to obtain the division of alternatives in the three domains of the $1st$ decision-level.
The decision will be computed sequentially following the above steps.
After the decisions are obtained for multi-level of the granular structure, a set of classification and the expected perceived utility of each alternative will be derived.
Fig. \ref{SLEsexample} and \ref{SLEsexample4} demonstrate the classification and ranking results of the $G{l_1}$, $G{l_2}$, $G{l_3}$, and $G{l_4}$ represent $1st$, $2nd$, $3rd$, and $4th$ decision-levels, respectively.

Fig. \ref{SLEsexample} clearly shows that the number of alternatives in the negative domain decreases as the decision-making progresses.
This trend conforms with the real decision-making scenarios.
In the positive and negative domains, the number of alternatives increases or remains unchanged as the decision-making progresses.
At the $1st$ decision-level, an initial determination can be made between patient $x_2$ and patients $x_1$, $x_5$ and $x_6$ based on the most important conditional attribute $a_2$.
Patient $x_2$ has a negative utility of $a_2$, indicating that her situation is not good.
Patients $x_1$, $x_5$, and $x_6$ do not have enough negative utility under $a_2$ to support them, so it is recommended that no further testing be done and monitoring is sufficient.
Those who handled it differently from them are patients $x_3$, $x_4$, $x_7$, $x_8$.
Further examinations are requested for these four patients since they may no longer be sufficient to receive a definitive medical guidance under $a_2$.
At the $2nd$ decision-level, the conditional attribute $a_1$ is added, providing a better decision-making basis. This makes it possible to determine patient$x_8$'s exact situation at this level.
Patient $x_8$ only needs further monitoring and does not require additional tests for the time being, simplifying her care.
Unfortunately, Patient $x_4$ cannot receive a diagnosis at the $2nd$ decision-level.
However, at the $3rd$ decision-level, a new diagnosis is conducted.
She is suffering from SLE, and she is referred for precise treatment.
This scenario demonstrates that the decision model closely aligns with real-life medical decision-making scenarios, making it a practical tool for such applications.
Given limited healthcare resources and varying degrees of patient conditions, physicians often face challenging situations.
By determining the severity of a patient's condition, a personalized, patient-centered treatment plan can be developed based on their specific needs.
This approach helps optimize doctors' work and makes the most efficient use of medical resources.
The model outlined in this paper presents an application scenario where different alternatives can be prioritized based on the severity of patients' illnesses.

Fig.  \ref{SLEsexample4} is the basis for the plotting of Fig. \ref{SLEsexample}, as well as the judgment of the severity of the disease in the eight patients.
Fig.  \ref{SLEsexample4} presents the trend of the expected perceived utility values for the eight patients.
According to the decision rule in Section \ref{S3WD}, the final ranking result is obtained: ${x_2} > {x_4} > {x_7} > {x_3} > {x_1} > {x_5} > {x_8} > {x_6}$,which can be used to determine the severity of illness for eight patients.
Patient $x_2$ and patient $x_4$ are already diagnosed at the $4th$ decision-level,
Among them, patient $x_2$'s condition is worse, as the condition attribute used in this case has a larger value, indicating that the target concept is closer to being sick.
For patients $x_3$ and $x_7$, new tests are required to determine if they are sick.
Although the ranking rules suggest that patient  $x_3$ may be at a higher risk of being sick, there is still significant uncertainty in the ordering of the boundary domain.
While patients $x_1$, $x_5$, $x_6$, and $x_8$ are not diagnosed based on the evaluation of these six conditional attributes, it is important to monitor them regularly and maintain good follow-up records.
During the experiments, it was found that the ranking results obtained at the end of the $1st$ decision-level are highly correlated with the above ranking results.
The turning point that leads to a different ranking result is the patient $x_8$ in the boundary domain.
\begin{table}[h]
\centering
\renewcommand{\arraystretch}{1.4}%
\setlength{\abovecaptionskip}{1.5pt}%
\captionsetup{justification=centering, singlelinecheck=false, labelsep=quad}
\caption{The ranking results of different decision-levels.}
\label{Comparison of the ranking}
\setlength{\tabcolsep}{3mm}{
\begin{tabular}{cc}
\hline
         Decision-level  & The ranking orders \\ \hline
$1st$   &  ${x_2} > {x_4} > {x_7} > {x_3} > {x_8} > {x_1} > {x_5} > {x_6}$  \\
$2nd$   & ${x_2} > {x_4} > {x_7} > {x_3} > {x_1} > {x_5} > {x_8} > {x_6}$  \\
$3rd$  & ${x_2} > {x_4} > {x_7} > {x_3} > {x_1} > {x_5} > {x_8} > {x_6}$  \\
$4th$  & ${x_2} > {x_4} > {x_7} > {x_3} > {x_1} > {x_5} > {x_8} > {x_6}$  \\ \hline
\end{tabular}
}
\end{table}
In subfigure (a) of Fig. \ref{SLEsexample4}, it is evident that the conditional attribute $a_2$ has a more significant impact on the expected perceived utility, whereas subfigures (b), (c), and (d) exhibit a gradual stabilization as new conditional attributes are included.
This implies that, in this case, the conditional attribute $a_2$ has the most crucial influence on the decision results.

From Table \ref{Comparison of the ranking}, the changes in ranking from the $1st$ decision-level to the $4th$ decision-level can be observed.
The ranking results of $1st$ decision-level and $2nd$ decision-level is broadly consistent, but slightly different.
At $3rd$ decision-level of decision-making, the ranking remains the same, with very little change.
By $4th$ decision-level, the ranking of all alternative objects is stabilized.
These results indicate that the initial ranking achieved at the $1st$ decision-level, under coarse granularity, is nearly identical to the final ranking at the $4th$ decision-level.
This suggests that utilizing such a granular structure significantly enhances decision-making efficiency, as relative accurate results can be achieved even at the initial, less detailed levels.

The multi-level S3W-GDM method also provides a more reasonable semantic interpretation of the classification and ranking results for the above example.
In clinical diagnosis, the indicator based on the $a_2$ Degree of Anti-dsDNA is important for preliminary screening and rapid analysis.
Patient 2, as the most serious patient under this indicator, is screened out first to receive relevant treatment without delaying her condition.
Although each decision-level can produce a relevant ranking, it is still unclear which patients need additional examinations.
The boundary region provide the interpretive space for this process.
The decreasing number of boundary region objects and the stabilization of positive and negative region classifications reflect the model's ability to provide meaningful and interpretable decision results.
This aligns with the practical need for iterative and adaptive decision-making in medical diagnostics, where patient conditions and available information evolve over time.

\section{Comparative and sensitivity analysis}\label{section6}
The efficiency and rationality of the multi-level S3W-GDM method are verified through the comparative and sensitivity analysis.
In Section \ref{comparativeanalysis}, the efficiency and validity of the S3W-GDM method are verified using four types of data sets and seven other methods.
Additionally, the rationality of the proposed method is demonstrated by analyzing the results generated from the combination of parameters in Section \ref{sensitiveanalysis}.

In the previous sections, a novel S3W-GDM method is proposed to solve the GDM problem under DHHFLTS environment.
This decision-making process is more closely aligned with the diagnostic process in reality.
To validate the efficiency and validity of the proposed information fusion method incorporating granular computing, it is necessary to compare it with the established decision-making methods for different DHHFLTS methods.
Before making the comparison, the selected comparison methods and comparison data sets need to be explained.

(1) Regarding the comparative methods, some classical DHHFLTS methods are selected for comparison, and their differences are shown in Table \ref{methodsforcomparative}.
\begin{table}[htbp]
\centering
\renewcommand{\arraystretch}{1.2}%
\setlength{\abovecaptionskip}{1.5pt}%
\captionsetup{justification=centering, singlelinecheck=false, labelsep=quad}
\caption{Some methods for comparative analysis.}
\label{methodsforcomparative}
\setlength{\tabcolsep}{0.1mm}{
\begin{tabular}{lccccc}
\hline
Methods & Single expert  & Multiple expert & Fusion mode & Ranking & Classification\\ \hline
Method 1 \cite{krishankumar2019framework} &$\surd$  &$\surd$ &Traditional aggreagtion &$\surd$  &$\backslash$ \\
Method 2 \cite{zhang2021integrated} &$\surd$  &$\surd$ &Dempster-Shafer evidence &$\surd$  &$\backslash$ \\
S3W-GDM &$\surd$  &$\surd$ &Multi-level granularity  &$\surd$  &$\surd$  \\
Method 3 \cite{liu2023improved} &$\surd$  &$\backslash$ &$\backslash$ &$\surd$  &$\backslash$  \\
Method 4 \cite{liu2019novel} &$\surd$  &$\backslash$ &$\backslash$ &$\surd$  &$\backslash$  \\
Method 5 \cite{gou2018multiple}  &$\surd$  &$\backslash$ &$\backslash$ &$\surd$  &$\backslash$  \\
Method 6 \cite{GOU2017multimoora}  &$\surd$  &$\backslash$ &$\backslash$ &$\surd$  &$\backslash$  \\
Method 7 \cite{gou2021probabilistic}  &$\surd$  &$\backslash$ &$\backslash$ &$\surd$  &$\backslash$  \\
\hline
\end{tabular}
}
\end{table}
Method 1 and 2, as well as the S3W-GDM methods in this paper, focus on the comparison in dealing with multiple experts GDM problems.
The difference is that Method 1 used a traditional aggregation operator that fused information from multiple experts evaluations \cite{krishankumar2019framework}.
Method 2 employed the Dempster-Shafer evidence theory and proposed a novel method for fusing experts' information \cite{zhang2021integrated}.
Method 3 \cite{liu2023improved} addresses a multi-attribute group decision problem involving the selection of emergency logistics providers.
Methods 4-7 are traditional single expert decision-making methods, including the double hierarchy hesitant fuzzy linguistic generalized power average operator \cite{liu2019novel}, TOPSIS-based the generalized completely hybrid weighted Hausdorff-hesitance degree-based distance \cite{gou2018multiple}, MULTIMOORA \cite{GOU2017multimoora}, and VIKOR \cite{gou2021probabilistic}.
Method 4-7 are chosen to validate that the proposed methods are equally applicable to single-expert decision-making with better efficiency.

(2) Regarding the four types of comparative data sets, these evaluation data sets under three different perspectives are selected, as shown in Table \ref{comparativesources}.
The data set for the selection of financial products from Method 2 \cite{zhang2021integrated} is a regular GDM data set with multiple experts evaluating multiple alternatives under different attributes.
The breast cancer data set is also used for this comparison to further demonstrate the efficiency gains of the proposed method compared to the regular GDM methods.
The data is preprocessed and set up as a GDM problem with 3 experts and 30 alternatives.
The use of this data set with more alternatives allows for the validation of the proposed method in more complex scenarios.
Methods 1, 2 and the S3W-GDM method are applicable to both of two data sets for GDM.
The evaluation data set from literature \cite{liu2023improved} involves a multi-attribute decision-making problem with emergency logistics provider selection.
The novel S3W-GDM method proposed in this paper is designed to address the dynamics in GDM problems, making it compatible with the context of this evaluation data set.
Since the model proposed in this paper is to a greater extent a dynamic decision-making model, it is different from the stable decision-making where all the decision-making information is collected at once.
The effectiveness of the method in this paper is demonstrated by comparing it with the dynamic GDM problem.
Since there is only one decision table for this data set, some of the classical Methods 3-7 are compared with the  proposed method in this paper.
The evaluation data set from literature \cite{gou2018multiple} consists of reviews on Sichuan liquor brands.
The selection of these brands is based on their popularity among the general public, allowing for a more factual and rational evaluation of the decision results.
The comparison method chosen for this data set is a stationary multi-attribute decision-making Method 4-7.

\begin{table}[htbp]
\centering
\renewcommand{\arraystretch}{1.4}%
\setlength{\abovecaptionskip}{1.5pt}%
\captionsetup{justification=centering, singlelinecheck=false, labelsep=quad}
\caption{Four types of comparative data sets.}
\label{comparativesources}
\setlength{\tabcolsep}{2mm}{
\begin{tabular}{cccc}
\hline
Types &  \makecell[c]{Background of  data set}& Reasons for selection & Comparative methods \\ \hline
Type 1 &  \makecell[c]{The selection of \\ financial products} & Regular GDM  & \makecell[c]{Method 1 \cite{krishankumar2019framework},\\ Method 2 \cite{zhang2021integrated},\\ S3W-GDM}\\
\hline
Type 2 &  \makecell[c]{Diagnosis of \\breast cancer} & \makecell[c]{GDM with \\more alternatives}  & \makecell[c]{Method 1 \cite{krishankumar2019framework},\\ Method 2 \cite{zhang2021integrated},\\ S3W-GDM}\\
\hline
Type 3 &  \makecell[c]{Emergency logistics \\provider selection}& \makecell[c]{Dynamics of\\ decision-making}  & \makecell[c]{Method 3 \cite{liu2023improved},\\ Method 4 \cite{liu2019novel},\\Method 5  \cite{gou2018multiple},\\ Method 6 \cite{GOU2017multimoora},\\Method 7 \cite{gou2021probabilistic},\\S3W-GDM }\\
\hline
Type 4 &  \makecell[c]{The assessment of \\Sichuan liquor brand}& Easily identifiable  & \makecell[c]{Method 4 \cite{liu2019novel},\\Method 5  \cite{gou2018multiple},\\ Method 6 \cite{GOU2017multimoora},\\Method 7 \cite{gou2021probabilistic},\\S3W-GDM }\\
\hline
\end{tabular}
}
\end{table}

\subsection{Comparative analysis}\label{comparativeanalysis}

\textbf{Type 1.  Comparison of financial products selection}

This evaluation data set contains 6 different financial products $U = \left\{ {{x_1},{x_2},{x_3},{x_4},{x_5},{x_6}} \right\}$ and requires 3 experts to select a financial product by considering a combination of 4 dimensions: rate of return($a_1$), risk($a_2$), liquidity($a_3$), and tansparency($a_4$).
The conditional attribute weight vector $w = \left\{ {0.23,0.13,0.52,0.12} \right\}$ determined through BWM will be used here.
Method 2 treats decision information as evidence, with experts viewed as different sources of evidence.
The BPA function is constructed by calculating the confidence and evidence matrix for each piece of decision information, and the ranking of financial products is obtained by combining the DSET rule.
This process is quite tedious.
To combine the decision-making information of all experts under various conditional attributes, it is necessary to calculate different distances between experts, products, and attributes.
This computation is required before constructing the global BPA function, which introduces DSET rules.
Additionally, the different number of DHLTs in DHHFLEs need to be normalized.

As shown in Table \ref{finacialproducts}, the ranking results of these methods are almost the same.
The main difference is that the top two rankings of Method 1 are different from the other methods.
In Method 1, $x_1$ has an advantage over $x_3$, while all other alternatives maintain the same ranking as the remaining methods.
The last two ranking results are from the method proposed in this paper.
One ranking result uses only one sequential decision process, and the other completes the computation of all subsets of conditional attributes.
However, the final ranking results obtained are the same as Method 2.
The $1st$ decision-level delineated the $\eta=0.7$ partition in which the product is located.
Specifically, the novel multi-level S3W-GDM method can achieve the same results after the $1st$ decision-level.
The S3W-GDM method with multi-granularity thinking greatly enhances the efficiency of the GDM process.

\begin{table}[htbp]
\centering
\renewcommand{\arraystretch}{1.4}%
\setlength{\abovecaptionskip}{1.5pt}%
\captionsetup{justification=centering, singlelinecheck=false, labelsep=quad}
\caption{The ranking of financial products selection by different methods.}
\label{finacialproducts}
\setlength{\tabcolsep}{3mm}{
\begin{tabular}{cc}
    \hline
    Method   & Ranking \\
    \hline
    Method 1 \cite{krishankumar2019framework} &  ${x_1} > {x_3} > {x_5} > {x_4} > {x_2}> {x_6}$\\
    Method 2 \cite{zhang2021integrated} & ${x_3} > {x_1} > {x_5} > {x_2} > {x_4}> {x_6}$\\
    S3W-GDM after the $1st$ decision-level&  ${x_3} > {x_1} > {x_5} > {x_2} > {x_4}> {x_6}$\\
    S3W-GDM after the $4th$ decision-level&  ${x_3} > {x_1} > {x_5} > {x_2} > {x_4}> {x_6}$\\
    \hline
    \end{tabular}
}
\end{table}

\textbf{Type 2. Comparison of breast cancer diagnosis}

The Breast Cancer Coimbra Data Set (https://archive.ics.uci.edu/datasets) as row data is adapted to demonstrate the computational efficiency of the method for GDM scenarios.
Suppose the weight vector of conditional attribute is $w = \left\{ {0.07, 0.15, 0.28, 0.32, 0.34} \right\}$, the DHLTS with the first hierarchy linguistic term scale is $S = \left\{ {{s_{ - 2}} = very\;low,{s_{ - 1}} = low,{s_0}} \right.$\\$ = normal,\left. {{s_1} = high,{s_2} = very\;high} \right\}$, and the second hierarchy linguistic term scales includes ${O_1} = \left\{ {{o_{ - 2}} = only\;a\;little,{o_{ - 1}} = a\;little,{o_0} = just\;right,{o_1} = \;much,{o_2} = very\;much} \right\}$, ${O_2} = $$\left\{ {{o_{ - 2}} = only\;a\;little,{o_{ - 1}} = a\;little,{o_0} = just\;right} \right\}$, ${O_3} = \left\{ {{o_{ - 2}} = very\;much,{o_{ - 1}} = } \right.$\\$\left. {much,{o_0} = just\;right,{o_1} = \;only\;a\;little,{o_2} = little} \right\}$, ${O_4} = \left\{ {{o_{ - 2}} = very\;much,{o_{ - 1}} = much,} \right.$\\${o_0} = \left. {just\;right} \right\}$.
Three medical experts are invited to convert the crisp numbers of conditional attributes (BMI, Glucose, Insulin, Leptin, and Adiponectin) into DHHFLTS based on the linguistic term scales above.
These three experts have the same level of importance.
Only the data of the first 30 patients are required to be transformed by the experts.

\begin{figure}[htbp]
  \centering
  \includegraphics[width=10cm]{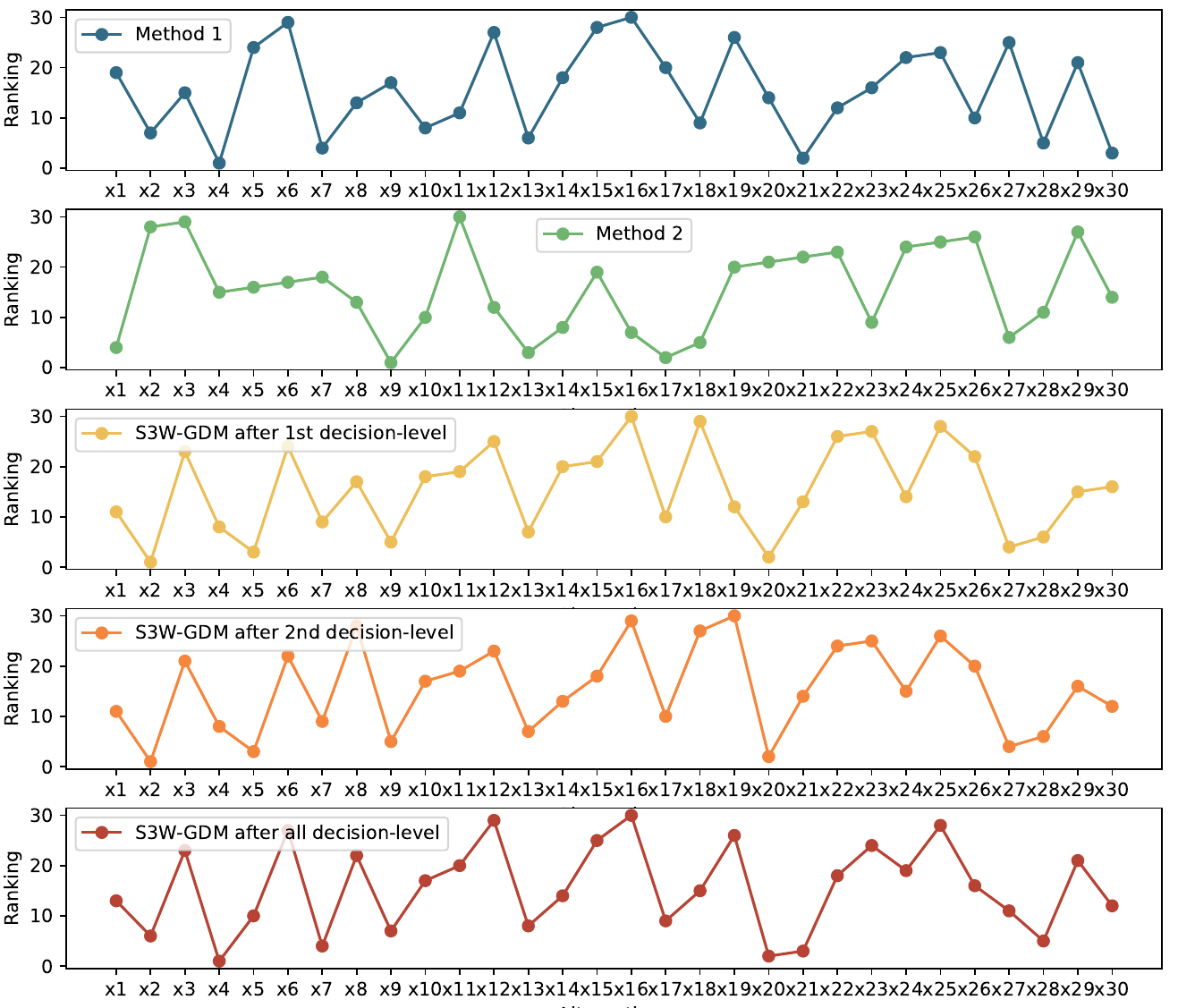}
  \caption{\textbf{.} The ranking comparison between S3W-GDM and others GDM methods.}
  \label{BCCDcomparision}
\end{figure}
In Fig. \ref{BCCDcomparision}, Method 1 and Method 2 are used as two different types of GDM methods for the comparison of program ranking.
The proposed method in this paper has 30 alternatives divided and completed the ranking after the $2nd$ decision-level.
The presented ranking results have a similar trend to the other two methods, while only $1st$ and $2nd$ decision-levels of S3W-GDM of computation are used, greatly improving the efficiency of the decision-making process.
However, the difference is due to the fact that the two compared methods use more comprehensive decision-making information, while the proposed method only uses partial information.
Therefore, all conditional attributes were added using the same parameters to execute the algorithm.
As can be observed from the bottom subplot in Fig. \ref{BCCDcomparision} the decision on the alternative is more rational after using the information from all the conditional attributes.
In particular, Method 1 shows a trend more similar to the method proposed in this paper in terms of the ranking of some alternatives.
For GDM problems with a large number of alternatives, it is more rational and intuitive to perform further decision-making by classifying the results of the alternatives.
Methods 1 and 2 lack a relevant process that brings semantic interpretation to the decision results.
The S3W-GDM method fills such a gap.
The specific classification of this data set will be shown in the sensitive analysis.

\textbf{Type 3. Comparison of emergency logistics provider selection}

This evaluation data set contains 5 emergency logistics providers in the food and beverage industry $U = \left\{ {{x_1},{x_2},{x_3},{x_4},{x_5}} \right\}$ and 6 evaluation attributes: cost($a_1$), product level($a_2$), quick response ability of supply($a_3$), quick response ability of transport($a_4$), management($a_5$), and reputation($a_6$).
The attribute weight results of optimization model-based distance under DHHFLTS environment is used in the comparison here to eliminate the impact might happen.
The vector of weight is $w = \left\{ {0.1011,0.1017,0.2591,0.1305,0.165,0.2426} \right\}$.

Table \ref{comparisionemergency} shows a comparison of the ranking results of the several methods under this data set.
Method 3 applied the normalized projection-based distance and bidirectional projection to DHHFLTS.
These improvements in the distance measurements bring about better superiority and rationality in the ranking results.
This ranking result is consistent with the traditional Methods 4-6. Method 7, as well as the methods proposed in this paper, are then consistent.
The difference lies mainly in the ranking of alternatives $x_1$, $x_3$, $x_5$.
However, there is a lack of a dynamic decision-making process and quick initial judgment for the characteristics of emergency decision-making.
It is well known that emergency decision-making has a higher demand for efficiency.
The multi-level S3W-GDM provides a more conclusive semantic interpretation after the completion of the $3rd$ decision-level ranking.
Although there are differences from most models, $x_4$ as the best supplier is reflected in the positive domain of the partition where it is located.
It is worth noting that the method proposed in this paper uses only half of the decision information at this decision-level.
For emergency decision-making scenarios, S3W-GDM provides a priori a solution as a decision-making result, supporting the rapid development of the action.

\begin{table}[htbp]
\centering
\renewcommand{\arraystretch}{1.4}%
\setlength{\abovecaptionskip}{1.5pt}%
\captionsetup{justification=centering, singlelinecheck=false, labelsep=quad}
\caption{The ranking of logistics provider selection by different methods.}
\label{comparisionemergency}
\setlength{\tabcolsep}{3mm}{
\begin{tabular}{cc}
    \hline
    Method   & Ranking \\
    \hline
    Method 3    &  ${x_4} > {x_2} > {x_5} > {x_1} > {x_3}$\\
    Method 4    &  ${x_4} > {x_2} > {x_5} > {x_1} > {x_3}$\\
    Method 5    &  ${x_4} > {x_2} > {x_5} > {x_1} > {x_3}$\\
    Method 6    &  ${x_4} > {x_2} > {x_5} > {x_1} > {x_3}$\\
    Method 7    &  ${x_4} > {x_2} > {x_1} > {x_3} > {x_5}$\\
    S3W-GDM after the $3rd$ decision-level &  ${x_4} > {x_1} > {x_3} > {x_2} > {x_5}$\\
    S3W-GDM after the $4th$ decision-level &  ${x_4} > {x_2} > {x_1} > {x_3} > {x_5}$\\
    \hline
    \end{tabular}
}
\end{table}

\textbf{Type 4.  Comparison of Sichuan liquor brand assessment}

This evaluation data set contains 5 Sichuan liquor brands: Wuliangye($x_1$), Luzhou Old Cellar($x_2$), Ichiro liquor ($x_3$), Tuopai liquor($x_4$) and Jian Nan Chun($x_5$).
The cognitions of consumers are used as a starting point to investigate four attributes: product price($a_1$), product classification($a_2$), consumer group($a_3$), and distribution channel($a_4$).
The attributes weights vector is $w = \left\{ {0.1,0.3,0.2,0.4} \right\}$.

From the Table \ref{liquorbrand}, these methods can be used to examine the liquor brand data set to improve the rationality of the decision results.
In the analysis of different methods' rankings of alternatives, it is observed that Method 4-6 display identical ranking patterns, suggesting similarities in their evaluation criteria.
Method 7 and S3W-GDM provide an alternative ranking result reveal that these methods may employ different decision-making logics or prioritize differently.
S3W-GDM presents a ranking almost similar to that of Method 7, both after the computation of all attribute subsets has been considered and only after the completion of the $3rd$ decision-level.
The only difference is the position of alternative $x_4$.
The ranking results obtained by the S3W-GDM method after the $4th$ decision-level differ from Methods 4-6 in the order of alternatives $x_1$ and $x_5$.
Methods 4-6 tend to prefer Wuliangye($x_1$) as the top alternative, indicating its widespread acceptance, while the varying rankings of Jian Nan Chun($x_5$) reflect significant differences in evaluations across methods.
The reason for the differences in the methods proposed in this paper goes back to the setup of this data set itself.
This evaluation comes from consumers' perceptions of Sichuan liquor brands.
Wuliangye($x_1$) is well known as a high-end brand.
However, Jian Nan Chun($x_5$), a mid-to-high end brand, is currently showing a rapid growth trend, becoming the ``meat and potatoes" of the liquor market.
On the one hand, its price and grade are more in line with the rational consumption concepts of young people.
On the other hand, the occupation position in the market rises higher than the space of the low-end categories.
Evaluating the condition attributes of Sichuan liquor brand, the largest weight is the distribution channel($a_4$), and Jian Nan Chun($x_5$) does have better distribution channels, gradually becoming the best occupied brand in the mid-to-high-end market.
Based on the perspective of distribution channels, the findings of the S3W-GDM method should be of better reference value in providing adjustment strategies for Sichuan liquor enterprises.

\begin{table}[htbp]
\centering
\renewcommand{\arraystretch}{1.4}%
\setlength{\abovecaptionskip}{1.5pt}%
\captionsetup{justification=centering, singlelinecheck=false, labelsep=quad}
\caption{The ranking of liquor brand by different methods.}
\label{liquorbrand}
\setlength{\tabcolsep}{3mm}{
\begin{tabular}{cc}
    \hline
    Method   & Ranking \\
    \hline
    Method 4        &  ${x_1} > {x_5} > {x_3} > {x_4} > {x_2}$\\
    Method 5     &  ${x_1} > {x_5} > {x_3} > {x_4} > {x_2}$\\
    Method 6  &  ${x_1} > {x_5} > {x_3} > {x_4} > {x_2}$\\
    Method 7 &  ${x_5} > {x_1} > {x_3} > {x_2} > {x_4}$\\
    \makecell[c]{S3W-GDM after the $3rd$ decision-level} &  ${x_5} > {x_1} > {x_4} > {x_3} > {x_2}$\\
    \makecell[c]{S3W-GDM after  the $4th$ decision-level} &  ${x_5} > {x_1} > {x_3} > {x_4} > {x_2}$\\
    \hline
    \end{tabular}
}
\end{table}

\subsection{Sensitivity analysis}\label{sensitiveanalysis}
The subsection will display how parameter variation affects the decision results.
sensitivity of Breast Cancer Coimbra Data Set.
For presentation purposes, the Breast Cancer Coimbra Data Set with more alternatives is used here as the sensitivity analysis.
The main study is the variation of the Gaussian kernel parameter $\sigma$ and the neighborhood cut parameter $\kappa  $ for different relative gain parameters $\eta $.
In Introduction \ref{Introduction} and Section \ref{3WD model}, extensive discussion has taken place regarding the parameter of relative gains.
The typical value range for this parameter is [0, 1].
In this data set, the experiments are conducted through varying $\eta $ from 0 to 1 with an interval of 0.1.
The conclusion is that when $\eta  \le 0.5$, all alternatives are completely classified into the positive and negative regions at the $1st$ decision-level, which is evidently unreasonable.
Given the large number of alternatives, although the proposed method aims to enhance decision efficiency, the limited decision information received at the $1st$ decision-level results in significant errors if classification is based solely on the first most important conditional attribute.
Consequently, this study does not consider $\eta  \le 0.5$ for this data set.
When $\eta $ is 0.6, although the classification of alternatives begins to show a general pattern, it remains unstable with variations in  $\sigma$ and $\kappa  $, and subsequential changes do not follow a consistent pattern.
When $\eta  \ge 0.7$, the classification of alternatives stabilizes, and the subsequent variations in $\sigma$ and $\kappa  $ conform to the discussions in Section \ref{S3WD}.

\begin{figure}[h]
    \centering
    \begin{minipage}[b]{0.3\textwidth}
        \centering
        \includegraphics[width=\textwidth]{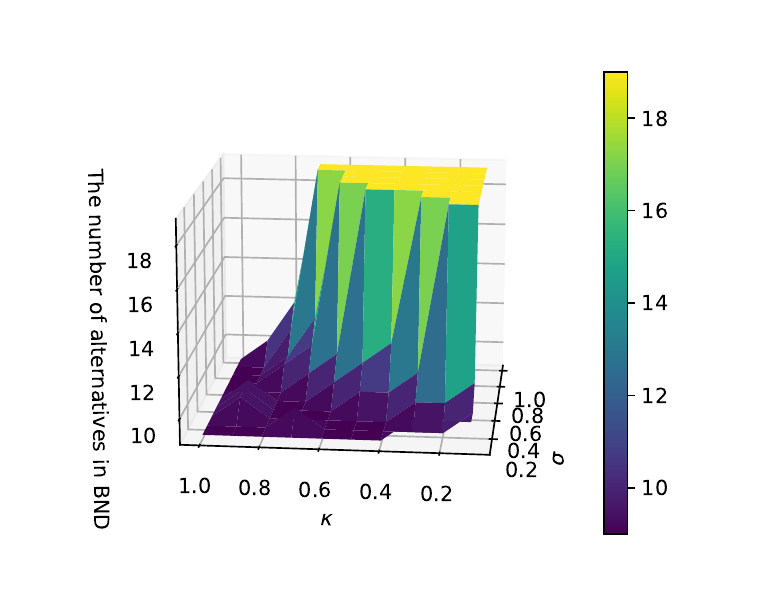}
        \small (a)  $\eta = 0.7$.
    \end{minipage}
    \hspace{0.02\textwidth} %
    \begin{minipage}[b]{0.3\textwidth}
        \centering
        \includegraphics[width=\textwidth]{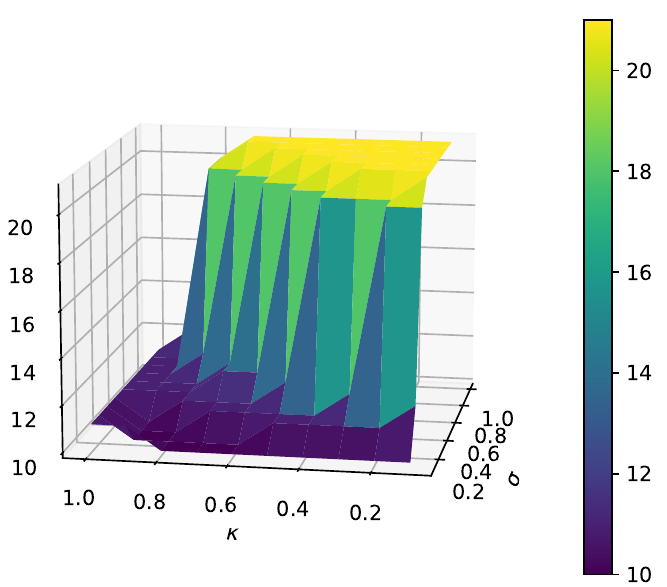}
        \small (b)  $\eta = 0.8$.
    \end{minipage}
    \hspace{0.02\textwidth} %
    \begin{minipage}[b]{0.3\textwidth}
        \centering
        \includegraphics[width=\textwidth]{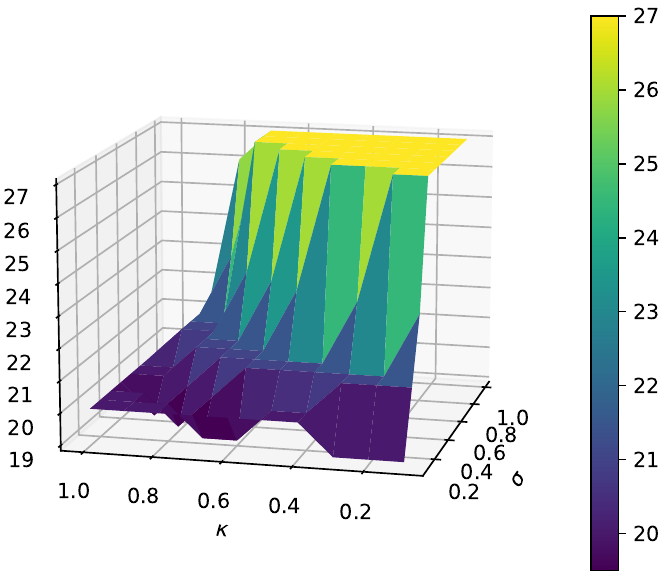}
        \small (c)  $\eta = 0.9$.
    \end{minipage}
    \caption{\textbf{.} Distribution of decision-level with the variation of parameters.}
\label{SENSITIVE ANALYSIS}
\end{figure}

The purpose of this work is to observe $\sigma$ and $\kappa  $ that are related to the sequential process, namely the gaussian kernel parameter and the neighborhood cut parameter with the relative gain parameter of 0.7, 0.8, and 0.9.
Then the interval in which these two parameters are observed is [0,1], with a step size of  0.1.
Fig. \ref{SENSITIVE ANALYSIS} provides a detailed illustration of the variations in $\sigma$ and $\kappa  $.
Subfigures (a), (b), and (c) demonstrate the variations in the number of boundary region alternatives as a function of the combination of $\sigma$ and $\kappa  $.
Even with different parameter settings, the boundary region alternatives exhibit a stable pattern at this decision-level.
Notably, the closer the combination of $\sigma$ and $\kappa  $ approaches (1, 1), the greater the number of alternatives in the boundary region.
This indicates that the decision conditions become stricter, aligning with the semantic interpretation of variations in these two parameters.

\begin{figure}[htbp]
  \centering
  \includegraphics[width=12cm]{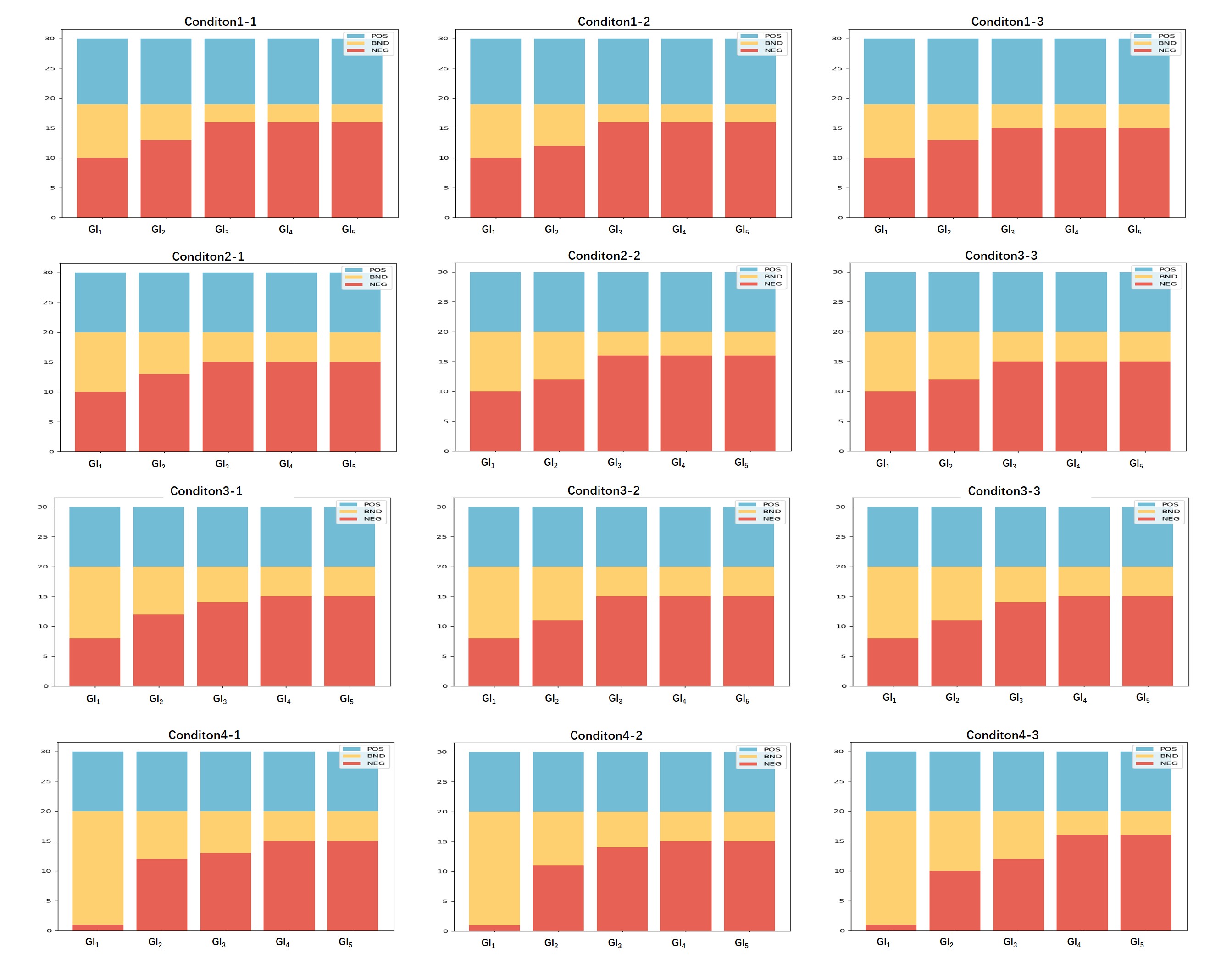}
  \caption{\textbf{.} Classification results under different combinations of parameters.}
  \label{BCCDclassification}
\end{figure}
Next, to control for variables, the classification results of 30 patients under $\eta  = 0.7$ are presented in Fig. \ref{BCCDclassification}.
This is to illustrate the classification trends under different parameter settings.
Fig. \ref{BCCDclassification} illustrates the classification of alternatives under four different combinations of $\sigma$ and $\kappa  $.
The four sets of parameter combinations are (0.9,0.9), (0.8,0.8), (0.8,0.7), and (0.9,0.7).
Each parameter combination results in distinct classification trends.
This is due to the ability to adjust and vary $\sigma$ and $\kappa  $ at each decision-level.
In this study, the value of $\sigma$ is fixed to maintain sensitivity in the calculations.
By adjusting the changes in $\kappa  $ at each decision-level, ensuring that the $\kappa  $ values gradually increase, the accuracy of the decision results is ensured, leading to diverse classification results while maintaining the overall trend.
In each set of sequential processes, $\kappa  $ in this way changes at the decision-level in the interval [0.8,1].
Regarding the properties of these two parameters, a $\sigma$ value closer to 1 results in a more sensitive Gaussian kernel function, while a $\kappa$ value closer to 1 indicates stricter equivalence division among alternatives.
Both settings contribute to the accuracy of the final classification results.

In general, the yellow area tends to shrink as the decision-making process progresses. Conversely, the blue and red areas may either remain constant or expand as the decision-making stages advance.
These two performance characteristics accurately reflect the actual decision-making situation. Furthermore, the more alternatives that are divided, the more varied the results will be. Different parameters can be adjusted to achieve different outcomes based on the specific decision-making scenario.
The model's classification rationality is demonstrated through sensitivity analysis.

\subsection{Discussion}
Since the method proposed in this paper is a dynamic decision-making model to a greater extent, unlike static decision-making which collects all the decision-making information at once,  mutli-level S3W-GDM makes decisions by increasing the granularity of the decision-making information in a level-by-level progression, which on the one hand improves the efficiency of decision-making, and on the other hand provides a buffer to reduce the risk of erroneous decision-making when the decision-making information is insufficient to support the decision.

The differences between the proposed method and others are listed as follows:

(1) Classical GDM methods \cite{krishankumar2019framework,zhang2021integrated} or multi-attribute decision-making methods \cite{liu2023improved,liu2019novel,gou2018multiple,GOU2017multimoora,gou2021probabilistic}   usually use 2WD methods.
These methods rank alternatives based on scores to produce decision results, but lack semantic interpretation of the decision results.
In contrast, the proposed method employs a S3WD process that provides meaningful explanations for situations such as medical diagnosis and emergency logistics service provider selection.

(2) In classical GDM methods \cite{krishankumar2019framework,zhang2021integrated}, information fusion is typically handled by aggregation operators that combine all experts' information at once. Non-operator-based information fusion methods consider information fusion from different perspective but still follow a holistic fusion approach.
The proposed S3W-GDM method, however, combines the concept of multi-granularity with conventional thinking.
It introduces a coarse-to-fine granularity approach to information fusion, where initial decisions are made using coarse-grained information, followed by progressively finer-grained analysis to refine the decisions.
This multi-level fusion approach enhances the decision-making process's efficiency and accuracy.
For breast cancer diagnosis, the S3W-GDM method utilizes a multi-level granularity approach that first focuses on key attributes and then continuously improves decision making to improve decision making efficiency.
For emergency logistics provider selection, the S3W-GDM method offers rapid a priori solutions, crucial in emergencies.
The S3W-GDM method achieves stable classification by the $3rd$ decision-level with only half the decision information, enhancing efficiency under time constraints.

(3) The most methods \cite{krishankumar2019framework, liu2023improved,liu2019novel,gou2018multiple,GOU2017multimoora} compared in this study do not adequately address the qualitative expression of decision preferences, particularly under DHHFLTS environment.
Classical methods lack work on the transition from qualitative evaluations to quantitative changes, leading to potential biases in decision-making.
The proposed S3W-GDM method addresses this gap by effectively capturing and incorporating qualitative decision preferences into the decision-making process, ensuring a balanced and comprehensive evaluation.

The advantages between the proposed method and others are summarized as follows:

(1) By incorporating a S3WD process and multi-granularity information fusion, the S3W-GDM method provides comprehensive and interpretable decision results.
This approach not only ranks the alternatives but also classifies them into positive, boundary, and negative regions, offering a clear semantic interpretation of the results.

(2) The S3W-GDM method uses a coarse-to-fine granularity approach, pioneering a new model of information fusion.
Initial decisions are made swiftly using coarse-grained key attributes, providing a rapid preliminary assessment.
Further refinements are then applied to alternatives requiring additional analysis, utilizing finer-grained information.
This multi-level fusion ensures that assessments are balanced and comprehensive, providing an effective way to use qualitative evaluation information.

(3) The novel S3WD of DHHFLTS method to address the problem of uncertainty in decision alternatives by redesigning the computation of conditional probabilities without relying on decision attributes, this approach improves the accuracy.
It incorporates relative utility into the evaluation of each alternative, capturing individual psychological behaviour and also improving decision-making accuracy.

\section{Conclusion and future work}\label{conclusion}
With the progress of society and information science, GDM problems are becoming increasingly complex.
Classical GDM methods, which rely on aggregation operators to fuse information from different attributes and decision-makers at once, significantly increase the decision burden and constrain efficiency.
Moreover, these problems often exhibit vagueness, hesitation, and variation, adding to their complexity.
Existing relative works rarely take these characteristics into account while improving decision-making efficiency by changing the paradigm of information fusion.
Accordingly, the work of this paper is summarised as follows.
First, constructing a neighborhood relation matrix based on derived similarity degrees between alternatives and combining it with the outranking relation to refine conditional probability calculations.
Then, designing a new ``loss function" model for decision risk based on relative perceived utility, incorporating regret theory (RT). This includes defining expert decision tables and multi-level granular extraction and aggregation of evaluations.
These two steps establish the foundation of the novel S3WD of DHHFLTS model.
Further, the paper demonstrates the most efficient operator for aggregation in the decision-level information fusion process, defines a multi-level granular structure, and proposes decision-making strategies and semantic interpretations for each level.
The efficiency and rationality of the established method are validated through illustrative example and comparative analysis with other methods.

In future research, the following three points will be emphasized.
With the development of information science, the volume of tool-focused data for solving complex problems has become larger and larger.
Therefore, the DHHFLTS as a kind of natural language word computing needs to raise the level of dealing with decision-making problems to a large-scale group\cite{tang2021conventional,cheng2024large}, deal with a larger volume of data through machine learning or deep learning algorithms\cite{ding2020large,cheng2024opinions}, and promote the development of the integration of computer science and technology and management science engineering.
Additionally, when the volume of data becomes larger, how to effectively allocate computing resources will also become an important issue.
Finally, no matter what kind of decision-making the ultimate goal is to reach a consensus\cite{wang2020sequential}, the future will be centered on multi-granularity to do consensus research.

\section*{Acknowledgement}
This work was supported by
the National Natural Science Foundation of China (No.62276\\038, No. 62221005), the Joint Fund of Chongqing Natural Science Foundation for Innovation and Development under Grant (No.CSTB2023NSCQ-LZX0164), the Chongqing Talent Program (No.CQYC20210202215),
the Chongqing Municipal Education Commission (HZ2021008),
and the Doctoral Talent Training Program of Chongqing University of Posts and Telecommunications (No.BYJS202213),.

\bibliography{paper_2}\label{references}
\end{spacing}
\end{document}